\pdfoutput=1

\documentclass{article} 
\usepackage{iclr2023_conference,times}

\usepackage{amsmath,amsfonts,bm}

\def\eqref#1{equation~\ref{#1}}

\def\1{\bm{1}}

\DeclareMathAlphabet{\mathsfit}{\encodingdefault}{\sfdefault}{m}{sl}
\SetMathAlphabet{\mathsfit}{bold}{\encodingdefault}{\sfdefault}{bx}{n}

\usepackage[T1]{fontenc}    
\usepackage[utf8]{inputenc} 
\usepackage[backref=page]{hyperref}       
\usepackage{url}            
\usepackage{booktabs}       
\usepackage{amsfonts}       
\usepackage{nicefrac}       
\usepackage{microtype}      
\usepackage{graphicx}
\usepackage{colortbl}
\usepackage{xcolor}         
\usepackage{dcolumn}
\usepackage{siunitx}

\usepackage{setspace}
\usepackage{ifthen}
\usepackage{ascmac}
\usepackage{fancybox}
\usepackage{bm}
\usepackage{amssymb}
\usepackage{amsmath}
\usepackage{multirow}
\usepackage{subcaption}
\usepackage[capitalize]{cleveref}
\hypersetup{
    colorlinks=true,
    citecolor=cyan,
    linkcolor=red,
    urlcolor=magenta,
}

\title{Gromov-Wasserstein Autoencoders}

\author{%
Nao Nakagawa${}^1$, Ren Togo${}^2$, Takahiro Ogawa${}^2$, \& Miki Haseyama${}^2$ \\
${}^1$ Graduate School of Information Science and Technology, Hokkaido University, Japan \\
${}^2$ Faculty of Information Science and Technology, Hokkaido University, Japan \\
\texttt{\{nakagawa,togo,ogawa,mhaseyama\}@lmd.ist.hokudai.ac.jp}
}

\newcommand{\finalvspace}{-4.5px}

\newcommand{\writingmode}{FINAL}

\ifthenelse{\equal{\writingmode}{PREVIEW}}{ %
    \newcommand{\wip}[1]{{\PackageWarning{Todo}{Detection  TODO:#1}\textcolor{red}{\textbf{[[[WIP: #1]]]}}}} %
    \newcommand{\revise}[1]{\textcolor{red}{#1}} %
}{ %
    \newcommand{\wip}[1]{} %
    \newcommand{\revise}[1]{#1} %
}

\ifthenelse{\equal{\writingmode}{CORRECTION-DOUBLE}}{ %
}{ %
}

\usepackage{xspace}
\makeatletter
\DeclareRobustCommand\onedot{\futurelet\@let@token\@onedot}
\def\@onedot{\ifx\@let@token.\else.\null\fi\xspace}

\def\eg{\emph{e.g}\onedot} 
\def\ie{\emph{i.e}\onedot}

\def\wrt{w.r.t\onedot} 

\makeatother

\renewcommand{\cite}[1]{\citep{#1}}

\newcommand{\x}{\mathbf{x}}
\newcommand{\y}{\mathbf{y}}
\newcommand{\z}{\mathbf{z}}

\newcommand{\xsp}{\mathcal{X}}

\newcommand{\zsp}{\mathcal{Z}}

\newcommand{\paramEnc}{\bm{\phi}}
\newcommand{\idistro}{q_{\paramEnc}}
\newcommand{\idistropermuted}{\bar{q}_{\paramEnc}}
\newcommand{\enc}[2]{\idistro(#1|#2)}
\newcommand{\ijoint}[2]{\idistro(#1,#2)}
\newcommand{\iagg}[1]{\idistro(#1)}
\newcommand{\iaggpermuted}[1]{\idistropermuted(#1)}
\newcommand{\data}{\mathrm{data}}
\newcommand{\datadistro}{p_{\data}}
\newcommand{\empir}[1]{\datadistro(#1)}

\newcommand{\paramDec}{\bm{\theta}}
\newcommand{\gdistro}{p_{\paramDec}}
\newcommand{\dec}[2]{\gdistro(#1|#2)}

\newcommand{\gjoint}[2]{\gdistro(#1,#2)}
\newcommand{\gmodel}[1]{\gdistro(#1)}
\newcommand{\pidistro}{\pi}
\newcommand{\trainablePidistro}{\pi_{\paramDec}}
\newcommand{\prior}[1]{\pidistro(#1)}
\newcommand{\trainablePrior}[1]{\trainablePidistro(#1)}

\newcommand{\gwd}{GW}
\newcommand{\wsd}{W}

\newcommand{\real}{\mathbb{R}}

\newcommand{\couplingset}{\mathcal{P}}

\newcommand{\ev}[1]{\mathbb{E}_{#1}}

\newcommand{\tr}{\mathrm{tr}}
\newcommand{\T}{\mathsf{T}}

\newcommand{\kld}[2]{{D_\mathrm{KL}}\!\left(#1\middle\|#2\right)}
\newcommand{\totcol}[1]{\mathrm{TC}(#1)}

\newcommand{\distfunc}[1]{d_{#1}}
\newcommand{\distx}{\distfunc{\xsp}}

\newcommand{\distz}{\distfunc{\zsp}}

\newcommand{\onedistro}{r}
\newcommand{\otherdistro}{s}

\newcommand{\elbo}[3]{\mathrm{ELBO}(#1;#2,#3)}
\newcommand{\maximize}{\mathop{\text{maximize}}}
\newcommand{\minimize}{\mathop{\text{minimize}}}
\newcommand{\subjectto}{\mathop{\text{subject to}}}

\newcommand{\loss}{\mathcal{L}}
\newcommand{\regularization}{\mathcal{R}}
\newcommand{\lossgw}{\loss_{GW}}
\newcommand{\lossw}{\loss_{W}}
\newcommand{\lossd}{\loss_{D}}

\newcommand{\lossh}{\regularization_{\mathcal{H}}}
\newcommand{\coeff}{\lambda}

\newcommand{\coeffw}{\coeff_{W}}
\newcommand{\coeffd}{\coeff_{D}}
\newcommand{\coeffgp}{\coeff_{GP}}
\newcommand{\coeffh}{\coeff_{\mathcal{H}}}

\newcommand{\worder}{\xi}
\newcommand{\gworder}{\rho}

\newcommand{\paramCritic}{\bm{\psi}}
\newcommand{\critic}{f_{\paramCritic}}
\newcommand{\xintp}{\tilde{\x}}
\newcommand{\zintp}{\tilde{\z}}

\newcommand{\facPrior}[2]{\tilde{\pi}_{\paramDec}^{(#1)}(#2)}

\newcommand{\disc}{\mathrm{Disc}}
\newcommand{\mmd}{\mathrm{MMD}}

\iclrfinalcopy 
\begin{document}

\maketitle

\begin{abstract}
Variational Autoencoder~(VAE)-based generative models offer flexible representation learning by incorporating meta-priors, general premises considered beneficial for downstream tasks.
However, the incorporated meta-priors often involve ad-hoc model deviations from the original likelihood architecture, causing undesirable changes in their training.
In this paper, we propose a novel representation learning method, Gromov-Wasserstein Autoencoders~(GWAE), which directly matches the latent and data distributions using the variational autoencoding scheme.
Instead of likelihood-based objectives, GWAE models minimize the Gromov-Wasserstein (GW) metric between the trainable prior and given data distributions.
The GW metric measures the distance structure-oriented discrepancy between distributions even with different dimensionalities, which provides a direct measure between the latent and data spaces.
By restricting the prior family, we can introduce meta-priors into the latent space without changing their objective.
The empirical comparisons with VAE-based models show that GWAE models work in two prominent meta-priors, disentanglement and clustering, with their GW objective unchanged.
\end{abstract}

\section{Introduction} \label{sec:intro}
One fundamental challenge in unsupervised learning is capturing the underlying low-dimensional structure of high-dimensional data because natural data (\eg, images) lie in low-dimensional manifolds~\cite{Carlson2008,Bengio2013}.
Since deep neural networks have shown their potential for non-linear mapping, representation learning has recently made substantial progress in its applications to high-dimensional and complex data~\cite{VAE,Rezende2014,Hsu2017,Hu2017}.
Learning low-dimensional representations is in mounting demand because the inference of concise representations extracts the essence of data to facilitate various downstream tasks~\cite{Thomas2017,DARLA,Creager2019,Locatello2019Fairness}.
For obtaining such general-purpose representations, several \emph{meta-priors} have been proposed~\cite{Bengio2013,Tschannnen2018}.
Meta-priors are general premises about the world, such as disentanglement~\cite{BetaVAE,BetaTCVAE,FactorVAE,GuidedVAE}, hierarchical factors~\cite{NVAE,VLadderAE,LadderVAE}, and clustering~\cite{AAE,DAGMM,Asano2020}.

A prominent approach to representation learning is a deep generative model based on the variational autoencoder~(VAE)~\cite{VAE}.
VAE-based models adopt the variational autoencoding scheme, which introduces an inference model in addition to a generative model and thereby offers bidirectionally tractable processes between observed variables~(data) and latent variables.
In this scheme, the reparameterization trick~\cite{VAE} yields representation learning capability since reparameterized latent codes are tractable for gradient computation.
The introduction of additional losses and constraints provides further regularization for the training process based on meta-priors.
However, controlling representation learning remains a challenging task in VAE-based models owing to the deviation from the original optimization.
Whereas the existing VAE-based approaches modify the latent space based on the meta-prior~\cite{FactorVAE,VLadderAE,DAGMM}, their training objectives still partly rely on the evidence lower bound~(ELBO).
Since the ELBO objective is grounded on variational inference, ad-hoc model modifications cause implicit and undesirable changes, \eg, posterior collapse~\cite{Dai2020} and implicit prior change~\cite{Hoffman2017} in $\beta$-VAE~\cite{BetaVAE}.
Under such modifications, it is also unclear whether a latent representation retains the underlying data structure because VAE models implicitly
interpolate data points to form a latent space using noises injected into latent codes by the reparameterization trick~\cite{GECO,TamingVAEs,NCPVAE}.

As another paradigm of variational modeling, the ELBO objective has been reinterpreted from the optimal transport~(OT) viewpoint~\cite{WAE}.
\citet{WAE} have derived a family of generative models called the Wasserstein autoencoder~(WAE) by applying the variational autoencoding model to high-dimensional OT problems as the couplings \revise{(\cref{sec:appendix-wae} for more details)}.
Despite the OT-based model derivation, the WAE objective is equivalent to that of InfoVAE~\cite{InfoVAE}, whose objective consists of the ELBO and the mutual information term.
The WAE formulation is derived from the estimation and minimization of the OT cost~\cite{WAE,WGAN} between the data distribution and the generative model, \ie, the generative modeling by applying the Wasserstein metric.
It furnishes a wide class of models, even when the prior support does not cover the entire variational posterior support.
The OT paradigm also applies to existing representation learning approaches originally derived from re-weighting the Kullback-Leibler~(KL) divergence term~\cite{TCWAE}.

Another technique for optimizing the VAE-based ELBO objective called implicit variational inference~(IVI)~\cite{Huszar2017} has been actively researched.
While the VAE model has an analytically tractable prior for variational inference, IVI aims at variational inference using \emph{implicit distributions}, in which one can use its sampler instead of its probability density function.
A notable approach to IVI is the density ratio estimation~\cite{Sugiyama2012Book}, which replaces the $f$-divergence term in the variational objective with an adversarial discriminator that distinguishes the origin of the samples.
For distribution matching, this algorithm shares theoretical grounds with generative models based on the generative adversarial networks~(GANs)~\cite{GAN,Sonderby2017}, which induces the application of IVI toward the distribution matching in complex and high-dimensional variables, such as images.
\revise{See \cref{sec:appendix-ivi} for more discussions.}

In this paper, we propose a novel representation learning methodology, \emph{Gromov-Wasserstein Autoencoder}~(GWAE) based on the Gromov-Wasserstein~(GW) metric~\cite{Memoli2011}, an OT-based metric between distributions applicable even with different dimensionality~\cite{Memoli2011,RAE,SFG}.
Instead of the ELBO objective, we apply the GW metric objective in the variational autoencoding scheme to directly match the latent marginal~(prior) and the data distribution.
The GWAE models obtain a latent representation retaining the distance structure of the data space to hold the underlying data information.
The GW objective also induces the variational autoencoding to perform the distribution matching of the generative and inference models, despite the OT-based derivation.
Under the OT-based variational autoencoding, one can adopt a prior of a GWAE model from a rich class of trainable priors depending on the assumed meta-prior even though the KL divergence from the prior to the encoder is infinite.
Our contributions are listed below.
\begin{itemize}
    \item 
        We propose a novel probabilistic model family GWAE, which matches the latent space to the given unlabeled data via the variational autoencoding scheme.
        The GWAE models estimate and minimize the GW metric between the latent and data spaces to directly match the latent representation closer to the data in terms of distance structure.
    \item 
        We propose several families of priors in the form of implicit distributions, adaptively learned from the given dataset using stochastic gradient descent~(SGD).
        The choice of the prior family corresponds to the meta-prior, thereby providing a more flexible modeling scheme for representation learning.
    \item 
        We conduct empirical evaluations on the capability of GWAE in prominent meta-priors: disentanglement and clustering.
        Several experiments on image datasets CelebA~\cite{CelebA}, MNIST~\cite{MNIST}, and 3D Shapes~\cite{3DShapes}, show that GWAE models outperform the VAE-based representation learning methods whereas their GW objective is not changed over different meta-priors.
\end{itemize}

\section{Related Work} \label{sec:related}

\textbf{VAE-based Representation Learning.}
%
%
%
VAE~\cite{VAE} is a prominent deep generative model for representation learning.
Following its theoretical consistency and explicit handling of latent variables, many state-of-the-art representation learning methods are proposed based on VAE with modification~\cite{BetaVAE,BetaTCVAE,FactorVAE,Achille2018Bottleneck,DIP-VAE,DAGMM,VLadderAE,LadderVAE,InfoVAE,DFCVAE,Detlefsen2019,GuidedVAE}.
The standard VAE learns an encoder and a decoder with parameters~$\paramEnc$ and~$\paramDec$, respectively, to learn a low-dimensional representation in its latent variables~$\z$ using a bottleneck layer of the autoencoder.
Using data~$\x \in \empir{\x}$ supported on the data space~$\xsp$, the VAE objective is the ELBO formulated by the following optimization problem:
\begin{align}
    \maximize_{\paramDec, \paramEnc} \quad & \ev{\empir{\x}} \left[ \ev{\enc{\z}{\x}} \left[ \log \dec{\x}{\z} \right] - \kld{\enc{\z}{\x}}{\prior{\z}} \right], \label{eq:vae-objective}
\end{align}
where the encoder~$\enc{\z}{\x}$ and decoder~$\dec{\x}{\z}$ are parameterized by neural networks, and the prior~$\prior{\z}$ is postulated before training.
The first and second terms~(called the \emph{reconstruction term} and the \emph{KL term}, respectively) in \cref{eq:vae-objective} are in a trade-off relationship~\cite{Tschannnen2018}.
This implies that learning is guided to autoencoding by the reconstruction term while matching the distribution of latent variables to the pre-defined prior using the KL term.

\textbf{Implicit Variational Inference.}
IVI solves the variational inference problem using implicit distributions~\cite{Huszar2017}.
A major approach to IVI is density ratio estimation~\cite{Sugiyama2012Book}, in which the ratio between probability distribution functions is estimated using a discriminator instead of their closed-form expression.
Since IVI-based and GAN-based models share density ratio estimation mechanisms in distribution matching~\cite{Sonderby2017}, the combination of VAEs and GANs has been actively studied, especially from the aspect of the matching of implicit distributions.
The successful results achieved by GAN-based models in high-dimensional data, such as natural images, have propelled an active application and research of IVI in unsupervised learning~\cite{VAEGAN,IAE}.

\textbf{Optimal Transport.}
The OT cost is used as a measure of the difference between distributions supported on high-dimensional space using SGD~\cite{WGAN,WAE,TCWAE}.
This provides the Wasserstein metric for the discrepancy between distributions.
For a constant $\xi \ge 1$, the $\xi$-Wasserstein metric between distributions~$\onedistro$ and~$\otherdistro$ is defined as
\begin{align}
    \wsd_{\xi}(\onedistro,\otherdistro) = \left(
        \inf_{\gamma \in \couplingset(\onedistro(\x), \otherdistro(\x'))}
            \ev{\gamma(\x,\x')} \left[
                d^{\xi}(\x,\x')
            \right]
    \right)^{1/{\xi}}, \label{eq:wsd-definition}
\end{align}
where $\x$ denotes the random variable in which the distributions~$\onedistro$ and~$\otherdistro$ are defined, and $\couplingset(\onedistro(\x), \otherdistro(\x'))$ denotes the set consisting of all couplings whose~$\x$-marginal is~$\onedistro(\x)$ and whose~$\x'$-marginal is~$\otherdistro(\x')$.
Owing to the difficulty of computing the exact infimum in \cref{eq:wsd-definition} for high-dimensional, large-scale data, several approaches try to minimize the estimated $\xi$-Wasserstein metric using neural networks and SGD~\cite{WAE,WGAN}.
The form in \cref{eq:wsd-definition} is the primal form of the Wasserstein metric, particularly compared with its dual form for the case of~$\xi=1$~\cite{WGAN}.
The two prominent approaches for the OT in high-dimensional, complex large-scale data are: (i) minimizing the primal form using a probabilistic autoencoder~\cite{WAE}, and (ii) adversarially optimizing the dual form using a generator-critic pair~\cite{WGAN}.

\textbf{Wasserstein Autoencoder~(WAE).}
WAE~\cite{WAE} is a family of generative models whose autoencoder estimates and minimizes the primal form of the Wasserstein metric between the generative model~$\gmodel{\x}$ and the data distribution~$\empir{\x}$ using SGD in the variational autoencoding settings, \ie, the VAE model architecture~\cite{VAE}.
This primal-based formulation induces a representation learning methodology from the OT viewpoint because the WAE objective is equivalent to that of InfoVAE~\cite{InfoVAE}, which learns the variational autoencoding model by retaining the mutual information of the probabilistic encoder.

\textbf{Kantorovich-Rubinstein Duality.}
The Wasserstein GAN models~\cite{WGAN} adopt an objective based on the~1-Wasserstein metric between the generative model~$\gmodel{\x}$ and data distribution~$\empir{\x}$.
This objective is estimated using the Kantorovich-Rubinstein duality~\cite{Villiani2010,WGAN}, which holds for the $1$-Wasserstein as
\begin{align}
    \wsd_1(\onedistro,\otherdistro) = \sup_{f:\text{1-Lipschitz}}
          \ev{\onedistro(\x)} \left[ f(\x) \right]
        - \ev{\otherdistro(\x)} \left[ f(\x) \right]. \label{eq:kr-duality}
\end{align}
To estimate this function $f$ using SGD, a 1-Lipschitz neural network called a \emph{critic} is introduced, as with a discriminator in the GAN-based models.
The training process using mini-batches is adversarially conducted, \ie, by repeating updates of the critic parameters and the generative parameters alternatively.
During this process, the critic maximizes the objective in \cref{eq:kr-duality} to approach the supremum, whereas the generative model minimizes the objective for the distribution matching~$\gmodel{\x} \approx \empir{\x}$.

\section{Proposed Method} \label{sec:method}

Our GWAE models minimize the OT cost between the data and latent spaces, based on generative modeling in the variational autoencoding.
GWAE models learn representations by matching the distance structure between the latent and data spaces, instead of likelihood maximization.


\subsection{Optimal Transport between Spaces}
\vspace{\finalvspace}

Although the OT problem induces a metric between probability distributions, its application is limited to distributions sharing one sample space.
The GW metric~\cite{Memoli2011} measures the discrepancy between metric measure spaces using the OT of distance distributions.
A metric measure space consists of a sample space, metric, and probability measure.
Given a pair of different metric spaces, \ie, sample spaces and metrics, the GW metric measures the discrepancy between probability distributions supported on the spaces.
In terms of the GW metric, two distributions are considered to be equal if there is an isometric mapping between their supports~\cite{Sturm2012,Sejourne2021}.
For a constant~$\rho \ge 1$, the formulation of the $\rho$-GW metric between probability distributions $\onedistro(\x)$ supported on a metric space~$(\xsp, \distx)$ and~$\otherdistro(\z)$ supported on~$(\zsp, \distz)$ is given by
\begin{align}
    \gwd_\gworder(\onedistro,\otherdistro)
    &:= \left(
        \inf_{\gamma \in \couplingset(\onedistro(\x), \otherdistro(\z))}
            \ev{\gamma(\x,\z)}
            \ev{\gamma(\x',\z')}
            \left[
                \left|
                    \distx(\x,\x') - \distz(\z,\z')
                \right|^\gworder
            \right]
        \right)^{1/\gworder},
    \label{eq:gwmetric}
\end{align}
where $\couplingset(\onedistro(\x), \otherdistro(\z))$ denotes the set of all couplings with $\onedistro(\x)$ as $\x$-marginal and $\otherdistro(\z)$ as $\z$-marginal.
The metrics~$\distx$ and~$\distz$ are the metrics in the spaces~$\xsp$ and~$\zsp$, respectively.

\subsection{Application to Representation Learning: Gromov-Wasserstein Autoencoder}
\vspace{\finalvspace}


In this work, we propose a novel GWAE modeling methodology based on the GW metric for distance structure modeling in the variational autoencoding formulation.
\revise{
The objectives of generative models typically aim for distribution matching in the data space, \eg, the likelihood~\cite{VAE} and the Jensen-Shannon divergence~\cite{GAN}.
The GWAE objective differs from these approaches and aims to directly match the latent and data distributions based on their distance structure.
}

\subsubsection{Model Settings: Variational Autoencoding \label{sec:model-settings}}
\vspace{\finalvspace}

Given an~$N$-sized set of data points~$\{\x_i\}_{i=1}^N$ supported on a data space~$\xsp$, representation learning aims to build a latent space~$\zsp$ and obtain mappings between both the spaces.
For numerical computation, we postulate that the spaces~$\xsp$ and~$\zsp$ respectively have tractable metrics~$\distx$ and~$\distz$ such as the Euclidean distance~(see \cref{sec:appendix-model} for details), and
let~$M,L\in\mathbb{N}\setminus\{0\}$, $\xsp \subseteq \real^M$, and $\zsp \subseteq \real^L$.
We mention the bottleneck case~$M \gg L$ similarly to the existing representation learning methods~\cite{VAE,BetaVAE,FactorVAE} because the data space~$\xsp$ is typically an~$L$-dimensional manifold~\cite{Carlson2008,Bengio2013}.

We construct a model with a trainable latent prior~$\trainablePrior{\z}$ to approach the data distribution~$\empir{\x}$ in terms of distance structure.
Following the standard VAE~\cite{VAE}, we consider a generative model~$\gjoint{\x}{\z}$ with parameters~$\paramDec$ and an inference model~$\ijoint{\x}{\z}$ with parameters~$\paramEnc$.
The generation process consists of the prior~$\trainablePrior{\z}$ and a decoder~$\dec{\x}{\z}$ parameterized with neural networks.
Since the inverted generation process~$\dec{\z}{\x}=\trainablePrior{\z}\dec{\x}{\z}/\gmodel{\x}$ is intractable in this scheme, an encoder~$\enc{\z}{\x} \approx \dec{\z}{\x}$ is instead established using neural networks for parameterization.
Thus, the generative~$\gjoint{\x}{\z}$ and inference~$\ijoint{\x}{\z}$ models are defined as
\begin{align}
    \gjoint{\x}{\z} &= \trainablePrior{\z} \dec{\x}{\z}, &
    \ijoint{\x}{\z} &= \empir{\x} \enc{\z}{\x}.
\end{align}
The empirical~$\hat{p}_{\data}(\x) = 1/N \sum_{i=1}^{N} \delta(\x-\x_i)$ is used for the estimation of $\empir{\x}$.
A Dirac decoder and a diagonal Gaussian encoder are used to alleviate deviations from the data manifold as in~\citet{WAE} (see \cref{sec:appendix-model} for these details and formulations).

\subsubsection{Optimal Transport Objective} \label{sec:ot-objective}
\vspace{\finalvspace}


Here, we focus on the latent space~$\zsp$ to transfer the underlying data structure to the latent space.
This highlights the main difference between the GWAE and the existing generative approaches.
The training objective of GWAE is the GW metric between the metric measure spaces~$(\xsp,\distx,\empir{\x})$ and~$(\zsp,\distz,\trainablePrior{\z})$ as
\begin{align}
    \minimize_{\paramDec} \quad & \gwd_\gworder(\empir{\x}, \trainablePrior{\z})^\gworder, \label{eq:min-gw}
\end{align}
where $\gworder \ge 1$ is a constant, and we adopt $\gworder=1$ to alleviate the effect of outlier samples distant from the isometry for training stability.
Computing the exact GW value is difficult owing to the high dimensionality of both~$\x$ and~$\z$.
Hence, we estimate and minimize the GW metric using the variational autoencoding scheme, which captures the latent factors of complex data in a stable manner.
\revise{We recast the GW objective} into a main GW estimator~$\lossgw$ with three regularizations: a reconstruction loss~$\lossw$, a joint dual loss~$\lossd$, and an entropy regularization~$\lossh$.

\textbf{Estimated GW metric~$\lossgw$.}
We use the generative model~$\gjoint{\x}{\z}$ as the coupling of \cref{eq:min-gw} similarly to the WAE~\cite{WAE} methodology.
The main loss~$\lossgw$ estimates the GW metric as:
\begin{align}
    \minimize_{\paramDec} \quad &
        \lossgw := \ev{ \gjoint{\x}{\z} } \ev{ \gjoint{\x'}{\z'} } \left[
            \left|
                \distx(\x,\x') - C \distz(\z,\z')
            \right|^\gworder
        \right], \label{eq:loss-gw}\\
    \subjectto \quad & \empir{\x} = \gmodel{\x}, \label{eq:loss-gw-cond}
\end{align}
where $C$ is a trainable scale constant to cancel out the scale degree of freedom, and $\gmodel{\x}$ denotes the marginal~$\gmodel{\x}=\int_{\zsp} \gjoint{\x}{\z} d\z$.

\textbf{WAE-based $\xsp$-marginal condition~$\lossw$.}
To obtain a numerical solution with stable training, \citet{WAE} relax the $\xsp$-matching condition of \cref{eq:loss-gw-cond} into $\worder$-Wasserstein minimization~$(\worder \ge 1)$ using the variational autoencoding coupling.
The WAE methodology~\cite{WAE} uses the inference model~$\ijoint{\x}{\z}$ to formulate the $\worder$-Wasserstein minimization as the reconstruction loss~$\lossw$ with a $\zsp$-matching condition as:
\begin{align}
    \minimize_{\paramDec, \paramEnc} \quad &
        \lossw := \ev{ \ijoint{\x}{\z} } \ev{ \dec{\x'}{\z} } \left[
                \distx(\x,\x')
        \right], \label{eq:loss-w} \\
    \subjectto \quad &
        \iagg{\z} = \trainablePrior{\z}. \label{eq:loss-w-cond}
\end{align}
where $\distx$ is a distance function based on the $L_{\worder}$ metric.
We adopt the settings~$\worder=2$ to retain the conventional Gaussian reconstruction loss.

\textbf{Merged sufficient condition~$\lossd$.}
We merge the marginal coupling conditions of \cref{eq:loss-gw-cond} and \cref{eq:loss-w-cond} into the joint $\xsp\times\zsp$-matching sufficient condition~$\gjoint{\x}{\z}=\ijoint{\x}{\z}$ to attain bidirectional inferences while preserving the stability of autoencoding.
Since such joint distribution matching can also be relaxed into the minimization of $W_1(\ijoint{\x}{\z}, \gjoint{\x}{\z})$, this condition is satisfied by minimizing the Kantorovich-Rubinstein duality introduced by \citet{WGAN} as in \cref{eq:kr-duality}.
Practically, a 1-Lipschitz neural network (critic)~$\critic$ estimates the supremum of \cref{eq:kr-duality}, and the main model minimizes this estimated supremum as:
\begin{align}
    \minimize_{\paramDec,\paramEnc} \  \maximize_{\paramCritic} \quad \lossd :=  \ev{\ijoint{\x}{\z}} \left[ \critic(\x,\z) \right]
    - \ev{\gjoint{\x}{\z}} \left[ \critic(\x,\z) \right], \label{eq:loss-d}
\end{align}
where $\paramCritic$ is the critic parameters.
To satisfy the 1-Lipschitz constraint, the critic~$\critic$ is implemented with techniques such as spectral normalization~\cite{SNGAN} and gradient penalty~\cite{WGANGP} (see \cref{sec:appendix-gp} for the details of the gradient penalty loss).

\textbf{Entropy regularization~$\lossh$.}
We further introduce the entropy regularization~$\lossh$ using the inference entropy to avoid degenerate solutions in which the encoder~$\enc{\z}{\x}$ becomes Dirac and deterministic for all data points.
In such degenerate solutions, the latent representation simply becomes a look-up table because such a point-to-point encoder maps the set of data points into a set of latent code points with measure zero~\cite{Hoffman2017,Dai2018}, causing overfitting into the empirical data distribution.
An effective way to avoid it is a regularization with the inference entropy~$\mathcal{H}_q$ of the latent variables~$\z$ conditioned on data~$\x$ as
\begin{align}
    \lossh
    &:= \mathcal{H}_q(\z|\x) = \ev{\ijoint{\x}{\z}} \left[
        -\log \enc{\z}{\x}
    \right].
\end{align}
Since the conditioned entropy~$\mathcal{H}_q(\z|\x)$ diverges to negative infinity in the degenerate solutions, the regularization term $-\lossh$ facilitates the probabilistic learning of GWAE models.

\textbf{Stochastic Training with Single Estimated Objective.}
Applying the Lagrange multiplier method to the aforementioned constraints, we recast the GW metric of \cref{eq:min-gw} into a single objective~$\loss$ with multipliers~$\coeffw$, $\coeffd$, and $\coeffh$ as
\begin{align}
    \minimize_{\paramDec,\paramEnc} \  \maximize_{\paramCritic} \quad &
        \loss := \lossgw + \coeffw \lossw + \coeffd \lossd - \coeffh \lossh. \label{eq:total-loss}
\end{align}
One efficient solution to optimize this objective is using the mini-batch gradient descent in alternative steps~\cite{GAN,WGAN}, which we can conduct in automatic differentiation packages, such as PyTorch~\cite{PyTorch}.
One step of mini-batch descent is the minimization of the total objective~$\loss$ in \cref{eq:total-loss}, and the other step is the maximization of the critic objective~$\lossd$ in \cref{eq:loss-d}.
By alternatively repeating these steps, the critic estimates the Wasserstein metric using the expected potential difference~$\lossd$~\cite{WGAN}.
Although the objective in \cref{eq:total-loss} involves three auxiliary regularizations including an adversarial term, the GWAE model can be efficiently optimized because the adversarial mechanism and the variational autoencoding scheme share the goal of distribution matching~$\gjoint{\x}{\z}\approx\ijoint{\x}{\z}$ (see \cref{sec:training-process} for more details).

\subsubsection{Prior by Sampling}
\vspace{\finalvspace}
GWAE models apply to the cases in which the prior~$\trainablePrior{\z}$ takes the form of an implicit distribution with a sampler.
An implicit distribution~$\trainablePrior{\z}$ provides its sampler~$\z \sim \trainablePrior{\z}$ while a closed-form expression of the probability density function is not available.
The adversarial algorithm of GWAE handles such cases and enables a wide class of priors to provide meta-prior-based inductive biases for unsupervised representation learning, \eg, for disentanglement~\cite{Locatello2019,Locatello2020SoberLook}.
Note that the GW objective in \cref{eq:min-gw} becomes a constant function in non-trainable prior cases.


\textbf{Neural Prior~(NP).}
A straightforward way to build a differentiable sampler of a trainable prior is using a neural network to convert noises.
The prior of the latent variables~$\z$ is defined via sampling using a neural network~$g_{\paramDec}: \real^L\to\real^L$ with parameters~$\paramDec$ (see \cref{sec:appendix-prior} for its formulation).
Notably, the neural network~$g_{\paramDec}$ need not be invertible unlike Normalizing Flow~\cite{Rezende2015} since the prior is defined as an implicit distribution not requiring a push-forward measure.

\textbf{Factorized Neural Prior~(FNP).}
For disentanglement, we can constitute a factorized prior using an element-wise independent neural network~$\tilde{g}_{\paramDec}=\{\tilde{g}^{(i)}_{\paramDec}\}_{i=1}^L$ (see \cref{sec:appendix-prior} for its formulation).
Such factorized priors can be easily implemented utilizing the 1-dimensional grouped convolution~\cite{AlexNet}.

\textbf{Gaussian Mixture Prior~(GMP).}
For clustering structure, we construct a class of Gaussian mixture priors.
Given that the prior contains $K$ components, the $k$-th component is parameterized using the weights~$w_k$, means~$\mathbf{m}_k \in \real^L$, and square-root covariances~$\mathbf{M}_k \in \real^{L \times L}$ as
\begin{align}
    \trainablePrior{\z} = \sum_{k=1}^K w_k \mathcal{N}(\z|\mathbf{m}_k, \mathbf{M}_k \mathbf{M}_k^\T),
\end{align}
where the weights~$\{ w_k \}_{k=1}^K$ are normalized as $\sum_{k=1}^K w_k = 1$.
To sample from a prior of this class, one randomly chooses a component $k$ from the $K$-way categorical distribution with probabilities~$(w_1,w_2,\ldots,w_k)$ and draws a sample~$\z$ as follows:
\begin{align}
    \z &= \mathbf{m}_k + \mathbf{M}_k \bm{\epsilon},
    \quad\quad\quad
    \bm{\epsilon} \sim \mathcal{N}(\mathbf{0}, \mathbf{I}_L),
\end{align}
where $\mathbf{0}$ and $\mathbf{I}_n$ denote the zero vector and the~$n$-sized identity matrix, respectively.
In this class of priors, the set of trainable parameters consists of~$\{(w_k, \mathbf{m}_k, \mathbf{M}_k)\}_{k=1}^K$.
Note that this parameterization can be easily implemented in differentiable programming frameworks because~$\mathbf{M}_k \mathbf{M}_k^\T$ is positive semidefinite for any~$\mathbf{M}_k \in \real^{L \times L}$.

\section{Experiments} \label{sec:experiments}
We investigated the wide capability of the GWAE models for learning representations based on meta-priors.\footnote{In the tables of the quantitative evaluations, $\uparrow$ and $\downarrow$ indicate scores in which higher and lower values are better, respectively.}
We evaluated GWAEs in two principal meta-priors: disentanglement and clustering.
To validate the effectiveness of GWAE on different tasks for each meta-prior, we conducted each experiment in corresponding experimental settings.
We further studied their autoencoding and generation for the inspection of general capability.

\subsection{Experimental Settings}
\vspace{\finalvspace}

We compared the GWAE models with existing representation learning methods (see \cref{sec:appendix-related} for the details of the compared methods).
For the experimental results in this section, we used four visual datasets: CelebA~\cite{CelebA}, MNIST~\cite{MNIST}, 3D Shapes~\cite{3DShapes}, and Omniglot~\cite{Omniglot} (see \cref{sec:appendix-datasets} for dataset details).
For quantitative evaluations, we selected hyperparameters from $\coeffw\in[10^0, 10^1]$, $\coeffd\in[10^0,10^1]$, and $\coeffh\in[10^{-4},10^0]$ using their performance on the validation set.
For fair comparisons, we trained the networks with a consistent architecture from scratch in all the methods~(see \cref{sec:appendix-architecture} for architecture details).

\subsection{Gromov-Wasserstein Estimation and Minimization} \label{sec:experiments-isometry}
\vspace{\finalvspace}


\begin{figure}[!t]
    \centering
    \hspace*{\fill}
    \begin{minipage}[t]{0.55\linewidth}
        \centering
        \includegraphics[width=\linewidth,clip]{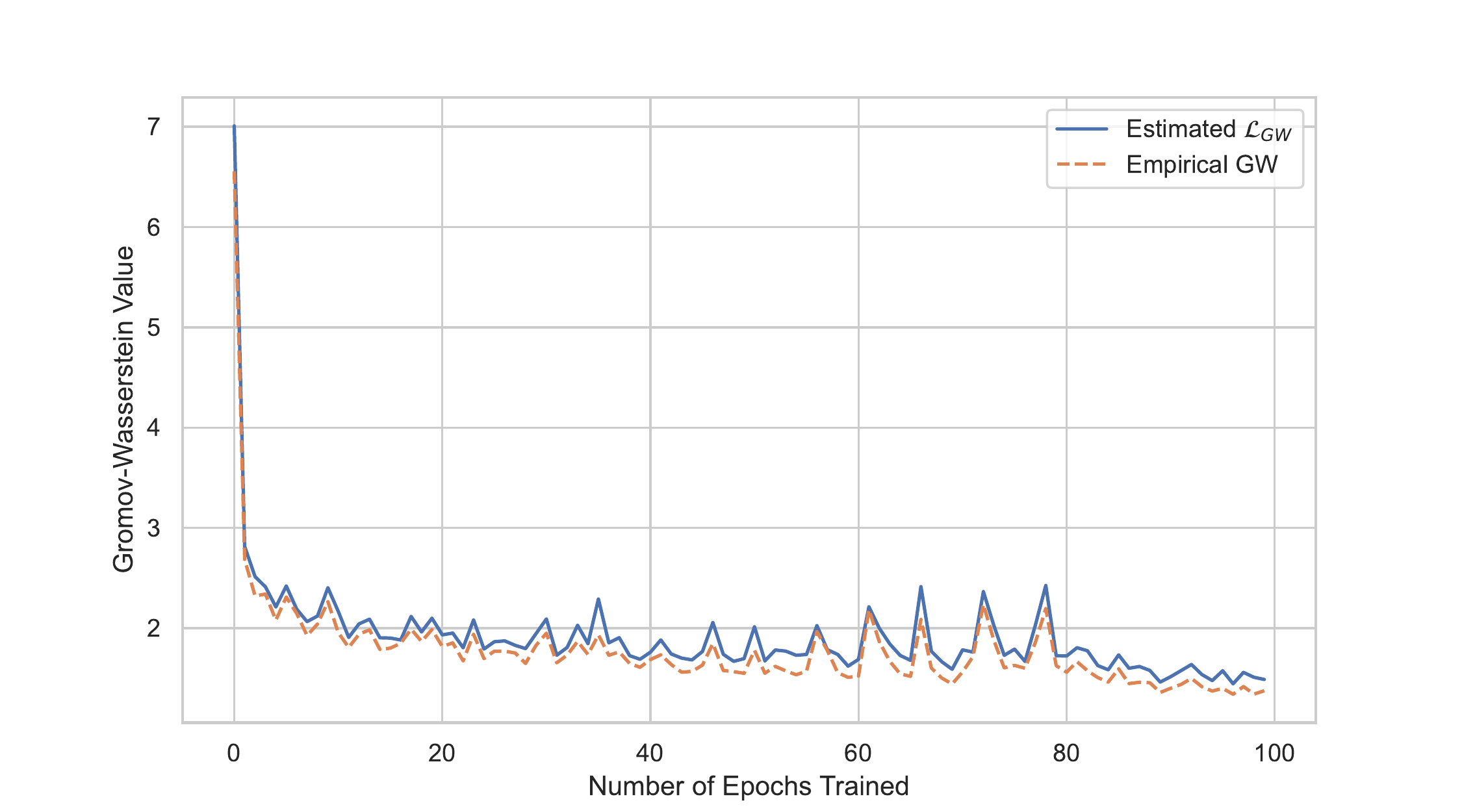}
        \subcaption{The estimation of the GW metric in each epoch.}
        \label{fig:gw-estmin-est}
    \end{minipage}
    \hfill
    \begin{minipage}[t]{0.4\linewidth}
        \centering
        \includegraphics[width=\linewidth,clip]{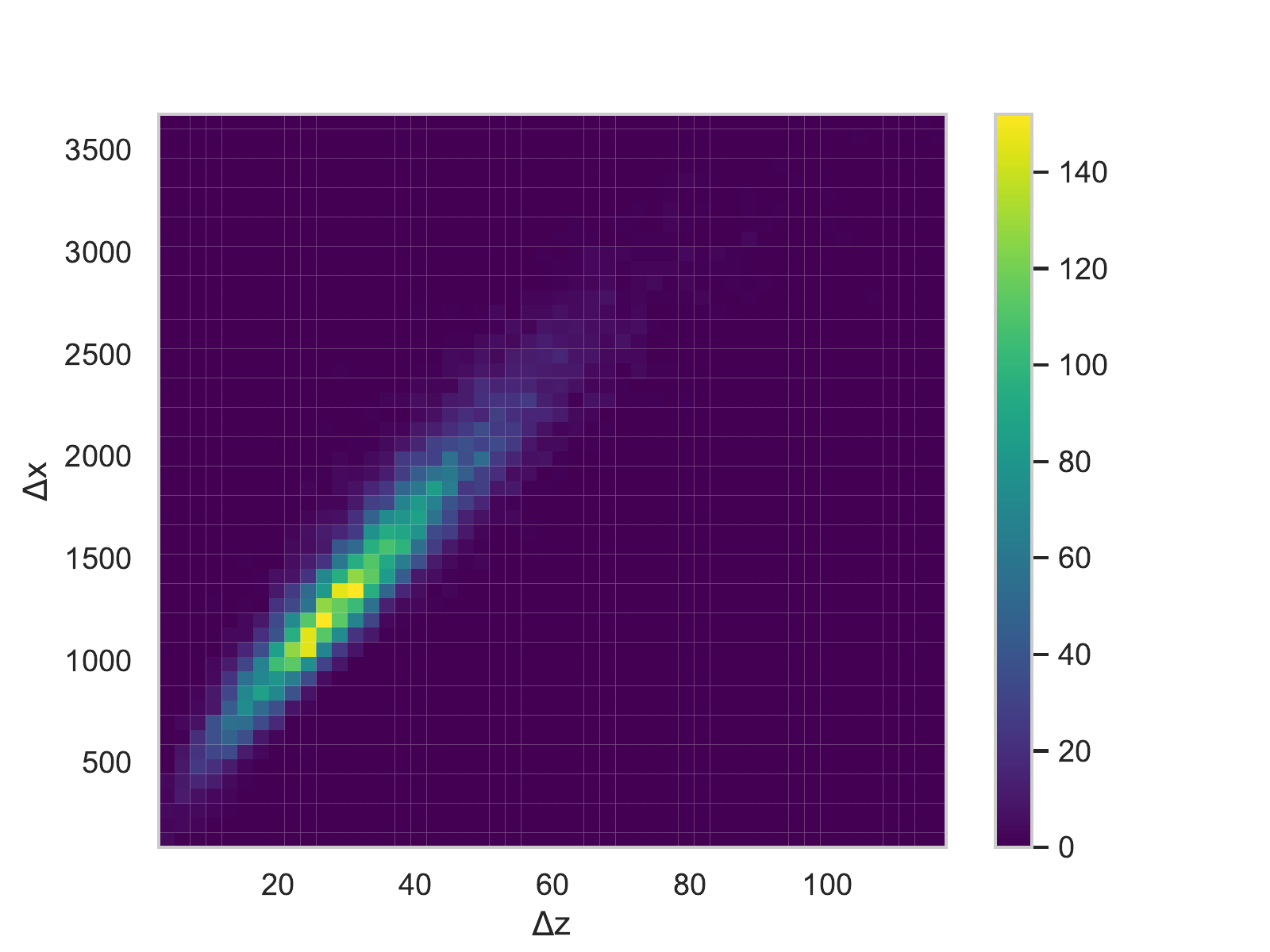}
        \subcaption{The isometry in GWAE.}
        \label{fig:gw-estmin-min}
    \end{minipage}
    \hspace*{\fill}
    \caption{
        The estimation and minimization of the GW metric. This trial of training is conduced in GWAE~(NP, $\coeffd$=1, $\coeffw$=1, $\coeffh$=1) using the MNIST~\cite{MNIST} dataset.
            (a) The curves show the GW values estimated by the loss term~$\lossgw$ (solid, blue) and the empirical GW computed by the POT package~\cite{Flamary2021POT} (dashed, orange).
            The values are computed using the validation set.
            (b) The axes~$\Delta x = \distx(\x,\x')$~(vertical) and $\Delta z = \distz(\z, \z')$~(horizontal) respectively denote the difference in the data and latent spaces between generated samples~$(\x,\z), (\x',\z') \sim \gjoint{\x}{\z}$.
            The histogram contains 10,000 generated sample pairs.
    }
    \label{fig:gw-estmin}
\end{figure}

We validated the estimation and minimization of the GW metric in \cref{fig:gw-estmin}.
First, to validate the estimation of the GW metric, we compared the GW metric estimated in GWAE and the empirical GW value computed in the conventional method in \cref{fig:gw-estmin-est}.
Against the GWAE models estimating the GW metric as in \cref{eq:loss-gw}, the empirical GW values are computed by the standard OT framework POT~\cite{Flamary2021POT}.
Although the estimated~$\lossgw$ is slightly higher than the empirical values, the curves behave in a very similar manner during the entire training process.
This result supports that the GWAE model successfully estimated the GW values and yielded their gradients to proceed with the distribution matching between the data and latent spaces.
Second, to validate the minimization of the GW metric, we show the histogram of the differences of generated samples in the data and latent space in \cref{fig:gw-estmin-min}.
The isometry of generated samples is attained if the generative coupling~$\gjoint{\x}{\z}$ attains the infimum in \cref{eq:gwmetric}.
This histogram result shows that the generative model~$\gjoint{\x}{\z}$ acquired nearly-isometric latent embedding, and suggests that the GW metric was successfully minimized although the objective of \cref{eq:total-loss} contains three regularization loss terms~(refer to \cref{sec:ablation-study} for ablation studies, and \cref{sec:isometry-comparison} for comparisons).
These two experimental results support that the GWAE models successfully estimated and optimized the GW objective.

\subsection{Learning Representations Based on Meta-Priors}
\vspace{\finalvspace}

\newcommand{\disentanglementwidth}{0.3}
\newcommand{\disentanglementinnerwidth}{1.0}
\begin{figure}[!t]
    \centering
    \hspace*{\fill}
    \begin{minipage}[t]{\disentanglementwidth\linewidth}
        \centering
        \includegraphics[width=\disentanglementinnerwidth\linewidth,clip]{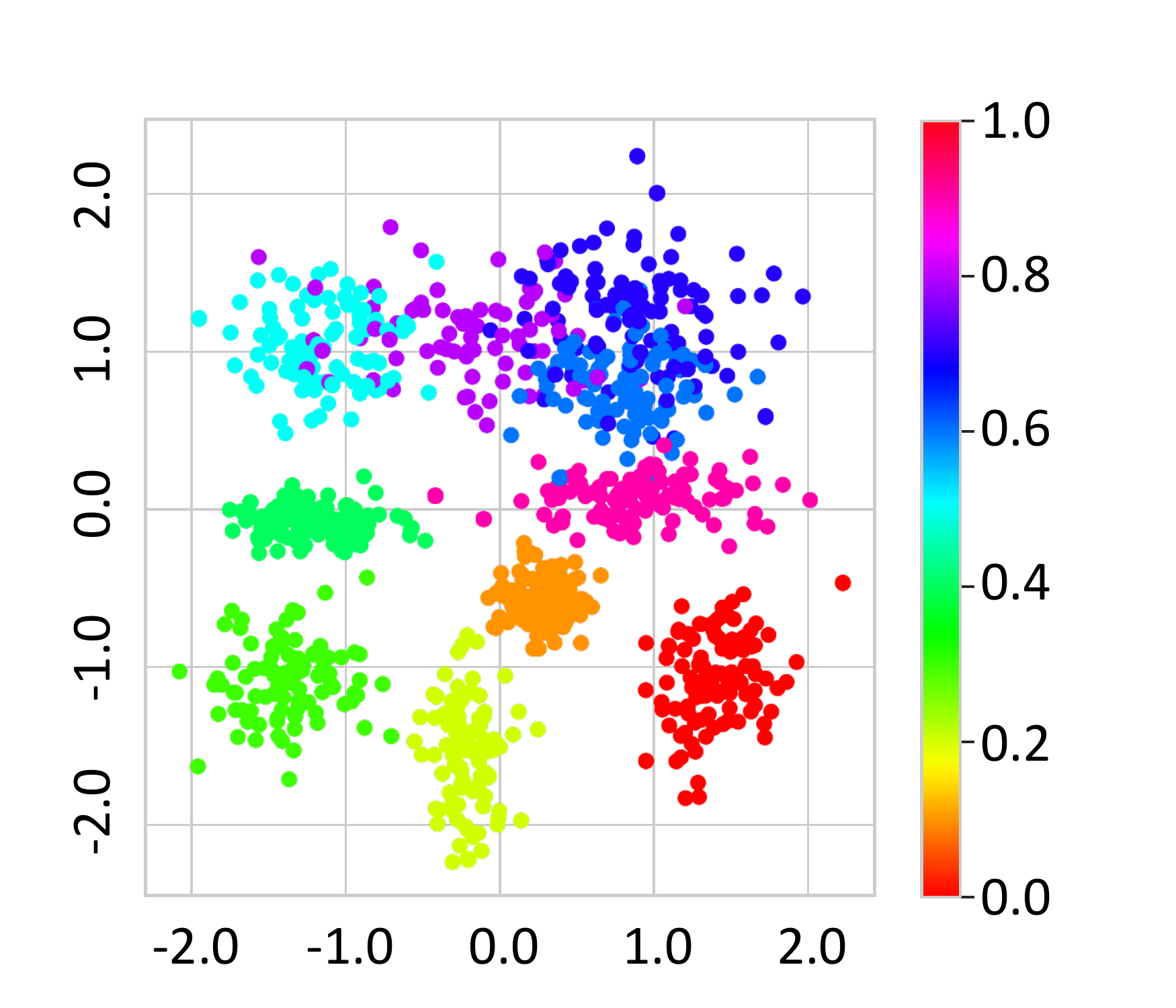}
        \subcaption{VAE~\cite{VAE}.}
    \end{minipage}
    \hfill
    \begin{minipage}[t]{\disentanglementwidth\linewidth}
        \centering
        \includegraphics[width=\disentanglementinnerwidth\linewidth,clip]{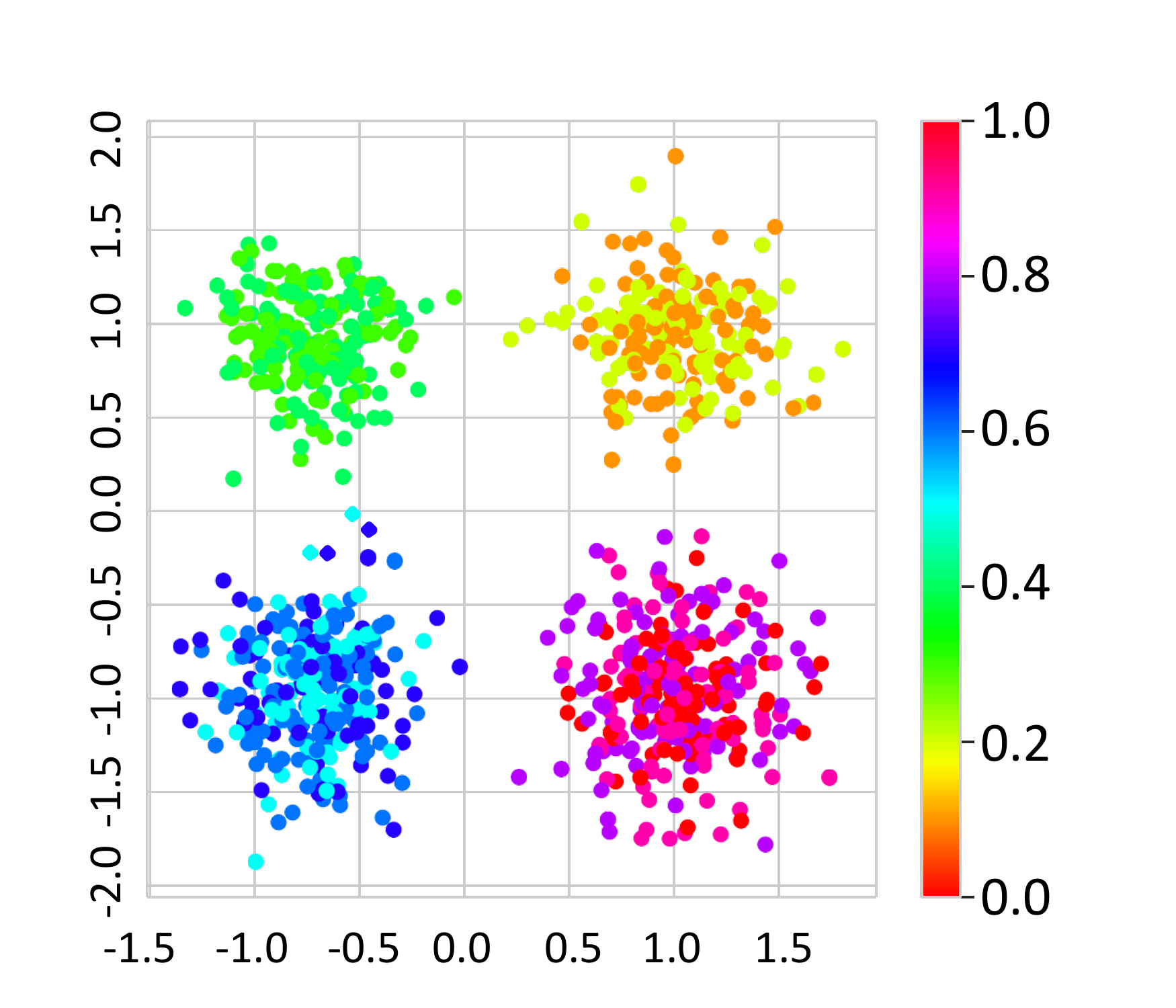}
        \subcaption{FactorVAE~\cite{FactorVAE} ($\gamma$=10).}
    \end{minipage}
    \hfill
    \begin{minipage}[t]{\disentanglementwidth\linewidth}
        \centering
        \includegraphics[width=\disentanglementinnerwidth\linewidth,clip]{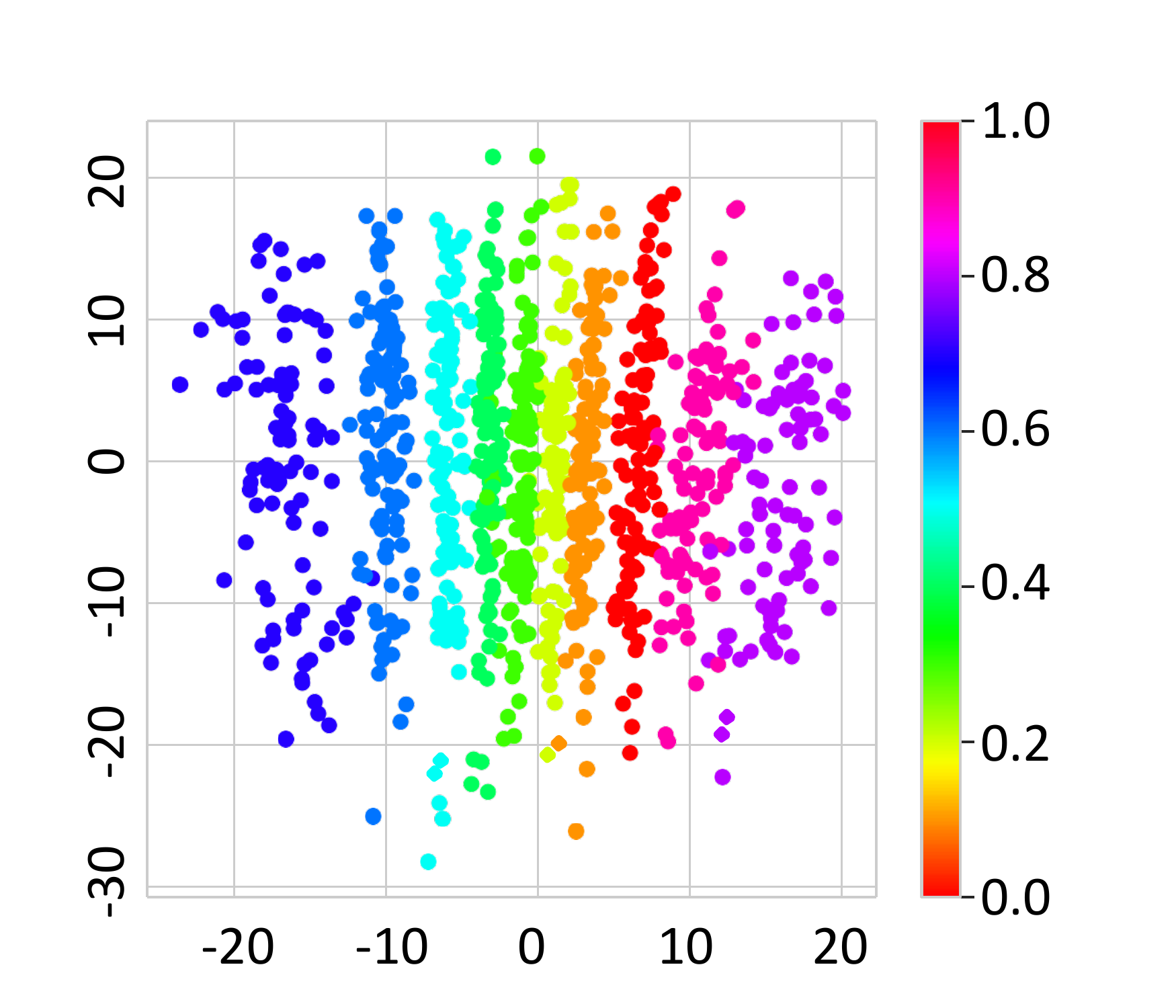}
        \subcaption{GWAE (FNP, $\coeffw$=1, $\coeffd$=10, $\coeffh$=0.3).}
    \end{minipage}
    \hspace*{\fill}
    \caption{
        Comparison of the learned latent spaces in 3D Shapes~\cite{3DShapes} and $L=16$.
        The vertical and horizontal axes in the scatter plots respectively represent two of the 16~($=L$) latent variables with the highest and the second-highest informativeness~\cite{Do2020} \wrt the \textit{object hue} factor.
        Note that a single factor value varies along only one axis in a disentangled representation.
    }
    \label{fig:scatter}
\end{figure}

\begin{table}[!t]
    \centering
    \caption{
        Quantitative comparison of disentanglement.
        The reported scores were calculated in 3D Shapes~\cite{3DShapes}, and the latent size~$L=16$.
        Since the latent size $L$ is larger than the number of the ground truth factors, the hyperparameter tuning was based on the validation set DCI-C~\cite{DCIScore} values.
        To deal with the probabilistic scores~\cite{Zaidi2021}, we reported the ranges for five measurements.
        The details of the scores are provided in \cref{sec:appendix-metrics}.
        \label{tab:disentanglement}
    }
    \small
    \begin{tabular}{rccc}
        \toprule
        Model & 
        DCI-C $\uparrow$ &
        DCI-D $\uparrow$ &
        DCI-I $\uparrow$ \\
        \midrule
        VAE~\cite{VAE}
            & 0.7734 $\pm$ 0.0004
            & 0.6831 $\pm$ 0.0002
            & 0.9914 $\pm$ 0.0003 \\
        $\beta$-VAE~\cite{BetaVAE}
            & 0.8245 $\pm$ 0.0002
            & 0.7328 $\pm$ 0.0002
            & 0.9796 $\pm$ 0.0002 \\
        WAE~\cite{WAE}
            & 0.8288 $\pm$ 0.0004
            & \textbf{0.7544 $\pm$ 0.0004}
            & 0.9959 $\pm$ 0.0001 \\
        $\beta$-TCVAE~\cite{BetaTCVAE}
            & 0.8347 $\pm$ 0.0003
            & 0.7085 $\pm$ 0.0002
            & 0.9880 $\pm$ 0.0002 \\
        FactorVAE~\cite{FactorVAE}
            & 0.7963 $\pm$ 0.0004
            & 0.7390 $\pm$ 0.0004
            & 0.9961 $\pm$ 0.0002 \\
        DIP-VAE-I~\cite{DIP-VAE}
            & 0.8609 $\pm$ 0.0003
            & 0.6984 $\pm$ 0.0003
            & 0.9961 $\pm$ 0.0001 \\
        DIP-VAE-II~\cite{DIP-VAE}
            & 0.8236 $\pm$ 0.0001
            & 0.7498 $\pm$ 0.0003
            & 0.9957 $\pm$ 0.0002 \\
        \midrule
        GWAE~(FNP)
            & \textbf{0.9080 $\pm$ 0.0002}
            & 0.7024 $\pm$ 0.0002
            & \textbf{0.9966 $\pm$ 0.0002} \\
        \bottomrule
        \multicolumn{4}{l}{* The ranges are denoted by $(\text{mean}) \pm (\text{standard error of the mean})$.}
    \end{tabular}
\end{table}

\textbf{Disentanglement.}
We investigated the disentanglement of representations obtained using GWAE models and compared them with conventional VAE-based disentanglement methods.
Since the element-wise independence in the latent space is postulated as a meta-prior for disentangled representation learning, we used the FNP class for the prior~$\trainablePrior{\z}$.
Considering practical applications with unknown ground-truth factor, we set relatively large latent size~$L$ to avoid the shortage of dimensionality.
The qualitative and quantitative results are shown in \cref{fig:scatter} and \cref{tab:disentanglement}, respectively.
These results support the ability to learn a disentangled representation in complex data.
The scatter plots in \cref{fig:scatter} suggest that the GWAE model successfully extracted one underlying factor of variation~(object hue) precisely along one axis, whereas the standard VAE~\cite{VAE} formed several clusters for each value, and FactorVAE~\cite{FactorVAE} obtained the factor in quadrants.

\newcommand{\oodwidth}{0.3}
\newcommand{\oodinnerwidth}{1.0}
\begin{figure}[!t]
    \centering
    \hspace*{\fill}
    \begin{minipage}[t]{\oodwidth\linewidth}
        \centering
        \includegraphics[width=\oodinnerwidth\linewidth,clip]{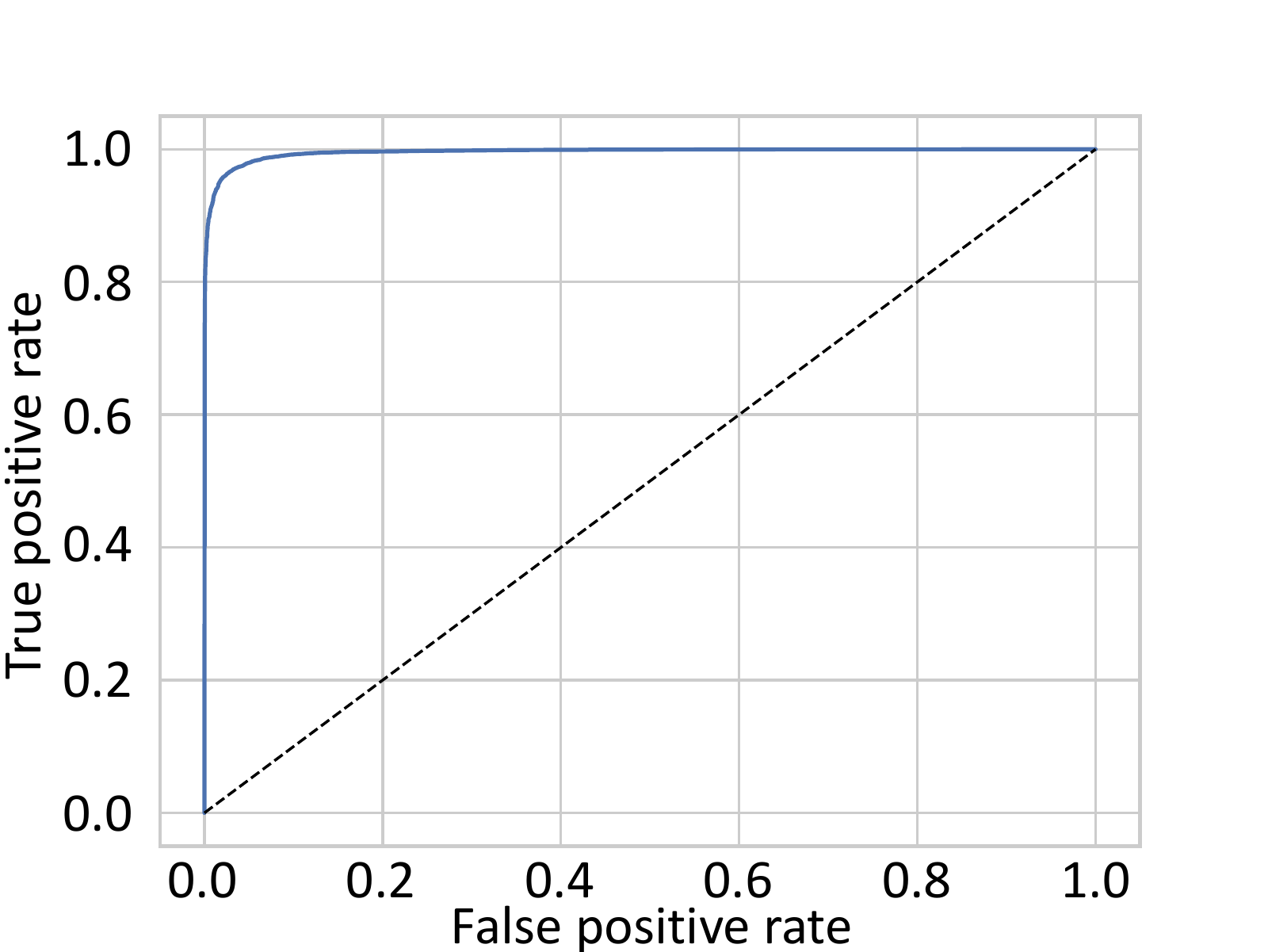}
        \subcaption{VAE~\cite{VAE}, AUC=0.9957.}
    \end{minipage}
    \hfill
    \begin{minipage}[t]{\oodwidth\linewidth}
        \centering
        \includegraphics[width=\oodinnerwidth\linewidth,clip]{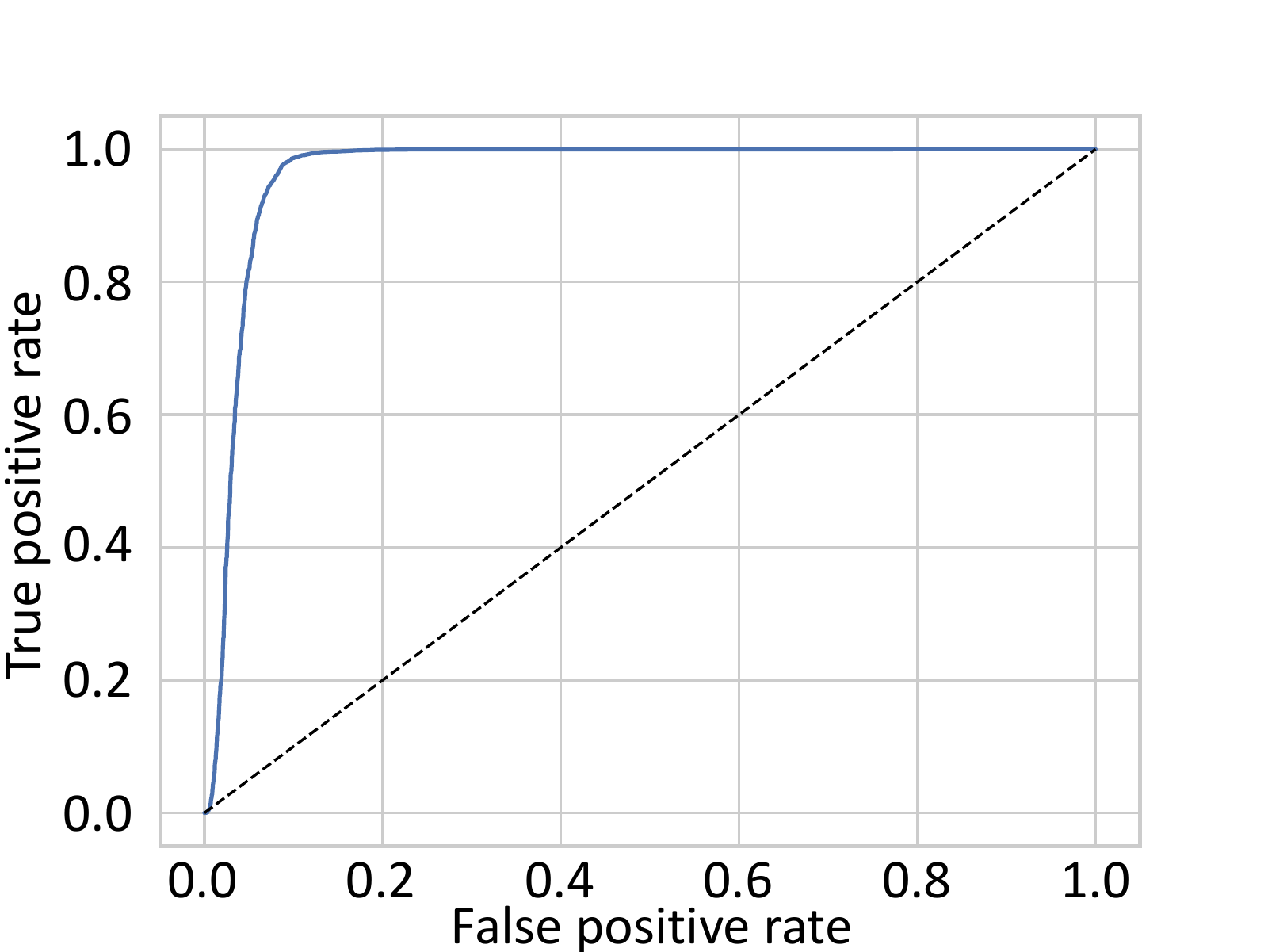}
        \subcaption{DAGMM~\cite{DAGMM}, AUC=0.9654.}
    \end{minipage}
    \hfill
    \begin{minipage}[t]{\oodwidth\linewidth}
        \centering
        \includegraphics[width=\oodinnerwidth\linewidth,clip]{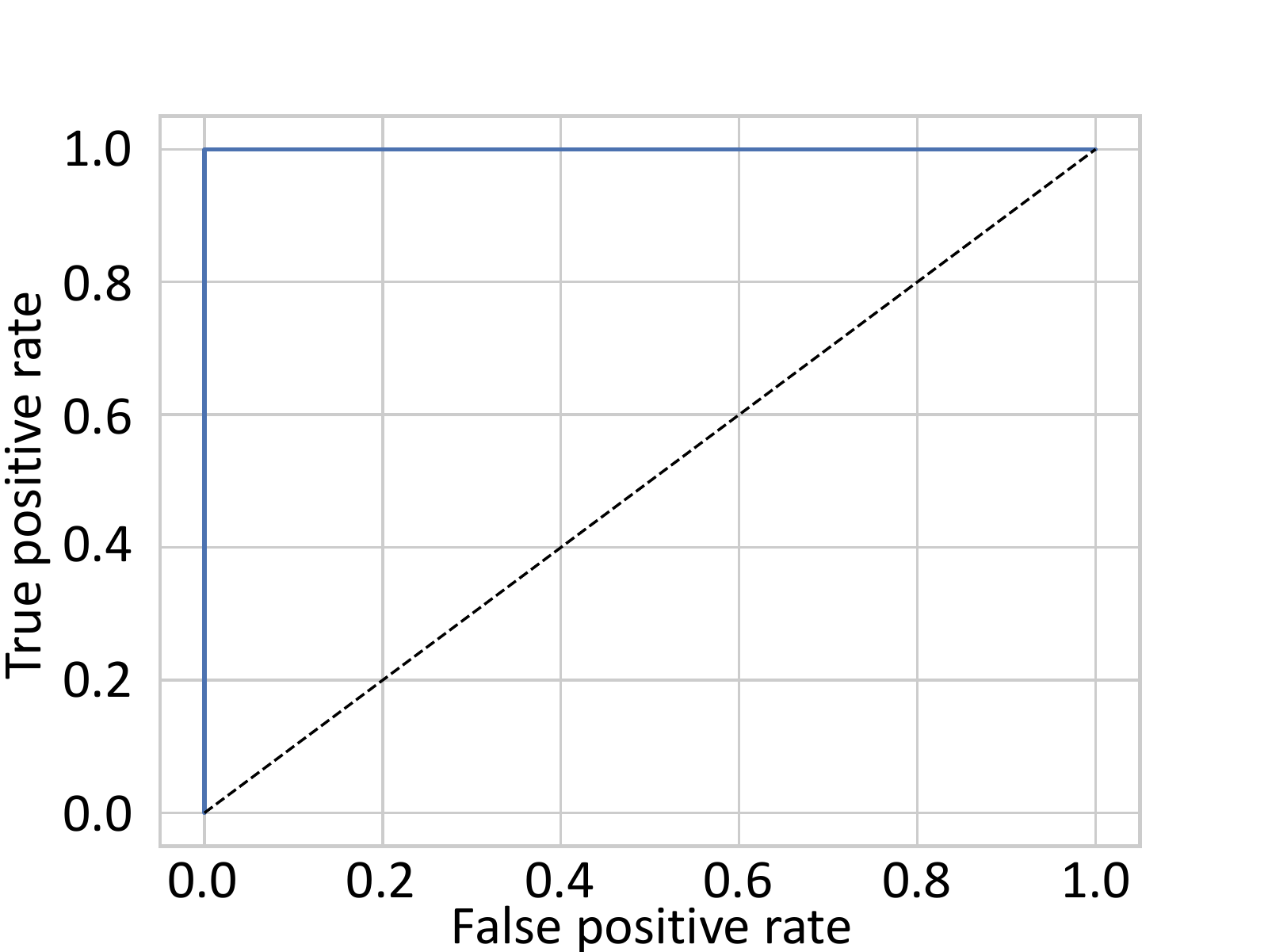}
        \subcaption{GWAE~(GMP), AUC=1.0000.}
    \end{minipage}
    \hspace*{\fill}
    \caption{
        The ROC curves of the OoD detection in MNIST~\cite{MNIST} against Omniglot~\cite{Omniglot}.
        We trained these models using MNIST as ID samples and used Omniglot as OoD samples.
        We upsampled Omniglot to 10,000 samples for data balancing.
        For the anomaly detection using the latent codes~$\z$, we applied the negative log-likelihood energy~$-\log\prior{\z}$ for VAE and DAGMM, and used the estimated Kantorovich potential~$\ev{\dec{\tilde{\x}}{\z}} \left[\critic(\tilde{\x},\z)\right]$ for GWAE (see \cref{sec:appendix-ood-prior} for more latent space details).
    }
    \label{fig:ood}
\end{figure}

\textbf{Clustering Structure.}
We empirically evaluated the capabilities of capturing clusters using MNIST~\cite{MNIST}.
We compared the GWAE model using GMP with other VAE-based methods considering the out-of-distribution~(OoD) detection performance in \cref{fig:ood}.
We used MNIST images as in-distribution~(ID) samples for training and Omniglot~\cite{Omniglot} images as unseen OoD samples.
Quantitative results show that the GWAE model successfully extracted the clustering structure, empirically implying the applicability of multimodal priors.

\subsection{Autoencoding Model}
\vspace{\finalvspace}

We additionally studied the autoencoding and generation performance of GWAE models in \cref{tab:fid} (see \cref{sec:appendix-qualitative} for qualitative evaluations).
Although the distribution matching~$\gmodel{\x} \approx \empir{\x}$ is a collateral condition of \cref{eq:loss-gw}, quantitative results show that the GWAE model also favorably compares with existing autoencoding models in terms of generative capacity.
This result suggests the substantial capture of the underlying low-dimensional distribution in GWAE models, which can lead to the applications to other types of meta-priors.

\definecolor{tablesubcolor}{HTML}{DDDDDD}
\newcommand{\tablesubcolor}{tablesubcolor}
\newcommand{\cctsc}{\cellcolor{\tablesubcolor}}
\begin{table}[!t]
    \centering
    \caption{
        Quantitative comparisons of generation and reconstruction.
        The FID scores~\cite{FID} evaluate a random sample set from the generative model~$\gmodel{\x}$ (without using dataset images) against the entire test set, and both consist of an equal number of 19,962 samples.
        The PSNR scores measure the reconstruction~$\iagg{\z}\dec{\x}{\z}$ using test images (see \cref{sec:appendix-metrics} for details).
        All reported values were computed in CelebA~\cite{CelebA} with a latent size of~$L=64$.
        For all the methods, we applied early stopping~(patience=10) and hyperparameter tuning using the validation set.
        The bold and underlined values respectively denote \textbf{the best} and \underline{the second-best} performance in each score.
        \label{tab:fid}
    }
    \small
    \begin{tabular}{rlcc}
        \toprule
        \multicolumn{2}{l}{Model} & FID $\downarrow$ & PSNR [dB] $\uparrow$ \\
        \midrule
                Baseline & VAE~\cite{VAE} & 130.9 & 19.96 \\
                \cctsc & \cctsc $\beta$-VAE~\cite{BetaVAE} & \cctsc 92.6 & \cctsc \underline{22.71} \\
                \cctsc & \cctsc GECO~\cite{GECO} & \cctsc 162.1 & \cctsc 21.19 \\
            \multirow{-3}{*}{\cctsc KL re-weighting}
                & \cctsc $\sigma$-VAE~\cite{SigmaVAE} & \cctsc 53.13${}^*$ & \cctsc 20.03 \\
            \multirow{2}{*}{Hierarchical factors}
                & LadderVAE~\cite{LadderVAE} & 255.6 & 12.35 \\
                & VLadderAE~\cite{VLadderAE} & 147.1 & 19.76 \\
                \cctsc & \cctsc WAE~\cite{WAE} & \cctsc 55${}^*$ & \cctsc 22.70 \\
                \cctsc & \cctsc WVI~\cite{WVI} & \cctsc 295.0 & \cctsc 14.45 \\
                \cctsc & \cctsc SWAE~\cite{SWAE} & \cctsc 102.2 & \cctsc 21.85 \\
            \multirow{-4}{*}{\cctsc OT-based models}
                 & \cctsc RAE~\cite{RAE} & \cctsc 52.20${}^*$ & \cctsc 21.34 \\
            \multirow{2}{*}{Trainable priors}
                & VampPrior~\cite{VampPrior} & 243.8 & 16.23 \\
                & 2-Stage VAE~\cite{Dai2019} & \textbf{34${}^*$} & 16.15 \\
                \cctsc & \cctsc VAE-GAN~\cite{VAEGAN} & \cctsc 111.8 & \cctsc 19.51 \\
                \cctsc & \cctsc AVB~\cite{AVB} & \cctsc 93.0 & \cctsc 22.60 \\
            \multirow{-3}{*}{\cctsc IVI-based models}
                & \cctsc ALI~\cite{ALI} & \cctsc 171.8 & \cctsc 12.26 \\
        \midrule
            \multirow{1}{*}{Ours}
                & GWAE (NP) & \underline{45.3} & \textbf{22.82} \\
        \bottomrule
        \multicolumn{4}{l}{{\footnotesize * The values are cited from the original papers annotated after the model names.}}
    \end{tabular}
\end{table}


\section{Conclusion} \label{sec:conclusion}
In this work, we have introduced a novel representation learning method that performs the distance distribution matching between the given unlabeled data and the latent space.
Our GWAE model family transfers distance structure from the data space into the latent space in the OT viewpoint, replacing the ELBO objective of variational inference with the GW metric.
The GW objective provides a direct measure between the latent and data distribution.
Qualitative and quantitative evaluations empirically show the performance of GWAE models in terms of representation learning.
In future work, further applications also remain open to various types of meta-priors, such as spherical representations and non-Euclidean embedding spaces.

\newpage

\section*{Reproducibility Statement}
We describe the implementation details in \cref{sec:experiments}, \cref{sec:appendix-method}, and \cref{sec:appendix-experiments}.
The dataset details are provided in \cref{sec:appendix-datasets}.
To ensure reproducibility, our code is available online at~\url{https://github.com/ganmodokix/gwae} and is provided as the supplementary material.

\section*{Acknowledgments}
This work was partly supported by AMED Grant Number JP21zf0127004 and JSPS KAKENHI Grant Number JP21H03456.

\bibliographystyle{iclr2023_conference}
\bibliography{ref}

\begin{thebibliography}{82}
\providecommand{\natexlab}[1]{#1}
\providecommand{\url}[1]{\texttt{#1}}
\expandafter\ifx\csname urlstyle\endcsname\relax
  \providecommand{\doi}[1]{doi: #1}\else
  \providecommand{\doi}{doi: \begingroup \urlstyle{rm}\Url}\fi

\bibitem[Achille \& Soatto(2018)Achille and Soatto]{Achille2018Bottleneck}
Alessandro Achille and Stefano Soatto.
\newblock Information dropout: Learning optimal representations through noisy
  computation.
\newblock \emph{{IEEE Transactions on Pattern Analysis \& Machine
  Intelligence}}, 40\penalty0 (12):\penalty0 2897--2905, 2018.
\newblock \doi{10.1109/TPAMI.2017.2784440}.

\bibitem[Alemi et~al.(2018)Alemi, Fischer, Dillon, and Murphy]{VIB}
Alexander~A. Alemi, Ian Fischer, Joshua~V. Dillon, and Kevin Murphy.
\newblock Deep variational information bottleneck.
\newblock In \emph{Proceedings of the International Conference on Learning
  Representations (ICLR)}, pp.\  1--19, 2018.
\newblock URL \url{https://openreview.net/forum?id=HyxQzBceg}.

\bibitem[Ambrogioni et~al.(2018)Ambrogioni, G\"{u}\c{c}l\"{u},
  G\"{u}\c{c}l\"{u}t\"{u}rk, Hinne, van Gerven, and Maris]{WVI}
Luca Ambrogioni, Umut G\"{u}\c{c}l\"{u}, Ya\u{g}mur G\"{u}\c{c}l\"{u}t\"{u}rk,
  Max Hinne, Marcel A.~J. van Gerven, and Eric Maris.
\newblock Wasserstein variational inference.
\newblock In \emph{Proceedings of Neural Information Processing Systems
  (NIPS)}, pp.\  2473--2482, 2018.
\newblock URL
  \url{https://papers.nips.cc/paper/2018/hash/2c89109d42178de8a367c0228f169bf8-Abstract.html}.

\bibitem[Aneja et~al.(2021)Aneja, Schwing, Kautz, and Vahdat]{NCPVAE}
Jyoti Aneja, Alex Schwing, Jan Kautz, and Arash Vahdat.
\newblock A contrastive learning approach for training variational autoencoder
  priors.
\newblock In \emph{Proceedings of Neural Information Processing Systems
  (NeurIPS)}, pp.\  480--493, 2021.
\newblock URL
  \url{https://proceedings.neurips.cc/paper/2021/hash/0496604c1d80f66fbeb963c12e570a26-Abstract.html}.

\bibitem[Arjovsky \& Bottou(2017)Arjovsky and Bottou]{Arjovsky2017}
Mart{\'{\i}}n Arjovsky and L{\'{e}}on Bottou.
\newblock Towards principled methods for training generative adversarial
  networks.
\newblock In \emph{Proceedings of the International Conference on Learning
  Representations (ICLR)}, pp.\  1--17, 2017.
\newblock URL \url{https://openreview.net/forum?id=Hk4_qw5xe}.

\bibitem[Arjovsky et~al.(2017)Arjovsky, Chintala, and Bottou]{WGAN}
Martin Arjovsky, Soumith Chintala, and L{\'e}on Bottou.
\newblock Wasserstein generative adversarial networks.
\newblock In \emph{Proceedings of the International Conference on Machine
  Learning (ICML)}, pp.\  214--223, 2017.
\newblock URL \url{https://proceedings.mlr.press/v70/arjovsky17a.html}.

\bibitem[Asano et~al.(2020)Asano, Rupprecht, and Vedaldi]{Asano2020}
Yuki~M. Asano, Christian Rupprecht, and Andrea Vedaldi.
\newblock Self-labelling via simultaneous clustering and representation
  learning.
\newblock In \emph{Proceedings of the International Conference on Learning
  Representations (ICLR)}, 2020.
\newblock URL \url{https://openreview.net/forum?id=Hyx-jyBFPr}.

\bibitem[Bengio et~al.(2013)Bengio, Courville, and Vincent]{Bengio2013}
Yoshua Bengio, Aaron Courville, and Pascal Vincent.
\newblock Representation learning: A review and new perspectives.
\newblock \emph{IEEE Transactions on Pattern Analysis and Machine
  Intelligence}, 35\penalty0 (8):\penalty0 1798--1828, 2013.
\newblock \doi{10.1109/TPAMI.2013.50}.

\bibitem[Breiman(2001)]{RandomForest}
Leo Breiman.
\newblock Random forests.
\newblock \emph{Machine Learning}, 45\penalty0 (1):\penalty0 5--32, 2001.
\newblock \doi{10.1023/A:1010933404324}.

\bibitem[Burgess \& Kim(2018)Burgess and Kim]{3DShapes}
Chris Burgess and Hyunjik Kim.
\newblock {3D Shapes Dataset}.
\newblock \url{https://github.com/deepmind/3d-shapes/}, 2018.
\newblock Accessed May 13, 2022.

\bibitem[Carlsson et~al.(2008)Carlsson, Ishkhanov, de~Silva, and
  Zomorodian]{Carlson2008}
Gunnar Carlsson, Tigran Ishkhanov, Vin de~Silva, and Afra Zomorodian.
\newblock On the local behavior of spaces of natural images.
\newblock \emph{International Journal of Computer Vision (IJCV)}, 76\penalty0
  (1):\penalty0 1--12, 2008.
\newblock \doi{10.1007/s11263-007-0056-x}.

\bibitem[Chen et~al.(2018)Chen, Li, Grosse, and Duvenaud]{BetaTCVAE}
Ricky T.~Q. Chen, Xuechen Li, Roger Grosse, and David Duvenaud.
\newblock Isolating sources of disentanglement in variational autoencoders.
\newblock In \emph{Proceedings of Neural Information Processing Systems
  (NIPS)}, pp.\  2610--2620, 2018.
\newblock URL
  \url{https://proceedings.neurips.cc/paper/2018/hash/1ee3dfcd8a0645a25a35977997223d22-Abstract.html}.

\bibitem[Chu et~al.(2020)Chu, Minami, and Fukumizu]{Chu2020}
Casey Chu, Kentaro Minami, and Kenji Fukumizu.
\newblock Smoothness and stability in {GANs}.
\newblock In \emph{Proceedings of the International Conference on Learning
  Representations (ICLR)}, pp.\  1--15, 2020.
\newblock URL \url{https://openreview.net/forum?id=HJeOekHKwr}.

\bibitem[Creager et~al.(2019)Creager, Madras, Jacobsen, Weis, Swersky, Pitassi,
  and Zemel]{Creager2019}
Elliot Creager, David Madras, Joern-Henrik Jacobsen, Marissa Weis, Kevin
  Swersky, Toniann Pitassi, and Richard Zemel.
\newblock Flexibly fair representation learning by disentanglement.
\newblock In \emph{Proceedings of the International Conference on Machine
  Learning (ICML)}, pp.\  1436--1445, 2019.
\newblock URL \url{https://proceedings.mlr.press/v97/creager19a.html}.

\bibitem[Dai \& Wipf(2019)Dai and Wipf]{Dai2019}
Bin Dai and David Wipf.
\newblock Diagnosing and enhancing vae models.
\newblock In \emph{Proceedings of the International Conference on Learning
  Representations (ICLR)}, pp.\  1--12, 2019.
\newblock URL \url{https://openreview.net/forum?id=B1e0X3C9tQ}.

\bibitem[Dai et~al.(2018)Dai, Wang, Aston, Hua, and Wipf]{Dai2018}
Bin Dai, Yu~Wang, John Aston, Gang Hua, and David Wipf.
\newblock Connections with robust {PCA} and the role of emergent sparsity in
  variational autoencoder models.
\newblock \emph{Journal of Machine Learning Research}, 19\penalty0
  (41):\penalty0 1--42, 2018.
\newblock URL \url{http://jmlr.org/papers/v19/17-704.html}.

\bibitem[Dai et~al.(2020)Dai, Wang, and Wipf]{Dai2020}
Bin Dai, Ziyu Wang, and David Wipf.
\newblock The usual suspects? {R}eassessing blame for {VAE} posterior collapse.
\newblock In \emph{Proceedings of the International Conference on Machine
  Learning (ICML)}, pp.\  2313--2322, 2020.
\newblock URL \url{https://proceedings.mlr.press/v119/dai20c.html}.

\bibitem[Deng et~al.(2009)Deng, Dong, Socher, Li, Li, and Fei-Fei]{ImageNet}
Jia Deng, Wei Dong, Richard Socher, Li-Jia Li, Kai Li, and Li~Fei-Fei.
\newblock {ImageNet}: A large-scale hierarchical image database.
\newblock In \emph{Proceedings of the IEEE Conference on Computer Vision and
  Pattern Recognition (CVPR)}, pp.\  248--255, 2009.
\newblock \doi{10.1109/CVPR.2009.5206848}.

\bibitem[Detlefsen \& Hauberg(2019)Detlefsen and Hauberg]{Detlefsen2019}
Nicki~Skafte Detlefsen and S{\o}ren Hauberg.
\newblock Explicit disentanglement of appearance and perspective in generative
  models.
\newblock In \emph{Proceedings of Neural Information Processing Systems
  (NeurIPS)}, pp.\  1018--1028, 2019.
\newblock URL
  \url{https://proceedings.neurips.cc/paper/2019/hash/3493894fa4ea036cfc6433c3e2ee63b0-Abstract.html}.

\bibitem[Ding et~al.(2020)Ding, Xu, Xu, Parmar, Yang, Welling, and
  Tu]{GuidedVAE}
Zheng Ding, Yifan Xu, Weijian Xu, Gaurav Parmar, Yang Yang, Max Welling, and
  Zhuowen Tu.
\newblock Guided variational autoencoder for disentanglement learning.
\newblock In \emph{Proceedings of the IEEE/CVF Conference on Computer Vision
  and Pattern Recognition (CVPR)}, pp.\  7920--7929, 2020.
\newblock URL
  \url{https://openaccess.thecvf.com/content_CVPR_2020/html/Ding_Guided_Variational_Autoencoder_for_Disentanglement_Learning_CVPR_2020_paper.html}.

\bibitem[Do \& Tran(2020)Do and Tran]{Do2020}
Kien Do and Truyen Tran.
\newblock Theory and evaluation metrics for learning disentangled
  representations.
\newblock In \emph{Proceedings of the International Conference on Learning
  Representations (ICLR)}, pp.\  1--30, 2020.
\newblock URL \url{https://openreview.net/forum?id=HJgK0h4Ywr}.

\bibitem[Donahue et~al.(2017)Donahue, Kr{\"{a}}henb{\"{u}}hl, and
  Darrell]{BiGAN}
Jeff Donahue, Philipp Kr{\"{a}}henb{\"{u}}hl, and Trevor Darrell.
\newblock Adversarial feature learning.
\newblock In \emph{Proceedings of the International Conference on Learning
  Representations (ICLR)}, pp.\  1--18, 2017.
\newblock URL \url{https://openreview.net/forum?id=BJtNZAFgg}.

\bibitem[Dumoulin et~al.(2017)Dumoulin, Belghazi, Poole, Mastropietro, Lamb,
  Arjovsky, and Courville]{ALI}
Vincent Dumoulin, Ishmael Belghazi, Ben Poole, Olivier Mastropietro, Alex Lamb,
  Martin Arjovsky, and Aaron Courville.
\newblock Adversarially learned inference.
\newblock In \emph{Proceedings of the International Conference on Learning
  Representations (ICLR)}, pp.\  1--18, 2017.
\newblock URL \url{https://openreview.net/forum?id=B1ElR4cgg}.

\bibitem[Eastwood \& Williams(2018)Eastwood and Williams]{DCIScore}
Cian Eastwood and Christopher K.~I. Williams.
\newblock A framework for the quantitative evaluation of disentangled
  representations.
\newblock In \emph{Proceedings of the International Conference on Learning
  Representations (ICLR)}, pp.\  1--15, 2018.
\newblock URL \url{https://openreview.net/forum?id=By-7dz-AZ}.

\bibitem[Flamary et~al.(2021)Flamary, Courty, Gramfort, Alaya, Boisbunon,
  Chambon, Chapel, Corenflos, Fatras, Fournier, Gautheron, Gayraud, Janati,
  Rakotomamonjy, Redko, Rolet, Schutz, Seguy, Sutherland, Tavenard, Tong, and
  Vayer]{Flamary2021POT}
R{\'e}mi Flamary, Nicolas Courty, Alexandre Gramfort, Mokhtar~Z. Alaya,
  Aur{\'e}lie Boisbunon, Stanislas Chambon, Laetitia Chapel, Adrien Corenflos,
  Kilian Fatras, Nemo Fournier, L{\'e}o Gautheron, Nathalie~T.H. Gayraud,
  Hicham Janati, Alain Rakotomamonjy, Ievgen Redko, Antoine Rolet, Antony
  Schutz, Vivien Seguy, Danica~J. Sutherland, Romain Tavenard, Alexander Tong,
  and Titouan Vayer.
\newblock {POT}: Python optimal transport.
\newblock \emph{Journal of Machine Learning Research}, 22\penalty0
  (78):\penalty0 1--8, 2021.
\newblock URL \url{http://jmlr.org/papers/v22/20-451.html}.

\bibitem[Gaujac et~al.(2021)Gaujac, Feige, and Barber]{TCWAE}
Benoit Gaujac, Ilya Feige, and David Barber.
\newblock Learning disentangled representations with the wasserstein
  autoencoder.
\newblock In \emph{Proceedings of the European Conference on Machine Learning
  and Principles and Practice of Knowledge Discovery in Databases (ECML PKDD),
  Part III}, pp.\  69--84, 2021.
\newblock \doi{10.1007/978-3-030-86523-8_5}.

\bibitem[Goodfellow et~al.(2014)Goodfellow, Pouget-Abadie, Mirza, Xu,
  Warde-Farley, Ozair, Courville, and Bengio]{GAN}
Ian Goodfellow, Jean Pouget-Abadie, Mehdi Mirza, Bing Xu, David Warde-Farley,
  Sherjil Ozair, Aaron Courville, and Yoshua Bengio.
\newblock Generative adversarial nets.
\newblock In \emph{Proceedings of Neural Information Processing Systems
  (NIPS)}, pp.\  2672--2680, 2014.
\newblock URL
  \url{https://papers.nips.cc/paper/2014/hash/5ca3e9b122f61f8f06494c97b1afccf3-Abstract.html}.

\bibitem[Gulrajani et~al.(2017)Gulrajani, Ahmed, Arjovsky, Dumoulin, and
  Courville]{WGANGP}
Ishaan Gulrajani, Faruk Ahmed, Martin Arjovsky, Vincent Dumoulin, and Aaron
  Courville.
\newblock Improved training of wasserstein gans.
\newblock In \emph{Proceedings of Neural Information Processing Systems
  (NIPS)}, pp.\  5769–5779, 2017.
\newblock URL
  \url{https://papers.nips.cc/paper/2017/hash/892c3b1c6dccd52936e27cbd0ff683d6-Abstract.html}.

\bibitem[Hendrycks \& Gimpel(2016)Hendrycks and Gimpel]{GELU}
Dan Hendrycks and Kevin Gimpel.
\newblock Gaussian error linear units (gelus).
\newblock {arXiv: 1606.08415}, 2016.

\bibitem[Heusel et~al.(2017)Heusel, Ramsauer, Unterthiner, Nessler, and
  Hochreiter]{FID}
Martin Heusel, Hubert Ramsauer, Thomas Unterthiner, Bernhard Nessler, and Sepp
  Hochreiter.
\newblock Gans trained by a two time-scale update rule converge to a local nash
  equilibrium.
\newblock In \emph{Proceedings of Neural Information Processing Systems
  (NIPS)}, pp.\  6629--6640, 2017.
\newblock URL
  \url{https://papers.nips.cc/paper/2017/hash/8a1d694707eb0fefe65871369074926d-Abstract.html}.

\bibitem[Higgins et~al.(2017{\natexlab{a}})Higgins, Matthey, Pal, Burgess,
  Glorot, Botvinick, Mohamed, and Lerchner]{BetaVAE}
Irina Higgins, Loic Matthey, Arka Pal, Christopher Burgess, Xavier Glorot,
  Matthew Botvinick, Shakir Mohamed, and Alexander Lerchner.
\newblock {$\beta$-VAE}: Learning basic visual concepts with a constrained
  variational framework.
\newblock In \emph{Proceedings of the International Conference on Learning
  Representations (ICLR)}, pp.\  1--22, 2017{\natexlab{a}}.
\newblock URL \url{https://openreview.net/forum?id=Sy2fzU9gl}.

\bibitem[Higgins et~al.(2017{\natexlab{b}})Higgins, Pal, Rusu, Matthey,
  Burgess, Pritzel, Botvinick, Blundell, and Lerchner]{DARLA}
Irina Higgins, Arka Pal, Andrei~A. Rusu, Loic Matthey, Christopher Burgess,
  Alexander Pritzel, Matthew Botvinick, Charles Blundell, and Alexander
  Lerchner.
\newblock {DARLA}: Improving zero-shot transfer in reinforcement learning.
\newblock In \emph{Proceedings of the International Conference on Machine
  Learning (ICML)}, pp.\  1480--1490, 2017{\natexlab{b}}.
\newblock URL \url{http://proceedings.mlr.press/v70/higgins17a.html}.

\bibitem[Hoffman et~al.(2017)Hoffman, Riquelme, and Johnson]{Hoffman2017}
Matt Hoffman, Carlos Riquelme, and Matthew Johnson.
\newblock The beta {VAE}'s implicit prior.
\newblock In \emph{Proceedings of Neural Information Processing Systems (NIPS)
  Workshop on Bayesian Deep Learning}, pp.\  1--5, 2017.
\newblock URL \url{https://research.google/pubs/pub47350/}.

\bibitem[Hou et~al.(2019)Hou, Sun, Shen, and Qiu]{DFCVAE}
Xianxu Hou, Ke~Sun, Linlin Shen, and Guoping Qiu.
\newblock Improving variational autoencoder with deep feature consistent and
  generative adversarial training.
\newblock \emph{Neurocomputing}, 341:\penalty0 183--194, 2019.
\newblock \doi{10.1016/j.neucom.2019.03.013}.

\bibitem[Hsu et~al.(2017)Hsu, Zhang, and Glass]{Hsu2017}
Wei-Ning Hsu, Yu~Zhang, and James Glass.
\newblock Learning latent representations for speech generation and
  transformation.
\newblock In \emph{Proceedings of the Annual Conference of the International
  Speech Communication Association (INTERSPEECH)}, pp.\  1273--1277, 2017.
\newblock \doi{10.21437/Interspeech.2017-349}.

\bibitem[Hu et~al.(2017)Hu, Yang, Liang, Salakhutdinov, and Xing]{Hu2017}
Zhiting Hu, Zichao Yang, Xiaodan Liang, Ruslan Salakhutdinov, and Eric~P. Xing.
\newblock Toward controlled generation of text.
\newblock In \emph{Proceedings of the International Conference on Machine
  Learning (ICML)}, pp.\  1587--1596, 2017.
\newblock URL \url{https://proceedings.mlr.press/v70/hu17e.html}.

\bibitem[Husz{\'a}r(2017)]{Huszar2017}
Ferenc Husz{\'a}r.
\newblock Variational inference using implicit distributions.
\newblock {arXiv: 1702.08235}, 2017.

\bibitem[Kim \& Mnih(2018)Kim and Mnih]{FactorVAE}
Hyunjik Kim and Andriy Mnih.
\newblock Disentangling by factorising.
\newblock In \emph{Proceedings of the International Conference on Machine
  Learning (ICML)}, pp.\  2649--2658, 2018.
\newblock URL \url{http://proceedings.mlr.press/v80/kim18b.html}.

\bibitem[Kingma \& Ba(2015)Kingma and Ba]{Adam}
Diederik~P. Kingma and Jimmy Ba.
\newblock Adam: A method for stochastic optimization.
\newblock In \emph{Proceedings of the International Conference on Learning
  Representations (ICLR)}, pp.\  1--13, 2015.
\newblock URL \url{https://openreview.net/forum?id=8gmWwjFyLj}.

\bibitem[Kingma \& Welling(2014)Kingma and Welling]{VAE}
Diederik~P. Kingma and Max Welling.
\newblock Auto-encoding variational bayes.
\newblock In \emph{Proceedings of the International Conference on Learning
  Representations (ICLR)}, pp.\  1--13, 2014.
\newblock URL \url{https://openreview.net/forum?id=33X9fd2-9FyZd}.

\bibitem[Kolouri et~al.(2019)Kolouri, Pope, Martin, and Rohde]{SWAE}
Soheil Kolouri, Phillip~E. Pope, Charles~E. Martin, and Gustavo~K. Rohde.
\newblock Sliced wasserstein auto-encoders.
\newblock In \emph{Proceedings of International Conference on Learning
  Representations (ICLR)}, 2019.
\newblock URL \url{https://openreview.net/forum?id=H1xaJn05FQ}.

\bibitem[Krizhevsky \& Hinton(2009)Krizhevsky and Hinton]{CIFAR10}
Alex Krizhevsky and Geoffrey Hinton.
\newblock Learning multiple layers of features from tiny images, 2009.
\newblock {Master's thesis, Technical Report, University of Toronto.}

\bibitem[Krizhevsky et~al.(2012)Krizhevsky, Sutskever, and Hinton]{AlexNet}
Alex Krizhevsky, Ilya Sutskever, and Geoffrey~E Hinton.
\newblock Imagenet classification with deep convolutional neural networks.
\newblock In \emph{Proceedings of Neural Information Processing Systems
  (NIPS)}, pp.\  1097--1105, 2012.
\newblock URL
  \url{https://papers.nips.cc/paper/2012/hash/c399862d3b9d6b76c8436e924a68c45b-Abstract.html}.

\bibitem[Kumar et~al.(2018)Kumar, Sattigeri, and Balakrishnan]{DIP-VAE}
Abhishek Kumar, Prasanna Sattigeri, and Avinash Balakrishnan.
\newblock Variational inference of disentangled latent concepts from unlabeled
  observations.
\newblock In \emph{Proceedings of the International Conference on Learning
  Representations (ICLR)}, pp.\  1--16, 2018.
\newblock URL \url{https://openreview.net/forum?id=H1kG7GZAW}.

\bibitem[Lake et~al.(2015)Lake, Salakhutdinov, and Tenenbaum]{Omniglot}
Brenden~M. Lake, Ruslan Salakhutdinov, and Joshua~B. Tenenbaum.
\newblock Human-level concept learning through probabilistic program induction.
\newblock \emph{Science}, 350\penalty0 (6266):\penalty0 1332--1338, 2015.
\newblock \doi{10.1126/science.aab3050}.

\bibitem[Larsen et~al.(2016)Larsen, S{\o}nderby, Larochelle, and
  Winther]{VAEGAN}
Anders Boesen~Lindbo Larsen, Søren~Kaae S{\o}nderby, Hugo Larochelle, and Ole
  Winther.
\newblock Autoencoding beyond pixels using a learned similarity metric.
\newblock In \emph{Proceedings of the International Conference on Machine
  Learning (ICML)}, pp.\  1558--1566, 2016.
\newblock URL \url{http://proceedings.mlr.press/v48/larsen16.html}.

\bibitem[LeCun et~al.(1998)LeCun, Bottou, Bengio, and Haffner]{MNIST}
Yann LeCun, L\'{e}eon Bottou, Yoshua Bengio, and Patrick Haffner.
\newblock Gradient-based learning applied to document recognition.
\newblock \emph{Proceedings of the IEEE}, 86\penalty0 (11):\penalty0
  2278--2324, 1998.
\newblock \doi{10.1109/5.726791}.

\bibitem[Liu et~al.(2015)Liu, Luo, Wang, and Tang]{CelebA}
Ziwei Liu, Ping Luo, Xiaogang Wang, and Xiaoou Tang.
\newblock Deep learning face attributes in the wild.
\newblock In \emph{Proceedings of the IEEE International Conference on Computer
  Vision (ICCV)}, pp.\  3730--3738, 2015.
\newblock \doi{10.1109/ICCV.2015.425}.

\bibitem[Locatello et~al.(2019{\natexlab{a}})Locatello, Abbati, Rainforth,
  Bauer, Sch\"{o}lkopf, and Bachem]{Locatello2019Fairness}
Francesco Locatello, Gabriele Abbati, Thomas Rainforth, Stefan Bauer, Bernhard
  Sch\"{o}lkopf, and Olivier Bachem.
\newblock On the fairness of disentangled representations.
\newblock In \emph{Proceedings of Neural Information Processing Systems
  (NeurIPS)}, pp.\  14584--14597, 2019{\natexlab{a}}.
\newblock URL
  \url{https://proceedings.neurips.cc/paper/2019/hash/1b486d7a5189ebe8d8c46afc64b0d1b4-Abstract.html}.

\bibitem[Locatello et~al.(2019{\natexlab{b}})Locatello, Bauer, Lucic, Gelly,
  Sch{\"{o}}lkopf, and Bachem]{Locatello2019}
Francesco Locatello, Stefan Bauer, Mario Lucic, Sylvain Gelly, Bernhard
  Sch{\"{o}}lkopf, and Olivier Bachem.
\newblock Challenging common assumptions in the unsupervised learning of
  disentangled representations.
\newblock In \emph{Proceedings of the International Conference on Machine
  Learning (ICML)}, pp.\  4114--4124, 2019{\natexlab{b}}.
\newblock URL \url{http://proceedings.mlr.press/v97/locatello19a.html}.

\bibitem[Locatello et~al.(2020)Locatello, Bauer, Lucic, Raetsch, Gelly,
  Sch{{\"o}}lkopf, and Bachem]{Locatello2020SoberLook}
Francesco Locatello, Stefan Bauer, Mario Lucic, Gunnar Raetsch, Sylvain Gelly,
  Bernhard Sch{{\"o}}lkopf, and Olivier Bachem.
\newblock A sober look at the unsupervised learning of disentangled
  representations and their evaluation.
\newblock \emph{Journal of Machine Learning Research}, 21\penalty0
  (209):\penalty0 1--62, 2020.
\newblock URL \url{http://jmlr.org/papers/v21/19-976.html}.

\bibitem[Maas et~al.(2013)Maas, Hannun, and Ng]{LeakyReLU}
Andrew~L. Maas, Awni~Y. Hannun, and Andrew~Y. Ng.
\newblock Rectifier nonlinearities improve neural network acoustic models.
\newblock In \emph{Proceedings of the Workshop on Deep Learning for Audio,
  Speech, and Language Processing, ICML (WDLASL)}, pp.\  3--9, 2013.
\newblock URL
  \url{http://robotics.stanford.edu/~amaas/papers/relu_hybrid_icml2013_final.pdf}.

\bibitem[Makhzani(2018)]{IAE}
Alireza Makhzani.
\newblock Implicit autoencoders.
\newblock {arXiv: 1805.09804}, 2018.

\bibitem[M\'{e}moli(2011)]{Memoli2011}
Facundo M\'{e}moli.
\newblock Gromov-wasserstein distances and the metric approach to object
  matching.
\newblock \emph{Foundations of Computational Mathematics}, 11\penalty0
  (1):\penalty0 417--487, 2011.
\newblock \doi{10.1007/s10208-011-9093-5}.

\bibitem[Mescheder et~al.(2017)Mescheder, Nowozin, and Geiger]{AVB}
Lars Mescheder, Sebastian Nowozin, and Andreas Geiger.
\newblock Adversarial variational bayes: Unifying variational autoencoders and
  generative adversarial networks.
\newblock In \emph{Proceedings of the International Conference on Machine
  Learning (ICML)}, pp.\  2391--2400, 2017.
\newblock URL \url{http://proceedings.mlr.press/v70/mescheder17a.html}.

\bibitem[Miyato et~al.(2018)Miyato, Kataoka, Koyama, and Yoshida]{SNGAN}
Takeru Miyato, Toshiki Kataoka, Masanori Koyama, and Yuichi Yoshida.
\newblock Spectral normalization for generative adversarial networks.
\newblock In \emph{Proceedings of International Conference on Learning
  Representations (ICLR)}, pp.\  1--26, 2018.
\newblock URL \url{https://openreview.net/forum?id=B1QRgziT-}.

\bibitem[Nguyen et~al.(2021)Nguyen, Nguyen, Ho, Pham, and Bui]{SFG}
Khai Nguyen, Son Nguyen, Nhat Ho, Tung Pham, and Hung Bui.
\newblock Improving relational regularized autoencoders with spherical sliced
  fused gromov wasserstein.
\newblock In \emph{Proceedings of the International Conference on Learning
  Representations (ICLR)}, pp.\  1--11, 2021.
\newblock URL \url{https://openreview.net/forum?id=DiQD7FWL233}.

\bibitem[Paszke et~al.(2019)Paszke, Gross, Massa, Lerer, Bradbury, Chanan,
  Killeen, Lin, Gimelshein, Antiga, Desmaison, Kopf, Yang, DeVito, Raison,
  Tejani, Chilamkurthy, Steiner, Fang, Bai, and Chintala]{PyTorch}
Adam Paszke, Sam Gross, Francisco Massa, Adam Lerer, James Bradbury, Gregory
  Chanan, Trevor Killeen, Zeming Lin, Natalia Gimelshein, Luca Antiga, Alban
  Desmaison, Andreas Kopf, Edward Yang, Zachary DeVito, Martin Raison, Alykhan
  Tejani, Sasank Chilamkurthy, Benoit Steiner, Lu~Fang, Junjie Bai, and Soumith
  Chintala.
\newblock Pytorch: An imperative style, high-performance deep learning library.
\newblock In \emph{Proceedings of Neural Information Processing Systems
  (NeurIPS)}, pp.\  8024--8035, 2019.
\newblock URL
  \url{https://papers.nips.cc/paper/2019/hash/bdbca288fee7f92f2bfa9f7012727740-Abstract.html}.

\bibitem[Rezende \& Viola(2018{\natexlab{a}})Rezende and Viola]{GECO}
Danilo~J. Rezende and Fabio Viola.
\newblock Generalized elbo with constrained optimization, geco.
\newblock In \emph{Proceedings of Neural Information Processing Systems (NIPS)
  Workshop on Bayesian Deep Learning}, pp.\  1--11, 2018{\natexlab{a}}.
\newblock URL \url{http://bayesiandeeplearning.org/2018/papers/33.pdf}.

\bibitem[Rezende \& Viola(2018{\natexlab{b}})Rezende and Viola]{TamingVAEs}
Danilo~J. Rezende and Fabio Viola.
\newblock Taming vaes.
\newblock {arXiv: 1810.00597}, 2018{\natexlab{b}}.

\bibitem[Rezende \& Mohamed(2015)Rezende and Mohamed]{Rezende2015}
Danilo~Jimenez Rezende and Shakir Mohamed.
\newblock Variational inference with normalizing flows.
\newblock In \emph{Proceedings of the International Conference on Machine
  Learning (ICML)}, pp.\  1530--1538, 2015.
\newblock URL \url{http://proceedings.mlr.press/v37/rezende15.html}.

\bibitem[Rezende et~al.(2014)Rezende, Mohamed, and Wierstra]{Rezende2014}
Danilo~Jimenez Rezende, Shakir Mohamed, and Daan Wierstra.
\newblock Stochastic backpropagation and approximate inference in deep
  generative models.
\newblock In \emph{Proceedings of the International Conference on Machine
  Learning (ICML)}, pp.\  1278--1286, 2014.
\newblock URL \url{https://proceedings.mlr.press/v32/rezende14.html}.

\bibitem[Rybkin et~al.(2021)Rybkin, Daniilidis, and Levine]{SigmaVAE}
Oleh Rybkin, Kostas Daniilidis, and Sergey Levine.
\newblock Simple and effective vae training with calibrated decoders.
\newblock In \emph{Proceedings of the International Conference on Machine
  Learning (ICML)}, pp.\  9179--9189, 2021.
\newblock URL \url{http://proceedings.mlr.press/v139/rybkin21a.html}.

\bibitem[Sejourne et~al.(2021)Sejourne, Vialard, and Peyr\'{e}]{Sejourne2021}
Thibault Sejourne, Francois-Xavier Vialard, and Gabriel Peyr\'{e}.
\newblock The unbalanced gromov wasserstein distance: Conic formulation and
  relaxation.
\newblock In \emph{Proceedings of Neural Information Processing Systems
  (NeurIPS)}, volume~34, pp.\  8766--8779, 2021.
\newblock URL
  \url{https://proceedings.neurips.cc/paper/2021/hash/4990974d150d0de5e6e15a1454fe6b0f-Abstract.html}.

\bibitem[S{\o}nderby et~al.(2016)S{\o}nderby, Raiko, Maal{\o}e, S{\o}nderby,
  and Winther]{LadderVAE}
Casper~Kaae S{\o}nderby, Tapani Raiko, Lars Maal{\o}e, S{\o}ren~Kaae
  S{\o}nderby, and Ole Winther.
\newblock Ladder variational autoencoders.
\newblock In \emph{Proceedings of Neural Information Processing Systems
  (NIPS)}, pp.\  3745--3753, 2016.
\newblock URL
  \url{https://papers.nips.cc/paper/2016/hash/6ae07dcb33ec3b7c814df797cbda0f87-Abstract.html}.

\bibitem[S{\o}nderby et~al.(2017)S{\o}nderby, Caballero, Theis, Shi, and
  Husz{\'a}r]{Sonderby2017}
Casper~Kaae S{\o}nderby, Jose Caballero, Lucas Theis, Wenzhe Shi, and Ferenc
  Husz{\'a}r.
\newblock Amortised map inference for image super-resolution.
\newblock In \emph{Proceedings of International Conference on Learning
  Representations (ICLR)}, pp.\  1--17, 2017.
\newblock URL \url{https://openreview.net/forum?id=S1RP6GLle}.

\bibitem[Sturm(2012)]{Sturm2012}
Karl-Theodor Sturm.
\newblock The space of spaces: curvature bounds and gradient flows on the space
  of metric measure spaces.
\newblock {arXiv: 1208.0434}, 2012.

\bibitem[Sugiyama et~al.(2012)Sugiyama, Suzuki, and Kanamori]{Sugiyama2012Book}
Masashi Sugiyama, Taiji Suzuki, and Takafumi Kanamori.
\newblock \emph{Density Ratio Estimation in Machine Learning}.
\newblock Cambridge University Press, 2012.
\newblock \doi{10.1017/CBO9781139035613}.

\bibitem[Thomas et~al.(2017)Thomas, Bengio, Fedus, Pondard, Beaudoin,
  Larochelle, Pineau, Precup, and Bengio]{Thomas2017}
Valentin Thomas, Emmanuel Bengio, William Fedus, Jules Pondard, Philippe
  Beaudoin, Hugo Larochelle, Joelle Pineau, Doina Precup, and Yoshua Bengio.
\newblock Disentangling the independently controllable factors of variation by
  interacting with the world.
\newblock In \emph{Proceedings of Neural Information Processing Systems (NIPS)
  Workshop, Learning Disentangled Representations: from Perception to Control},
  pp.\  1--9, 2017.
\newblock URL
  \url{https://acsweb.ucsd.edu/~wfedus/pdf/ICF_NIPS_2017_workshop.pdf}.

\bibitem[Tishby et~al.(1999)Tishby, Pereira, and Bialek]{InformationBottleneck}
Naftali Tishby, Fernando~C. Pereira, and William Bialek.
\newblock The information bottleneck method.
\newblock In \emph{Proceedings of the 37th Annual Allerton Conference on
  Communication, Control, and Computing}, pp.\  368--377, 1999.
\newblock URL
  \url{https://www.cs.huji.ac.il/labs/learning/Papers/allerton.pdf}.

\bibitem[Tolstikhin et~al.(2018)Tolstikhin, Bousquet, Gelly, and
  Schoelkopf]{WAE}
Ilya Tolstikhin, Olivier Bousquet, Sylvain Gelly, and Bernhard Schoelkopf.
\newblock Wasserstein auto-encoders.
\newblock In \emph{Proceedings of the International Conference on Learning
  Representations (ICLR)}, pp.\  1--16, 2018.
\newblock URL \url{https://openreview.net/forum?id=HkL7n1-0b}.

\bibitem[Tomczak \& Welling(2018)Tomczak and Welling]{VampPrior}
Jakub Tomczak and Max Welling.
\newblock Vae with a vampprior.
\newblock In \emph{Proceedings of the International Conference on Artificial
  Intelligence and Statistics (AISTATS)}, pp.\  1214--1223, 2018.
\newblock URL \url{https://proceedings.mlr.press/v84/tomczak18a.html}.

\bibitem[Tschannen et~al.(2018)Tschannen, Bachem, and Lucic]{Tschannnen2018}
Michael Tschannen, Olivier Bachem, and Mario Lucic.
\newblock Recent advances in autoencoder-based representation learning.
\newblock In \emph{Proceedings of Neural Information Processing Systems (NIPS)
  Workshop on Bayesian Deep Learning}, pp.\  1--25, 2018.
\newblock URL \url{https://www.mins.ee.ethz.ch/pubs/p/autoenc2018}.

\bibitem[Vahdat \& Kautz(2020)Vahdat and Kautz]{NVAE}
Arash Vahdat and Jan Kautz.
\newblock {NVAE}: A deep hierarchical variational autoencoder.
\newblock In \emph{Proceedings of Neural Information Processing Systems
  (NeurIPS)}, volume~33, pp.\  19667--19679, 2020.
\newblock URL
  \url{https://proceedings.neurips.cc/paper/2020/file/e3b21256183cf7c2c7a66be163579d37-Paper.pdf}.

\bibitem[van~der Maaten \& Hinton(2008)van~der Maaten and Hinton]{tSNE}
Laurens van~der Maaten and Geoffrey Hinton.
\newblock Visualizing data using {t-SNE}.
\newblock \emph{Journal of Machine Learning Research}, 9\penalty0
  (86):\penalty0 2579--2605, 2008.
\newblock URL \url{http://jmlr.org/papers/v9/vandermaaten08a.html}.

\bibitem[Villiani(2009)]{Villiani2010}
C{\'{e}}dric Villiani.
\newblock \emph{Optimal Transport: Old and New}.
\newblock Springer Berlin, 2009.
\newblock \doi{10.1007/978-3-540-71050-9}.

\bibitem[Xu et~al.(2020)Xu, Luo, Henao, Shah, and Carin]{RAE}
Hongteng Xu, Dixin Luo, Ricardo Henao, Svati Shah, and Lawrence Carin.
\newblock Learning autoencoders with relational regularization.
\newblock In \emph{Proceedings of the International Conference on Machine
  Learning (ICML)}, pp.\  10576--10586, 2020.
\newblock URL \url{https://proceedings.mlr.press/v119/xu20e.html}.

\bibitem[Zaidi et~al.(2021)Zaidi, Boilard, Gagnon, and Carbonneau]{Zaidi2021}
Julian Zaidi, Jonathan Boilard, Ghyslain Gagnon, and Marc-Andr{\'{e}}
  Carbonneau.
\newblock Measuring disentanglement: A review of metrics.
\newblock {arXiv: 2012.09276}, 2021.

\bibitem[Zhao et~al.(2018)Zhao, Kim, Zhang, Rush, and LeCun]{AAE}
Junbo Zhao, Yoon Kim, Kelly Zhang, Alexander Rush, and Yann LeCun.
\newblock Adversarially regularized autoencoders.
\newblock In \emph{Proceedings of the International Conference on Machine
  Learning (ICML)}, pp.\  5902--5911, 2018.
\newblock URL \url{https://proceedings.mlr.press/v80/zhao18b.html}.

\bibitem[Zhao et~al.(2017)Zhao, Song, and Ermon]{VLadderAE}
Shengjia Zhao, Jiaming Song, and Stefano Ermon.
\newblock Learning hierarchical features from generative models.
\newblock In \emph{Proceedings of the International Conference on Machine
  Learning (ICML)}, pp.\  4091--4099, 2017.
\newblock URL \url{https://proceedings.mlr.press/v70/zhao17c.html}.

\bibitem[Zhao et~al.(2019)Zhao, Song, and Ermon]{InfoVAE}
Shengjia Zhao, Jiaming Song, and Stefano Ermon.
\newblock {InfoVAE}: Balancing learning and inference in variational
  autoencoders.
\newblock In \emph{Proceedings of the AAAI Conference on Artificial
  Intelligence}, pp.\  5885--5892, 2019.
\newblock \doi{10.1609/aaai.v33i01.33015885}.

\bibitem[Zong et~al.(2018)Zong, Song, Min, Cheng, Lumezanu, Cho, and
  Chen]{DAGMM}
Bo~Zong, Qi~Song, Martin~Renquang Min, Wei Cheng, Cristian Lumezanu, Daeki Cho,
  and Haifeng Chen.
\newblock Deep autoencoding gaussian mixture model for unsupervised anomaly
  detection.
\newblock In \emph{Proceedings of the International Conference on Learning
  Representations (ICLR)}, pp.\  1--19, 2018.
\newblock URL \url{https://openreview.net/forum?id=BJJLHbb0-}.

\end{thebibliography}

\appendix

\section{Details of Related Work} \label{sec:appendix-related}

For self-containment, we describe VAE-based representation learning methods.
As with \cref{sec:method}, $\x$ and $\z$ denote data and latent variables, respectively, and the data~$\x$ are $M$-dimensional and the latent variables~$\z$ are $L$-dimensional.
Unless otherwise noted, each VAE-based model consists of a generative model~$\gjoint{\x}{\z}$ with parameters~$\paramDec$, an inference model~$\ijoint{\x}{\z}$ with parameters~$\paramEnc$, and a pre-defined~(non-trainable) prior~$\prior{\z}$ as in the standard VAE model architecture.

\subsection{VAE-based Models with ELBO Extension}

Utilizing the latent variables of VAE-based models is a prominent approach to representation learning.
Several models with extended ELBO-based objectives aim to overcome the shortcomings of the original VAE model, such as posterior collapse.
VAE-based models are mainly grounded on the ELBO objective, where we denote the ELBO for the data point~$\x$ as
\begin{align}
    \elbo{\x}{\paramDec}{\paramEnc}
    &= \ev{\enc{\z}{\x}} \left[ \log \dec{\x}{\z} \right] - \kld{\enc{\z}{\x}}{\prior{\z}}, \label{eq:elbo-objective}
\end{align}
which is mentioned as the expected objective of the original VAE~\cite{VAE} in \cref{eq:vae-objective}.

\subsubsection{$\beta$-VAE}

$\beta$-VAE~\cite{BetaVAE} is a VAE-based model for learning disentangled representations by re-weighting the KL term of the ELBO.
Given a KKT multiplier~$\beta>0$, the $\beta$-VAE objective is expressed as
\begin{align}
    \maximize_{\paramDec, \paramEnc} \quad \ev{\empir{\x}} \left[ \ev{\enc{\z}{\x}} \left[ \dec{\x}{\z} \right] - \beta \kld{\enc{\z}{\x}}{\prior{\z}} \right].
\end{align}
The KKT multiplier~$\beta$ works as the weight of the regularization to impose a factorized prior (\eg, the standard Gaussian~$\mathcal{N}(\mathbf{0},\mathbf{I}_L)$) on the latent variables.
This re-weighting induces the capability of disentanglement in the case of $\beta > 1$; however, a large value of~$\beta$ causes posterior collapse, in which the latent variables ``forget'' the information of the input data.

From the Information Bottleneck~(IB)~\cite{InformationBottleneck} point of view, the $\beta$-VAE objective is re-interpreted as the following optimization problem~\cite{VIB,Achille2018Bottleneck}:
\begin{align}
    \maximize_{\paramDec,\paramEnc} \quad & I_{\paramEnc}(\z;\y) \\
    \subjectto\quad & I_{\paramEnc}(\z;\x) \le I_c, \label{eq:vib-objective}
\end{align}
where $I_c$ is a bottleneck capacity, $\y$ is a task to be estimated, and $I_{\paramEnc}(\cdot;\cdot)$ denotes the mutual information on the inference model.
Introducing the Lagrange multiplier~$\beta$, the IB problem is given as
\begin{align}
    \maximize_{\paramDec,\paramEnc} \quad
        & I_{\paramEnc}(\z;\y) - \beta I_{\paramEnc}(\z;\x).
\end{align}
\citet{VIB} have given the lower bound of this IB objective as
\begin{align}
    I_{\paramEnc}(\z;\y) - \beta I_{\paramEnc}(\z;\x)
    &\ge \underbrace{
        \ev{\empir{\y}\enc{\z}{\y}} \left[ \log \dec{\y}{\z} \right] - \mathcal{H}(\y)
    }_{\text{The lower bound of~$I_{\paramEnc}(\z;\y)$}}
     - \beta \underbrace{
        \ev{\empir{\x}} \left[ \kld{\enc{\z}{\x}}{\prior{\z}} \right]
    }_{\text{The upper bound of~$I_{\paramEnc}(\z;\x)$}},
\end{align}
where the task entropy~$\mathcal{H}(\y)$ is independent of the parameters~$\paramDec$ and~$\paramEnc$.
The autoencoding task~$\y=\x$ gives the objective equivalent to that of the original VAE.
This IB-based formulation of the $\beta$-VAE objective implies that the larger value of the multiplier~$\beta$ guides the training process to minimize the mutual information~$I_{\paramEnc}(\z;\x)$ to make the encoder forget the input data, \ie, to cause posterior collapse.

\subsubsection{FactorVAE}

FactorVAE~\cite{FactorVAE} is a state-of-the-art disentanglement method that minimizes the Total Correlation~(TC) of the aggregated posterior~$\iagg{\z}=\ev{\empir{\x}}[\enc{\z}{\x}]$ in addition to the original ELBO objective.
The TC is expressed as the KL divergence between a distribution and its factorized counterpart.
In the FactorVAE case, the TC of the aggregated posterior is the KL divergence from the factorized aggregated posterior~$\iaggpermuted{\z}=\prod_{i=1}^L \iagg{z_i}$ to the aggregated posterior~$\iagg{\z}$.
The training objective of FactorVAE is the weighted sum of the ELBO and the TC term as
\begin{align}
    \maximize_{\paramDec, \paramEnc} \quad \elbo{\x}{\paramDec}{\paramEnc} - \gamma\totcol{\iagg{\z}},
\end{align}
where $\totcol{\z}$ denotes the TC of the latent variables~$\z$ defined as
\begin{align}
    \totcol{\z} &=
        \kld{\iagg{\z}}{\iaggpermuted{\z}} \\
        &=
          \ev{\iagg{\z}} \left[ \log \frac{ \disc(\z) }{ 1-\disc(\z) } \right]. \label{eq:factorvae-tc}
\end{align}
In \cref{eq:factorvae-tc}, $\disc(\z)$ denotes a discriminator to estimate the TC term by density ratio estimation~\cite{Sugiyama2012Book} as
\begin{align}
        \disc(\z) = \mathrm{arg}\max_{f:\zsp\to[0,1]}
              \ev{\iagg{\z}} \left[ \log f(\z) \right]
            + \ev{\iaggpermuted{\z}} \left[ \log(1-f(\z)) \right].
\end{align}
Practically, the discriminator is estimated using SGD in parallel using samples from~$\iaggpermuted{\z}$ by permuting the latent codes along the batch dimension independently in each latent variable.

\subsubsection{InfoVAE}

InfoVAE~\cite{InfoVAE} is an extension of VAE to prevent posterior collapse by the retention of data information in the latent variables.
The InfoVAE objective is the sum of the ELBO and the inference model mutual information~$I_{\paramEnc}$ in \cref{eq:vib-objective}.
To this end, the following maximization problem is solved via SGD:
\begin{align}
    \maximize_{\paramDec,\paramEnc}\quad
    & \ev{\empir{\x}} \left[ \elbo{\x}{\paramDec}{\paramEnc} \right] + I_{\paramEnc}(\x;\z) \\
    &= \ev{\empir{\x}} \ev{\enc{\z}{\x}} \left[ \dec{\x}{\z} \right] - \kld{\iagg{\z}}{\prior{\z}} \label{eq:infovae-objective}
\end{align}
The main difference between the VAE and InfoVAE objectives is using the regularization term~$\kld{\iagg{\z}}{\prior{\z}}$ instead of the original VAE regularization~$\kld{\enc{\z}{\x}}{\prior{\z}}$.
The original KL term becomes zero if all the data points are encoded into the standard Gaussian~$\mathcal{N}(\mathbf{0},\mathbf{I}_L)$ to cause posterior collapse.
The InfoVAE KL term~$\kld{\iagg{\z}}{\prior{\z}}$ alleviates this problem by adopting the aggregated posterior~$\iagg{\z}$ for optimization instead of the encoder~$\enc{\z}{\x}$.
The authors of InfoVAE~\cite{InfoVAE} further provide the model family in which the KL term is replaced with other divergences.
They introduce an alternative divergence~$\mathcal{D}(\iagg{\z},\prior{\z})$ and its weight~$\lambda$ to conduct representation learning by the following training objective:
\begin{align}
    \maximize_{\paramDec,\paramEnc}\quad
    & \ev{\empir{\x}} \left[ \elbo{\x}{\paramDec}{\paramEnc} \right] + I_{\paramEnc}(\x;\z) \\
    &= \ev{\empir{\x}} \ev{\enc{\z}{\x}} \left[ \dec{\x}{\z} \right] - \lambda \mathcal{D}(\iagg{\z},\prior{\z}).
\end{align}
In the original InfoVAE paper~\cite{InfoVAE}, the authors reported that the Maximum-Mean Discrepancy~(MMD) is the best choice for the divergence~$\mathcal{D}$.
The MMD divergence~$\mmd(\iagg{\z}, \prior{\z})$ is defined as
\begin{align}
    \mmd(\iagg{\z},\prior{\z})
    &= \ev{\iagg{\z}} \ev{\iagg{\z'}} \left[ k(\z,\z') \right]
     + \ev{\prior{\z}} \ev{\prior{\z'}} \left[ k(\z,\z') \right] \nonumber \\
    &\quad\quad - 2\ev{\iagg{\z}} \ev{\prior{\z'}} \left[ k(\z,\z') \right],
\end{align}
where $k(\cdot,\cdot)$ is any universal kernel, such as the radial basis function kernel
\begin{align}
    k(\z,\z')=\exp(-\|\z-\z'\|_2^2/\sigma^2)    
\end{align}
for a constant~$\sigma>0$.

\subsection{VAE-based Methods based on Hierarchical Factors}

Several VAE-based methods postulate the existence of hierarchical factors as its meta-prior to learn representations with the abstractness of different levels~\cite{LadderVAE,VLadderAE}.
These methods involve the change in their network architecture to utilize the feature hierarchy often captured in the hidden layers of deep neural networks.

\subsubsection{Ladder Variational Autoencoder~(LadderVAE)}
Ladder Variational Autoencoder~(LadderVAE)~\cite{LadderVAE} introduces hierarchical latent variables to the VAE model.
Whereas the objective is still the ELBO, the LadderVAE model structure has hierarchical latent variables.
The generative process is modeled as the Markov chain of several latent variable groups, and the inference model consists of deterministic feature encoders and the decoders shared with generative models.
In the original paper~\cite{LadderVAE}, the authors claim that the LadderVAE models provide tighter log-likelihood lower bounds than the standard VAE.

\subsubsection{Variational Ladder Autoencoder~(VLadderAE)}
Variational Ladder Autoencoder~(VLadderAE)~\cite{VLadderAE} is a VAE-based model for hierarchical factors.
Instead of the hierarchical models based on Markov chains, the VLadderAE models introduce the hierarchical structure in the network architecture parameterizing the generative and the inference model.
Since it constrains feature hierarchy by the process of feature extraction, VLadderAE also performs disentanglement, \eg, the latent variables from different hidden convolutional layers capture textural or global features of visual data.

\subsection{VAE-based Methods involving Prior Learning}

The standard VAE model has a pre-defined prior, which may cause the discrepancy between the underlying data structure and the postulated prior~\cite{Dai2019}.
Several methods overcome this problem by involving the prior itself in the training process.

\subsubsection{VampPrior}
VampPrior~\cite{VampPrior} is a type of prior consisting of the mixture of the encoder distributions from several pseudo-input.
The pseudo-inputs are introduced as trainable parameters, which are input into the encoder to build a mixture prior.
Thus, the VAE models with VampPriors have trainable priors while retaining the main training procedure using the reparameterization trick to apply SGD.

\subsubsection{2-Stage VAE}
2-Stage VAE~\cite{Dai2019} is a generative model with two probabilistic autoencoders.
The process of 2-Stage VAE consists of two steps: (i) training a standard VAE using the given dataset as the input, and (ii) training another VAE using the latent variables of the previous VAE as the input.
The 2-Stage VAE model attempts to overcome the discrepancy between the pre-defined prior and the learned latent representation by introducing the second VAE in stage~(ii), which yields the prior training using the VAE in stage~(i).

\subsection{Wasserstein Autoencoder~(WAE)} \label{sec:appendix-wae}

WAE~\cite{WAE} is a family of generative models whose autoencoder tries to estimate and minimize the primal form of the Wasserstein metric between the generative model~$\gmodel{\x}$ and the data distribution~$\empir{\x}$ using SGD with the following objective:
\begin{align}
    \minimize_{\paramDec,\paramEnc}\quad &
        \ev{ \empir{\x} }
            \ev{ \enc{\z}{\x} }
                \ev{ \dec{\x'}{\z} }
                    \left[ d(\x,\x') \right]
        + \lambda \mathcal{D}(\iagg{\z},\prior{\z}),
\end{align}
where $\lambda$ is a Lagrange multiplier, the generative model is defined as a latent variable model~$\gjoint{\x}{\z}=\prior{\z}\dec{\x}{\z}$ postulating the prior of the latent variables~$\prior{\z}$, and a conditional distribution $\enc{\z}{\x}$ is a probabilistic encoder to optimize instead of all couplings supported on~$\xsp\times\xsp$.
The WAE objective is indeed equivalent to that of InfoVAE~\cite{InfoVAE} in \cref{eq:infovae-objective}, which provides the OT-based perspective on VAE-based models.
Following the InfoVAE~\cite{InfoVAE}, we adopt the MMD for the divergence~$\mathcal{D}$, which is denoted by ``WAE-MMD'' in the original WAE paper~\cite{WAE}.
\revise{Although the WAE-based approaches rewrite VAE-based objectives with the Wasserstein metric, these metrics are between $\x$-marginal distributions and do not directly include the latent space~$\zsp$. To learn representations~$\z$, the Wasserstein-based objective is further modified~\cite{TCWAE}.}

\subsection{Relational Regularized Autoencoder (RAE)} \label{sec:appendix-rae}

Relational Regularized Autoencoder~(RAE)~\cite{RAE} is a variational autoencoding generative model with a regularization loss based on the fused Gromov-Wasserstein~(FGW) metric.
RAE introduces the FGW metric between the aggregated posterior and the latent prior as the regularization divergence to fortify the WAE constraint~$\trainablePrior{\z}=\iagg{\z}$ introduced by \citet{WAE} for generative modeling.
The FGW regularization is introduced with a weight hyperparameter~$\beta \in [0, 1]$ and given as
\begin{align}
    \minimize_{\paramDec,\paramEnc}\quad &
        \ev{ \empir{\x} }
            \ev{ \enc{\z}{\x} }
                \ev{ \dec{\x'}{\z} }
                    \left[ d(\x,\x') \right]
        + \lambda \mathcal{D}_{FGW}(\iagg{\z},\trainablePrior{\z}; \beta),
\end{align}
where $\mathcal{D}_{FGW}$ denotes the FGW metric being the upper bound of the weighted sum of the Wasserstein and Gromov-Wasserstein metrics.
The FGW metric~$\mathcal{D}_{FGW}$ is given as
\begin{align}
    & \mathcal{D}_{FGW}(\iagg{\z},\trainablePrior{\z}; \beta) \nonumber \\
    = \quad & \inf_{\gamma \in \couplingset(\iagg{\z}, \trainablePrior{\z})}
        \Big(
            (1-\beta) \ev{ \gamma(\z, \z') } [ \distz(\z, \z') ]
            + \beta \ev{ \gamma(\z_1, \z_1') \gamma(\z_2, \z_2') } [ | \distz(\z_1, \z_2) - \distz(\z_1', \z_2') | ]
        \Big) \\
    \ge \quad &
        (1-\beta) \underbrace{\inf_{\gamma \in \couplingset(\iagg{\z}, \trainablePrior{\z})}
             \ev{ \gamma(\z, \z') } [ \distz(\z, \z') ]}_{\text{Wasserstein term for direct comparison}} \nonumber \\
        & + \beta \underbrace{ \inf_{\gamma \in \couplingset(\iagg{\z}, \trainablePrior{\z})}
            \ev{ \gamma(\z_1, \z_1') \gamma(\z_2, \z_2') } [ | \distz(\z_1, \z_2) - \distz(\z_1', \z_2') |^2 ] }_{\text{Gromov-Wasserstein term for relational comparison}},
\end{align}
where $\couplingset(\iagg{\z}, \trainablePrior{\z})$ is a set of all couplings whose marginals are $\iagg{\z}, \trainablePrior{\z}$.
The discrepancy between the prior~$\trainablePrior{\z}$ and the aggregated posterior~$\iagg{\z}$ causes the degradation of generative performance since the processes of decoding~$\empir{\x}\enc{\z}{\x}\dec{\x}{\z}$ and generation~$\trainablePrior{\z}\dec{\x}{\z}$ are modeled in different regions of the latent space.
This formulation enables learning a prior distribution~$\iagg{\z}$ with flexibly assuming the structures of data, where the prior~$\trainablePrior{\z}$ is modeled as a Gaussian mixture model the original settings by \citet{RAE}.
They aim at matching the distributions on the latent space~$\zsp$, which can have an identical dimensionality but may differ in terms of distance structure.

\subsection{IVI Methods} \label{sec:appendix-ivi}

Beyond the analytically tractable distributions, implicit distributions are applied to variational inference.
An implicit distribution only requires its sampling method, which extends the variety of modeling and applications in variational inference and VAE-based models.

\subsubsection{Density Ratio Estimation by Adversarial Discriminators}

The density ratio estimation technique~\cite{Sugiyama2012Book} is essential to the mechanism of GANs~\cite{GAN} and IVI methods~\cite{Huszar2017}, which is conducted via an optimal discriminator~$f^*$ between distributions~$\onedistro(\x)$ and~$\otherdistro(\x)$ as
\begin{align}
    \kld{\onedistro(\x)}{\otherdistro(\x)}
        &= \ev{\onedistro(\x)} \left[
            \log \frac{ \onedistro(\x) }{ \otherdistro(\x) }
        \right]
        = \ev{\onedistro(\x)} \left[
            \log \frac{ f^*(\x) }{ 1-f^*(\x) }
        \right] \nonumber \\
        &= \ev{\onedistro(\x)} \left[
            \log f^*(\x) - \log (1-f^*(\x))
        \right], \label{eq:dre-kl} \\
        \text{where}\quad f^*(\x) &= \mathrm{arg}\max_{ f: \xsp \to (0,1) }
            \ev{\onedistro(\x)} \left[
                \log f(\x)
            \right]
            +
            \ev{\otherdistro(\x)} \left[
                \log (1-f(\x))
            \right]. \label{eq:discriminator}
\end{align}
The discriminator is estimated via maximizing \cref{eq:discriminator} with a neural network~$f \approx f^*$.
The training of discriminators often suffers from instability and mode collapse owing to its alternative parameter updates based on \cref{eq:dre-kl} and \cref{eq:discriminator}~\cite{Arjovsky2017,WGAN}.
One approach to tackle this problem is imposing the Lipschitz continuity on the discriminator based on the Kantorovich-Rubinstein duality~\cite{WGAN}.

\subsubsection{Adversarial Variational Bayes~(AVB)}

Adversarial Variational Bayes~(AVB)~\cite{AVB} is an ELBO optimization method using the adversarial training process instead of the analytical KL term.
Let us recall that the KL term in \cref{eq:vae-objective} is defined by the expected density ratio as
\begin{align}
    \kld{\enc{\z}{\x}}{\prior{\z}}
    = \ev{\enc{\z}{\x}}{\left[
        \frac{
            \enc{\z}{\x}
        }{
            \prior{\z}
        }
    \right]}.
\end{align}
Adopting the density ratio trick~\cite{Sugiyama2012Book}, the analytical KL term can be replaced with the optimal discriminator, which takes a data point~$\x$ and its encoder sample~$\z \sim \enc{\z}{\x}$ to output the density ratio~$\enc{\z}{\x}/\prior{\z}$.
It enables implicit distributions in the prior while retaining the ELBO objective of variational inference.

\subsubsection{Adversarially Learned Inference~(ALI) / Bidirectional Generative Adversarial Networks~(BiGAN)}

Adversarially Learned Inference~(ALI)~\cite{ALI} / Bidirectional Generative Adversarial Networks~(BiGAN)~\cite{BiGAN} are models introducing the distribution matching of the generative model and the inference model as implicit distributions.
These models have been proposed in different papers~\cite{ALI,BiGAN}; however, they share an equivalent methodology.
One can draw samples from the generative model~$\prior{\z}\dec{\x}{\z}$ by decoding prior samples and also from the inference model~$\empir{\x}\enc{\z}{\x}$ by encoding data points.
Here the ALI/BiGAN models introduce a discriminator to estimate the Jensen-Shannon divergence between the generative model~$\gjoint{\x}{\z}$ and the inference model~$\ijoint{\x}{\z}$.
The model matching between the encoder and the decoder also learns latent representations by the bidirectional mappings.

\subsubsection{VAE-GAN}

VAE-GAN~\cite{VAEGAN} is a hybrid model based on VAE and GANs.
The VAE-GAN models introduce a discriminator for the generative modeling \wrt the data~$\x$ and utilize the hidden layers of the discriminator to model the decoder likelihood $\dec{\x}{\z}$ along the manifolds supporting the data.
It provides the outstanding performance of data generation to the VAE framework by measuring the similarity of data utilizing the GANs-like network architecture.

\section{Details of Proposed Method} \label{sec:appendix-method}

\subsection{Modeling Details \label{sec:appendix-model}}

The decoder~$\dec{\x}{\z}$ is modeled with a neural network~$D_{\paramDec}: \zsp\to\real^M$ and its parameters~$\paramDec$ as
\begin{align}
    \dec{\x}{\z}
    &= \delta(\x - D_{\paramDec}(\z)).
\end{align}

Following the standard VAE settings~\cite{VAE}, the encoder~$\enc{\z}{\x}$ is defined as a diagonal Gaussian parameterized by neural networks~$\bm{\mu}_{\paramEnc}: \zsp\to\real^M$ and~$\bm{\sigma}^2_{\paramEnc}:\zsp\to\real_{+}^M$ with parameters~$\paramEnc$ as
\begin{align}
    \enc{\z}{\x} = \mathcal{N}(\z | \bm{\mu}_{\paramEnc}(\x), \mathrm{diag}(\bm{\sigma}_{\paramEnc}^2(\x))).
\end{align}

For the distance functions~$\distx$ and~$\distz$ in \cref{eq:loss-gw} and \cref{eq:loss-w}, we used the $L_2$ distance defined as
\begin{align}
    \distx(\x,\x') &= \frac{1}{\sqrt{2}} \| \x - \x' \|, \\
    \distz(\z,\z') &= \frac{1}{\sqrt{2}} \| \z - \z' \|.
\end{align}
As another choice, we also utilized the adversarially learned metric~\cite{VAEGAN} in \cref{eq:loss-w}. 
In the adversarially learned metric, the distance is measured in the feature space formed by the hidden outputs of the critic~$\critic$.
Let $h_{\paramCritic}(\x)$ denote the critic hidden outputs in which the critic takes $\x$ as its input.
We can then define a distance~$d'$ based on the adversarially learned metric as
\begin{align}
    d'(\x, \x') &= \sqrt{ \distx(\x, \x')^2 + \frac{1}{2} \left\| h_{\paramCritic}(\x) - h_{\paramCritic}(\x') \right\|_2^2 }.
\end{align}
Since the critic network~$\critic(\x,\z)$ has the Y-shaped architecture~(see \cref{sec:appendix-architecture}) and concatenates $\x$-based features and $\z$-based features in one of the hidden layers to take a pair $(\x, \z)$ as the inputs, we use the $\x$-side branch as~$h_{\paramCritic}(\x)$.

\subsection{Prior Details \label{sec:appendix-prior}}

\textbf{Neural Prior (NP).}
Formally, the NP~$\trainablePrior{\z}$ with a neural network~$g_{\paramDec}$ is defined as:
\begin{align}
    \trainablePrior{\z} = \int
        \prior{\bm{\epsilon}}
        \left| \mathrm{det} \frac{\partial g_{\paramDec}(\bm{\epsilon})}{\partial \bm{\epsilon}} \right|
    d\bm{\epsilon}, \\
    \mathrm{where} \quad
    \prior{\bm{\epsilon}} = \mathcal{N}(\bm{\epsilon}|\mathbf{0}, \mathbf{I}_L).
\end{align}
We can implement this class of prior with sampling noises~$\bm{\epsilon}$ as~$\z=g_{\paramDec}(\bm{\epsilon})$, avoiding the calculation of the integral.

\textbf{Factorized Neural Prior.}
For disentanglement in the variational autoencoding settings, element-wise independence is often imposed on latent variables~$\z$.
Following the standard VAE settings~\cite{VAE}, we postulate $\zsp=\real^L$, where the latent variables $\z \in \zsp$ are expressed as an $L$-dimensional vector~$\z=[z_1,z_2,\ldots,z_L]^\T$.
As with the NP, the FNP class of prior is defined as
\begin{align}
    & \trainablePrior{\z} = \prod_{i=1}^L \facPrior{i}{z_i}, \\
    \mathrm{where} \quad
    & \facPrior{i}{z_i} = \int \prior{\epsilon^{(i)}} \left| \frac{\partial \tilde{g}^{(i)}_{\paramDec}(\epsilon^{(i)})}{\partial \epsilon^{(i)}} \right| d\epsilon^{(i)}, & (i=1,2,\ldots,L) \\
    & \prior{\epsilon^{(i)}} = \mathcal{N}(\epsilon^{(i)}|0, 1). & (i=1,2,\ldots,L)
\end{align}
This prior can be implemented with $N$ disjoint neural networks, or 1-dimensional grouped convolutions.
The difference between the NP and the FNP is element-wise independence, in which the prior~$\trainablePrior{\z}$ is factorized into distributions for each latent variable.
Factorized priors enable disentanglement by obtaining a representation comprising independent factors of variation~\cite{BetaVAE,BetaTCVAE,FactorVAE}.

\subsection{Gradient Penalty \label{sec:appendix-gp}}
In the case of gradient penalty~\cite{WGANGP}, the maximization in \cref{eq:loss-d} is further modified as
\begin{align}
    \maximize_{\paramCritic} \quad &
        \lossd + \coeffgp \ev{\ijoint{\x}{\z}} \ev{\gjoint{\x'}{\z'}} \ev{\epsilon \sim \mathcal{U}(0,1)} \left[
            \left(
                \|\nabla_{(\xintp,\zintp)} \critic(\xintp,\zintp)\|_2 - 1
            \right)^2 \label{eq:loss-gp}
        \right],
\end{align}
where $\coeffgp > 0$ is a constant, and $\xintp=\epsilon\x + (1-\epsilon)\x'$ and~$\zintp=\epsilon\z + (1-\epsilon)\z'$ are interpolated samples by the random uniform noise~$\epsilon$.
We adopt $\coeffgp=10$ in all the experiments reported in this paper.
Introducing the gradient penalty together with other techniques such as spectral normalization~\cite{SNGAN} is effective and essential for adversarial learning in general~\cite{Chu2020,SNGAN}.

\section{Experimental Details} \label{sec:appendix-experiments}
For the reported experimental results, we used a single GPU of NVIDIA GeForce\textregistered\ RTX 2080 Ti, and a single run of the entire GWAE training process until convergence takes about eight hours.

\subsection{Dataset Details \label{sec:appendix-datasets}}

For the reported experiments in \cref{sec:experiments}, we used the following datasets:
\begin{description}
    \item[MNIST~\cite{MNIST}.]
        The MNIST dataset contains 70,000 handwritten digit images of 10 classes, comprising 60,000 training images and 10,000 test images.
        We used the original test set and randomly split the original training set into 54,000 training images and 6,000 validation images.
        We used the class information as its approximate factors of variation in the form of $10$-dimensional dummy variables.
        This dataset is available online\footnote{\url{http://yann.lecun.com/exdb/mnist/}} in its original format or via the \texttt{torchvision} package\footnote{\url{https://github.com/pytorch/vision}} in the PyTorch~\cite{PyTorch} tensor format.
        The MNIST dataset is licensed under the terms of the Creative Commons Attribution-Share Alike 3.0 license\footnote{\url{https://creativecommons.org/licenses/by-sa/3.0/}}.
    \item[CelebA~\cite{CelebA}.]
        The CelebA dataset contains 202,599 aligned face images with 40 binary attributes.
        We cropped $144 \times 144$ pixels in the center of the $178 \times 218$-sized aligned images in the original dataset to omit excessive backgrounds.
        We used the train/validation/test partitions that the original authors provided.
        We used the binary attributes as its approximate factors of variation in the form of $40$-dimensional vectors.
        As in the website of this dataset\footnote{\url{https://mmlab.ie.cuhk.edu.hk/projects/CelebA.html}}, the CelebA dataset is available for non-commercial research purposes only.
    \item[3D Shapes~\cite{3DShapes}.]
        The 3D Shapes dataset contains 480,000 synthetic images with six ground truth factors of variation.
        The images in this dataset contain a single-colored 3D object, a single-colored wall of a rectangular room, a single-colored floor.
        These images are procedurally generated from the independent factors of variation, \emph{floor colour}, \emph{wall colour}, \emph{object colour}, \emph{scale}, \emph{shape}, and \emph{orientation}~\cite{3DShapes}.
        We randomly split the entire dataset into 384,000/48,000/48,000 images for the train/validation/test set, respectively.
        Since the factor \emph{shape} is a categorical variable in four classes, we converted it into four dummy variables to obtain quantitative factors of variation in the form of $9$-dimensional vectors.
        The repository of this dataset\footnote{\url{https://github.com/deepmind/3d-shapes}} is licensed under Apache License 2.0\footnote{\url{http://www.apache.org/licenses/}}.
    \item[Omniglot~\cite{Omniglot}.]
        The Omniglot dataset contains 1,623 images of hand-written characters from 50 different alphabets written by 20 different people.
        The images are $105 \times 105$-sized, binary-valued.
        We used this dataset as OoD samples over MNIST in the evaluations on the OoD detection utilizing cluster structure.
        The repository of this dataset\footnote{\url{https://github.com/brendenlake/omniglot}} is licensed under the MIT License\footnote{\url{https://opensource.org/licenses/MIT}}.
    \revise{
    \item[CIFAR-10~\cite{CIFAR10}.]
        The CIFAR10 dataset contains 60,000 images with 10 classes, comprising 50,000 training images and 10,000 test images. The images are 32x32 color images in 10 natural image classes, such as airplane and cat. This dataset is provided online\footnote{\url{https://www.cs.toronto.edu/~kriz/cifar.html}} without any specific license.
    }
\end{description}
In all the datasets above, we used all the images as the raster (bitmap) representation and resized them to $64 \times 64$ pixels with three channels, where each image is a $3 \times 64 \times 64$-sized tensor value and~$M=12,288$. For gray-scale (one-channeled) images such as in MNIST, we repeated these images along the channel dimension three times to uniform these sizes to~$3 \times 64 \times 64$ elements.

\subsection{Architecture Details \label{sec:appendix-architecture}}

The architecture of neural networks in GWAE and the compared methods are built with convolutions and deconvolution (transposed convolution) in the same settings as shown in \cref{tab:architecture-encoder,tab:architecture-decoder}.
In all the experiments on GWAE, we applied the gradient penalty and the spectral normalization in the critic networks to impose the 1-Lipschitz continuity on the critic~$\critic$, as shown in \cref{tab:architecture-critic}.
In the neural samplers of GWAE models, we used the fully-connected architecture in \cref{tab:architecture-sampler-np} for NP and the grouped-convolutional architecture in \cref{tab:architecture-sampler-fnp} for FNP.
We used fully connected layers for unconstrained priors in NP, and 1-dimensional grouped convolution layers (converting sequences with length 1 and $L$ channels) for factorized priors in FNP.
For the optimizers of GWAE, we used RMSProp\footnote{\url{https://www.cs.toronto.edu/~tijmen/csc321/slides/lecture_slides_lec6.pdf}} with a learning rate of~$10^{-4}$ for the main autoencoder network and used RMSProp with a learning rate of~$5 \times 10^{-5}$ for the critic network.
For all the compared methods except for GWAE, we used the Adam~\cite{Adam} optimizer with a learning rate of~$10^{-4}$.
In the experiments, we used an equal batch size of 64 for all evaluated models.
The batch size is relatively small, since the computational cost of GWAE for each batch is quadratic to the batch size~$B$ and the GW estimation runs in time $O(NB)$ for each epoch using $\lceil N/B \rceil$ batches.

\newcommand{\encmu}{\bm{\mu}}
\newcommand{\encsg}{\bm{\sigma}}
\newcommand{\encmubn}{\tilde{\bm{\mu}}}
\newcommand{\encsgbn}{\tilde{\bm{\sigma}}}
\newcommand{\batchset}{\mathcal{B}}
\newcommand{\batchsize}{\#\batchset}
In the case that a batch normalization layer is introduced in the encoder outputs~$\enc{\z_i}{\x}=\mathcal{N}(\encmubn(\x), \mathrm{diag}(\encsgbn^2(\x)))$, the mean and variance are computed \wrt the aggregated posterior~$\iagg{\z}$ rather than the element-wise sample mean and variance of $L$-dimensional output values.
The normalized parameters~$(\encmubn(\x), \encsgbn(\x))$ against the original parameters~$(\encmu(\x), \encsg(\x))$ are given as
\begin{align}
    \encmubn(\x) &= \frac{ \encmu(\x) - \ev{\iagg{\z}}[\z] }{ \sqrt{ \mathbb{V}_{\iagg{\z}}[\z] } }, \\
    \encsgbn^2(\x) &= \frac{ \encsg^2(\x) }{ \mathbb{V}_{\iagg{\z}}[\z] },
\end{align}
where the division is element-wise conducted, and $\mathbb{V}$ denotes the variance.
The mean~$\ev{\iagg{\z}}[\z]$ and variance~$\mathbb{V}_{\iagg{\z}}[\z]$ are approximated using unbiased estimators consisting of mini-batch samples.
Given a mini-batch index set~$\batchset \subseteq \{1,2,\ldots,N\}$, the unbiased estimations are expressed using the law of total variance as
\begin{align}
    \ev{\iagg{\z}}[\z] &\approx \frac{1}{\batchsize} \sum_{i\in\batchset} \encmu(\x_i) =: \hat{\bm{\mu}}, \\
    \mathbb{V}_{\iagg{\z}}[\z] &\approx \frac{1}{\batchsize} \sum_{i\in\batchset} \encsg^2(\x_i) + \frac{1}{\batchsize-1} \sum_{i\in\batchset} (\encmu(\x) - \hat{\bm{\mu}})^2.
\end{align}

\newcommand{\imageshape}[3]{ $#1 \times #2 \times #3$ }
\begin{table}[!t]
    \centering
    \caption{
        Model architecture for the encoders in the GWAE models and the compared models.
        For the $64 \times 64$ RGB images used in the experiments, the input size is set to $(\text{Channels},\text{Height},\text{Width})=(3,64,64)$.
        FC and Conv denote fully-connected~(linear) layers and convolutional layers, respectively.
    }
    \label{tab:architecture-encoder}
    \begin{tabular}{cccc}
        \toprule
        Layer & Input Shape & Output Shape & Options \\
        \midrule
        \multicolumn{4}{c}{Inverse Sigmoid~$\sigma^{-1}(x)=\log \frac{x}{1 - x}$} \\
        Conv & \imageshape{3}{64}{64} & \imageshape{32}{32}{32} & kernel size=4, stride=2, padding=1 \\
        \multicolumn{4}{c}{SiLU activation~\cite{GELU}} \\
        Conv & \imageshape{32}{32}{32} & \imageshape{64}{16}{16} & kernel size=4, stride=2, padding=1 \\
        \multicolumn{4}{c}{SiLU activation~\cite{GELU}} \\
        Conv & \imageshape{64}{16}{16} & \imageshape{128}{8}{8} & kernel size=4, stride=2, padding=1 \\
        \multicolumn{4}{c}{SiLU activation~\cite{GELU}} \\
        Conv & \imageshape{128}{8}{8} & \imageshape{256}{4}{4} & kernel size=4, stride=2, padding=1 \\
        \multicolumn{4}{c}{SiLU activation~\cite{GELU}} \\
        FC & \imageshape{256}{4}{4} & 256 & bias=True \\
        \multicolumn{4}{c}{SiLU activation~\cite{GELU}} \\
        FC & 256 & $L$ for $\bm{\mu}$, $L$ for $\bm{\sigma}^2$ & bias=True \\
        \bottomrule
    \end{tabular}
\end{table}

\begin{table}[!t]
    \centering
    \caption{
        Model architecture for the decoders in the GWAE models and the compared models.
        The image shape is set to the same as \cref{tab:architecture-encoder}.
        FC and DeConv denote fully-connected layers and deconvolutional layers, respectively.
    }
    \label{tab:architecture-decoder}
    \begin{tabular}{cccc}
        \toprule
        Layer & Input Shape & Output Shape & Options \\
        \midrule
        FC & $L$ & 256 & bias=True \\
        \multicolumn{4}{c}{SiLU activation~\cite{GELU}} \\
        FC & 256 & \imageshape{256}{4}{4} & bias=True \\
        \multicolumn{4}{c}{SiLU activation~\cite{GELU}} \\
        DeConv & \imageshape{256}{4}{4} & \imageshape{128}{8}{8} & kernel size=4, stride=2, padding=1 \\
        \multicolumn{4}{c}{SiLU activation~\cite{GELU}} \\
        DeConv & \imageshape{128}{8}{8} & \imageshape{64}{16}{16} & kernel size=4, stride=2, padding=1 \\
        \multicolumn{4}{c}{SiLU activation~\cite{GELU}} \\
        DeConv & \imageshape{64}{16}{16} & \imageshape{32}{32}{32} & kernel size=4, stride=2, padding=1 \\
        \multicolumn{4}{c}{SiLU activation~\cite{GELU}} \\
        DeConv & \imageshape{32}{32}{32} & \imageshape{3}{64}{64} & kernel size=4, stride=2, padding=1 \\
        \multicolumn{4}{c}{Sigmoid~$\sigma(x)=\frac{1}{1+e^{-x}}$} \\
        \bottomrule
    \end{tabular}
\end{table}

\begin{table}[!t]
    \centering
    \caption{
        Model architecture for the samplers in the GWAE models with NP.
        FC denotes a fully-connected layer.
    }
    \label{tab:architecture-sampler-np}
    \begin{tabular}{cccc}
        \toprule
        Layer & Input Shape & Output Shape & Options \\
        \midrule
        FC & $L$ & 256 & bias=True \\
        \multicolumn{4}{c}{SiLU activation~\cite{GELU}} \\
        FC & 256 & 256 & bias=True \\
        \multicolumn{4}{c}{SiLU activation~\cite{GELU}} \\
        FC & 256 & 256 & bias=True \\
        \multicolumn{4}{c}{SiLU activation~\cite{GELU}} \\
        FC & 256 & $L$ & bias=True \\
        \multicolumn{4}{c}{Batch Normalization with affine=False} \\
        \bottomrule
    \end{tabular}
\end{table}

\begin{table}[!t]
    \centering
    \caption{
        Model architecture for the samplers in the GWAE models with FNP.
        GroupConv denotes 1-dimensional grouped convolutional layers.
    }
    \label{tab:architecture-sampler-fnp}
    \begin{tabular}{cccc}
        \toprule
        Layer & Input Shape & Output Shape & Options \\
        \midrule
        GroupConv & $L$ & 256 & bias=True, groups=$L$ \\
        \multicolumn{4}{c}{SiLU activation~\cite{GELU}} \\
        GroupConv & 256 & 256 & bias=True, groups=$L$ \\
        \multicolumn{4}{c}{SiLU activation~\cite{GELU}} \\
        GroupConv & 256 & 256 & bias=True, groups=$L$ \\
        \multicolumn{4}{c}{SiLU activation~\cite{GELU}} \\
        GroupConv & 256 & $L$ & bias=True, groups=$L$ \\
        \multicolumn{4}{c}{Batch Normalization with affine=False} \\
        \bottomrule
    \end{tabular}
\end{table}

\begin{table}[!t]
    \centering
    \caption{
        Model architecture for the critics in the GWAE models.
        We concatenated the outputs of the $\x$-side and $\z$-side branches and multiplied the concatenated outputs by $0.5$ to input into the stem network for the sake of the gradient norm, resulting in a Y-shaped network.
        We applied spectral normalization~\cite{SNGAN} to all the layers in the critic networks and used the LeakyReLU~\cite{LeakyReLU} activation for the critic to retain the 1-Lipschitz continuity.
        FC and Conv denote fully-connected layers and convolutional layers, respectively.
    }
    \label{tab:architecture-critic}
    \begin{tabular}{cccc}
        \toprule
        Layer & Input Shape & Output Shape & Options \\
        \midrule
        \multicolumn{4}{c}{$\x$-side branch} \\
        Conv & \imageshape{3}{64}{64} & \imageshape{8}{32}{32} & kernel size=4, stride=2, padding=1 \\
        \multicolumn{4}{c}{LeakyReLU activation~\cite{LeakyReLU} with negative slope 0.2} \\
        Conv & \imageshape{8}{32}{32} & \imageshape{16}{16}{16} & kernel size=4, stride=2, padding=1 \\
        \multicolumn{4}{c}{LeakyReLU activation~\cite{LeakyReLU} with negative slope 0.2} \\
        Conv & \imageshape{16}{16}{16} & \imageshape{32}{8}{8} & kernel size=4, stride=2, padding=1 \\
        \multicolumn{4}{c}{LeakyReLU activation~\cite{LeakyReLU} with negative slope 0.2} \\
        Conv & \imageshape{32}{8}{8} & \imageshape{64}{4}{4} & kernel size=4, stride=2, padding=1 \\
        \multicolumn{4}{c}{LeakyReLU activation~\cite{LeakyReLU} with negative slope 0.2} \\
        Conv & \imageshape{64}{4}{4} & \imageshape{128}{2}{2} & kernel size=4, stride=2, padding=1 \\
        \multicolumn{4}{c}{LeakyReLU activation~\cite{LeakyReLU} with negative slope 0.2} \\
        Conv & \imageshape{128}{2}{2} & \imageshape{256}{1}{1} & kernel size=4, stride=2, padding=1 \\
        \multicolumn{4}{c}{LeakyReLU activation~\cite{LeakyReLU} with negative slope 0.2} \\
        FC & 256 & 64 & bias=True \\
        \midrule
        \multicolumn{4}{c}{$\z$-side branch} \\
        FC & $L$ & 256 & bias=True \\
        \multicolumn{4}{c}{LeakyReLU activation~\cite{LeakyReLU} with negative slope 0.2} \\
        FC & 256 & 256 & bias=True \\
        \multicolumn{4}{c}{LeakyReLU activation~\cite{LeakyReLU} with negative slope 0.2} \\
        FC & 256 & 64 & bias=True \\
        \multicolumn{4}{c}{LeakyReLU activation~\cite{LeakyReLU} with negative slope 0.2} \\
        \midrule
        \multicolumn{4}{c}{Stem network} \\
        \multicolumn{4}{c}{$\z$-side branch} \\
        FC & 64+64 & 256 & bias=True \\
        \multicolumn{4}{c}{LeakyReLU activation~\cite{LeakyReLU} with negative slope 0.2} \\
        FC & 256 & 256 & bias=True \\
        \multicolumn{4}{c}{LeakyReLU activation~\cite{LeakyReLU} with negative slope 0.2} \\
        FC & 256 & 1 & bias=True \\
        \multicolumn{4}{c}{LeakyReLU activation~\cite{LeakyReLU} with negative slope 0.2} \\
        \bottomrule
    \end{tabular}
\end{table}

\subsection{Quantitative Evaluation Details \label{sec:appendix-metrics}}

For quantitative evaluations, we used the DCI scores~\cite{DCIScore} for disentanglement, the FID score~\cite{FID} for image generation, and the PSNR score for image reconstruction.

\subsubsection{DCI Scores}

The DCI scores~\cite{DCIScore} measure a representation in terms of disentangled representation learning.
In the DCI scores, disentanglement is measured from three aspects: (i) each representation variable represents a single factor of variation, (ii) each factor of variation is expressed by a single representation variable, and (iii) a representation is informative \wrt the original data.
The correspondence of variables and factors is computed via estimating the ground truth factors from the representation using random forest~\cite{RandomForest}.
DCI Disentanglement~(DCI-D) measures (i) the factor singleness for each variable.
DCI Completeness~(DCI-C) measures (ii) the variable singleness for each factor.
DCI Informativeness~(DCI-I) measures (iii) whether the representation is informative for estimating the ground truth factors.
These metrics are computed via the variable importances (\eg, the Gini impurity~\cite{RandomForest}) of the random forest~\cite{RandomForest}, in which the random forest regressor estimates the ground truth factors using the representation variables.
Using $L$-dimensional representation variables~$\z$, $V$-dimensional factors~$\y$ and their importance $R_{ij}$ of the $i$-th variable~$z_i$ for the $k$-th factor~$y_k$, the DCI-D and DCI-C scores for each variable and each factor are defined as
\begin{align}
    \text{DCI-D}_i &= 1 + \sum_{k=1}^V p_{ik} \log_V p_{ik}, & (i=1,2,\ldots,L) \\
    & \text{where } p_{ik} = \left. R_{ik} \middle/ \sum_{j=1}^V R_{ij} \right. , \\
    \text{DCI-C}_k &= 1 + \sum_{i=1}^L q_{ik} \log_V q_{ik}, & (k=1,2,\ldots,V) \\
    & \text{where } q_{ik} = \left. R_{ik} \middle/ \sum_{j=1}^V R_{jk} \right. .
\end{align}
The DCI-D score for the entire variable set is given by the weighted sum~$\sum_{i=1}^L \rho_i \text{DCI-D}_i$, where the weight~$\rho_i$ is weighted importance~$\rho_i=( \sum_{k=1}^V R_{ik} ) / (\sum_{i=1}^{L} \sum_{k=1}^{V} R_{ik} $).
The DCI-C score for the entire factor set is given by the average score~$1/V \sum_{k=1}^V \text{DCI-C}_k$.
The DCI-D and DCI-C metrics take values within the range~$[0, 1]$, where higher values indicate better performance.
For DCI-I, we used the normalized definition by \citet{Zaidi2021} because the normalized DCI-I values are within the range~$[0, 1]$ and the higher values mean better informativeness, while DCI-I score~$\text{DCI-D}_{Original}$ is the estimation mean square error in the original definition.
The DCI-I definition that we used is expressed as
\begin{align}
    \text{DCI-I} = 1 - 6 \times \text{DCI-I}_{Original}.
\end{align}
Following the original paper~\cite{DCIScore}, we set the number of random trees to 10 and decided the tree depth with cross-validation.

\subsubsection{Fr{\'e}chet Inception Distance~(FID)}

Fr{\'e}chet Inception Distance~(FID)~\cite{FID} is a score for evaluating the quality of the generated images by generative models.
The FID score is defined as the squared $2$-Wasserstein metric between the features of the real images with mean~$(\bm{\mu}_r, \bm{\Sigma}_r)$ and that of the generated images with mean~$(\bm{\mu}_g, \bm{\Sigma}_g)$.
Assuming that the features are normally distributed in the feature space, the FID score is expressed as
\begin{align}
    \text{FID}
    &= W_2^2(\mathcal{N}(\bm{\mu}_r, \bm{\Sigma}_r), \mathcal{N}(\bm{\mu}_g, \bm{\Sigma}_g)) \\
    &= \| \bm{\mu}_r - \bm{\mu}_g \|_2^2 + \tr( \bm{\Sigma}_r + \bm{\Sigma}_g - 2(\bm{\Sigma}_r \bm{\Sigma}_g)^{\frac{1}{2}} ).
\end{align}
Since the Wasserstein metric measures the discrepancy between distributions, lower values indicate better generation performance in the FID score.
Following the original FID paper~\cite{FID}, we used the features obtained from the final pooling layer outputs of the Inception-v3 pre-trained in the ImageNet dataset~\cite{ImageNet}.

\subsubsection{Peak Signal-to-Noise Ratio~(PSNR)}
For measuring the image reconstruction, we used the Peak Signal-to-Noise Ratio~(PSNR) value.
The PSNR value is defined as
\begin{align}
    \text{PSNR} &= 20 \log_{10}(\text{MAX}) - 10 \log_{10}(\text{MSE}),
\end{align}
where MAX denotes the maximum value of the pixel values, and MSE indicates the mean square error~(MSE).
In all the experiments conducted in \cref{sec:experiments}, the value of MAX is set to $\text{MAX}=1$ because the images input as a dataset~$\data$ are scaled within the range~$[0, 1]$.






\subsection{Isometry Comparison} \label{sec:isometry-comparison}

Regarding the evaluations in \cref{sec:experiments-isometry}, we further conducted comparisons on isometry in \cref{fig:isometry}.
The results show that the GWAE models provide more isometric autoencoders compared with other VAE-based representation learning methods.
The existing VAE-based methods did not yield as far as GWAE, which supports that the GW metric works as a different objective class from the ELBO.
This implies that the GW metric loss substantially affects the training procedure of learning representations.

\newcommand{\isometrywidth}{0.45}
\begin{figure}
    \centering
    \begin{minipage}[t]{\isometrywidth\linewidth}
        \centering
        \includegraphics[width=\linewidth,clip]{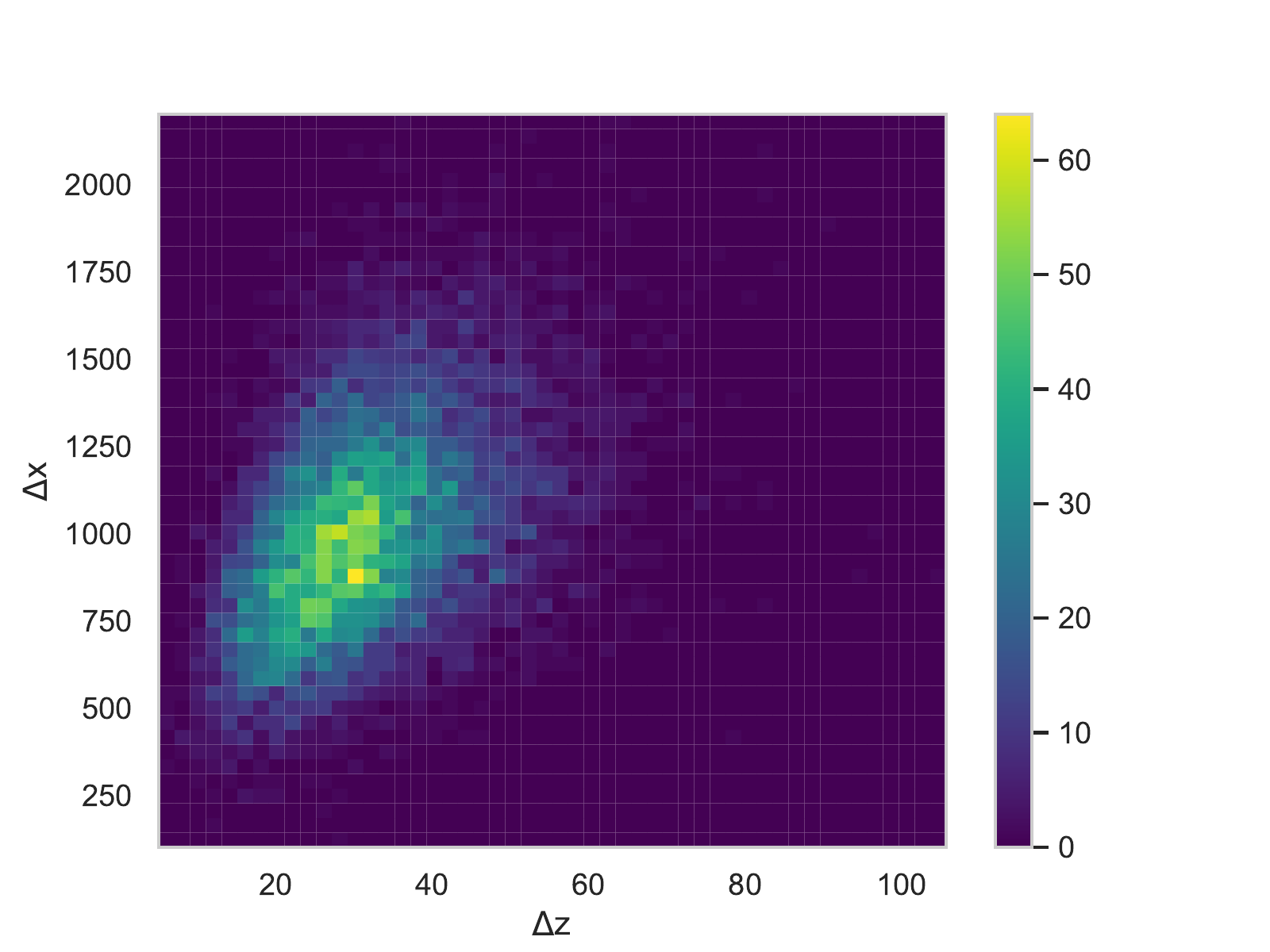}
        \subcaption{VAE~\cite{VAE}.}
    \end{minipage}
    \begin{minipage}[t]{\isometrywidth\linewidth}
        \centering
        \includegraphics[width=\linewidth,clip]{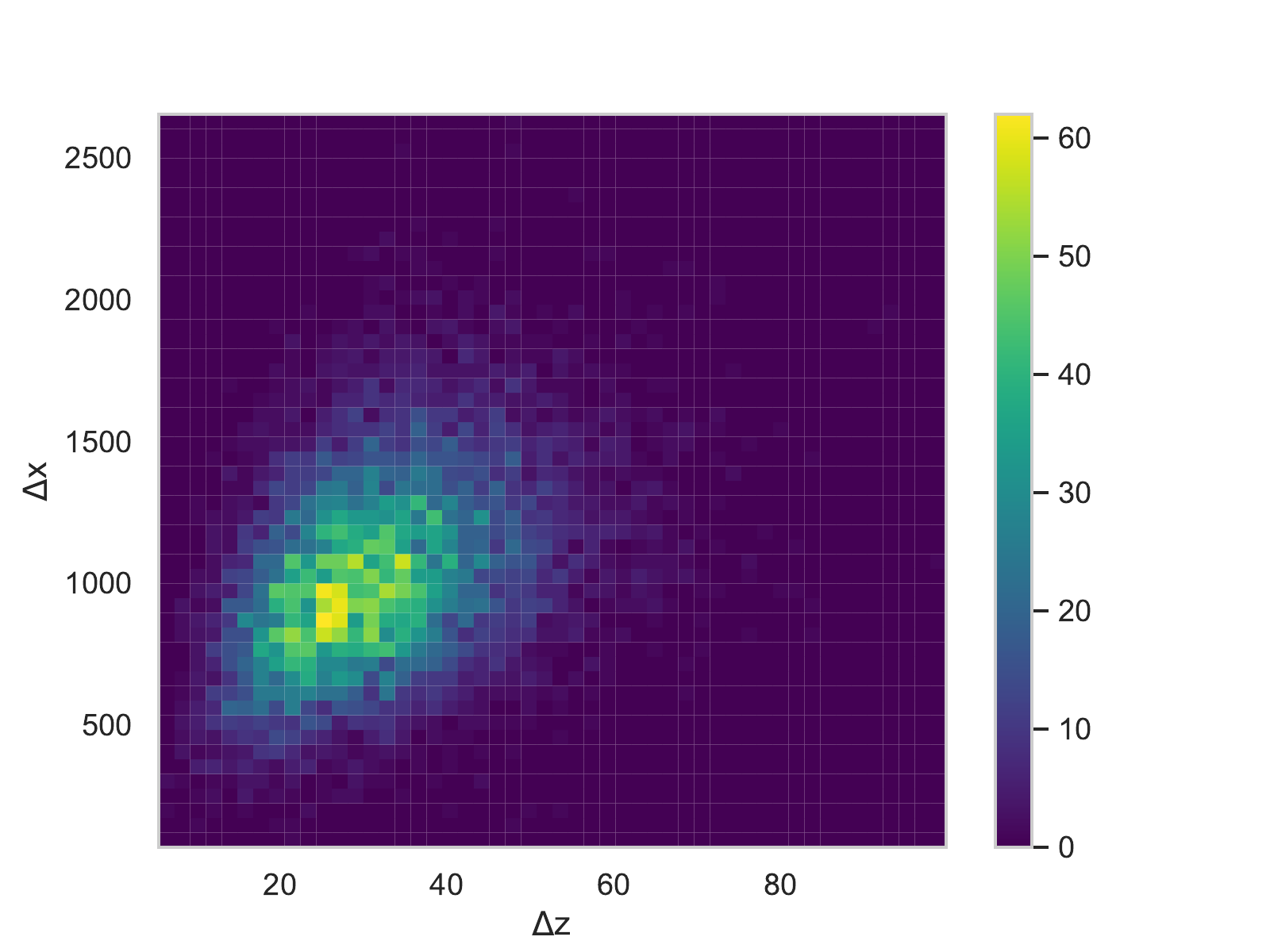}
        \subcaption{FactorVAE~\cite{FactorVAE}.}
    \end{minipage}
    \begin{minipage}[t]{\isometrywidth\linewidth}
        \centering
        \includegraphics[width=\linewidth,clip]{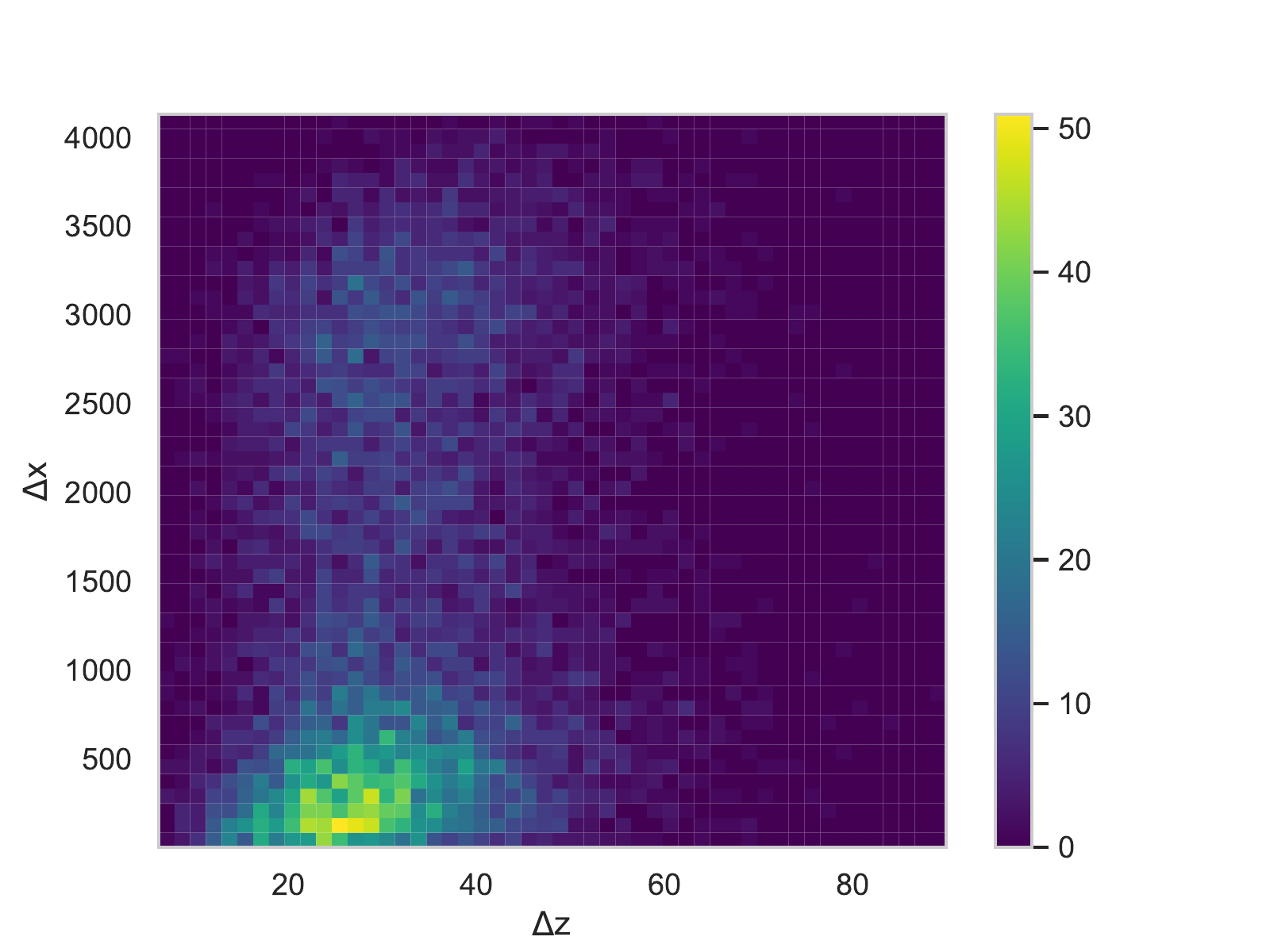}
        \subcaption{2-Stage VAE~\cite{Dai2019}.}
    \end{minipage}
    \begin{minipage}[t]{\isometrywidth\linewidth}
        \centering
        \includegraphics[width=\linewidth,clip]{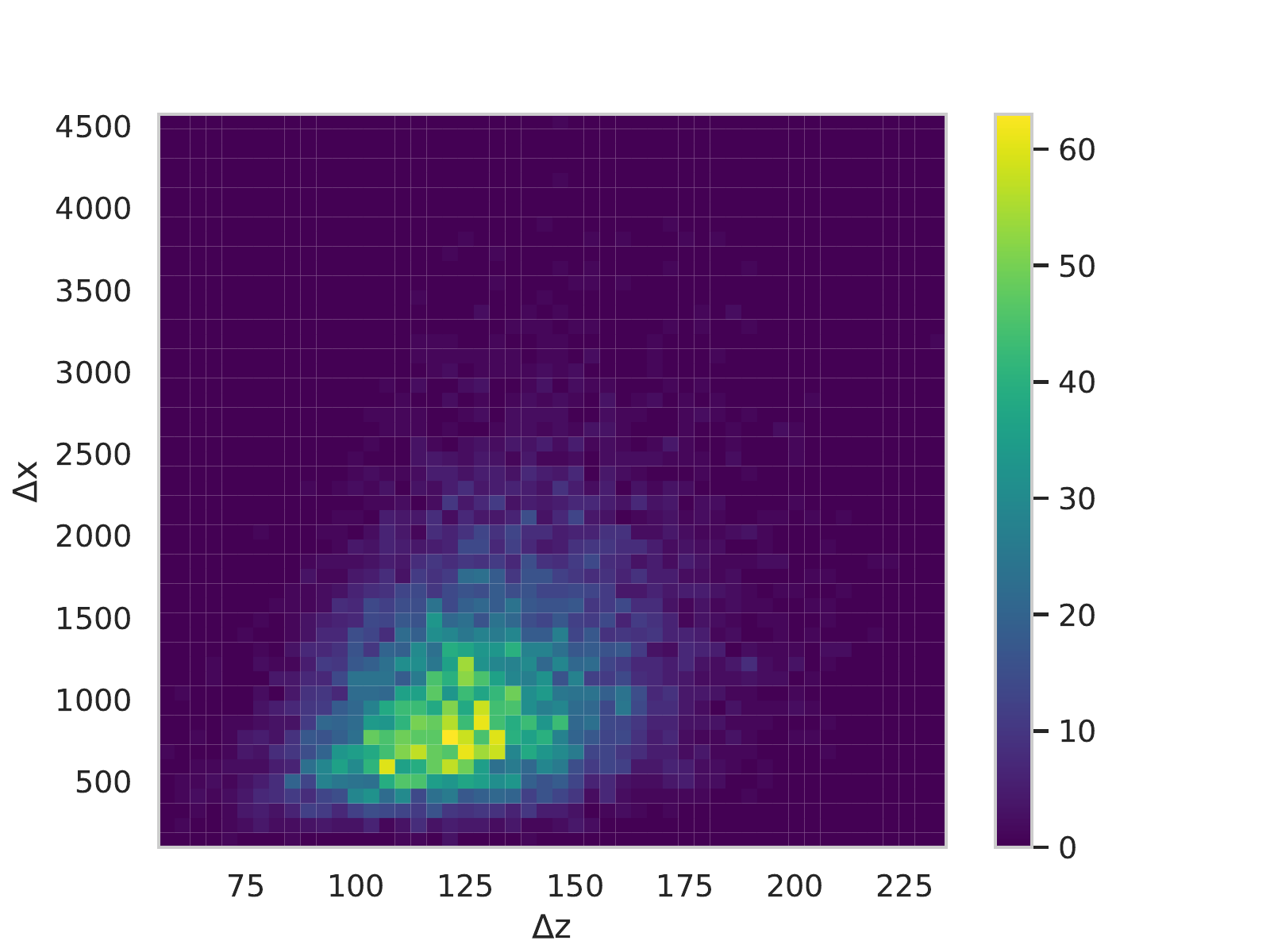}
        \subcaption{\revise{WAE~\cite{WAE}.}}
    \end{minipage}
    \begin{minipage}[t]{\isometrywidth\linewidth}
        \centering
        \includegraphics[width=\linewidth,clip]{metric_interspace_gwae.pdf}
        \subcaption{GWAE.}
    \end{minipage}
    \caption{Histograms of the differences in MNIST~\cite{MNIST}.
        Each histogram consists of 10,000 samples of $(\Delta x,\Delta z)$, where $\Delta x$~(vertical) and~$\Delta z$~(horizontal) respectively denote the differences~$\Delta x = \distx(\x,\x')$ and~$\Delta z = \distz(\z, \z')$ of two generative samples~$(\x,\z), (\x',\z') \sim \gjoint{\x}{\z}$.
        In all reported results including FactorVAE~\cite{FactorVAE}~($\gamma$=3)\revise{, WAE~\cite{WAE},} and GWAE~(NP, $\coeffd$=1, $\coeffw$=1, $\coeffh$=1), the latent dimension~$L$ was set to $L=16$, and their priors were set to the standard Gaussian.
    }
    \label{fig:isometry}
\end{figure}

\subsection{Training Process \label{sec:training-process}}
\revise{
We present the training process of GWAE in \cref{fig:training-process}.
Although the objective seems complex for its composition of four different losses, the training process successfully converged and the values of the terms $\lossgw$, $\lossw$, and $\lossd$ jointly descended in the most part of training.
Although the term $-\lossh$ increased, its values did not diverge to prevent the degenerate solutions.
These results imply that the three different losses~$\lossgw$, $\lossw$, and $\lossd$ did not conflict during the training process even for the complicated data, balancing these three terms against $-\lossh$ as in the trade-off of the reconstruction against the regularization in $\beta$-VAE~\cite{BetaVAE,Tschannnen2018}.
}
\newcommand{\trainingprocesswidth}{0.45}
\begin{figure}[!t]
    \centering
    \begin{minipage}[t]{\trainingprocesswidth\linewidth}
        \centering
        \includegraphics[width=\linewidth]{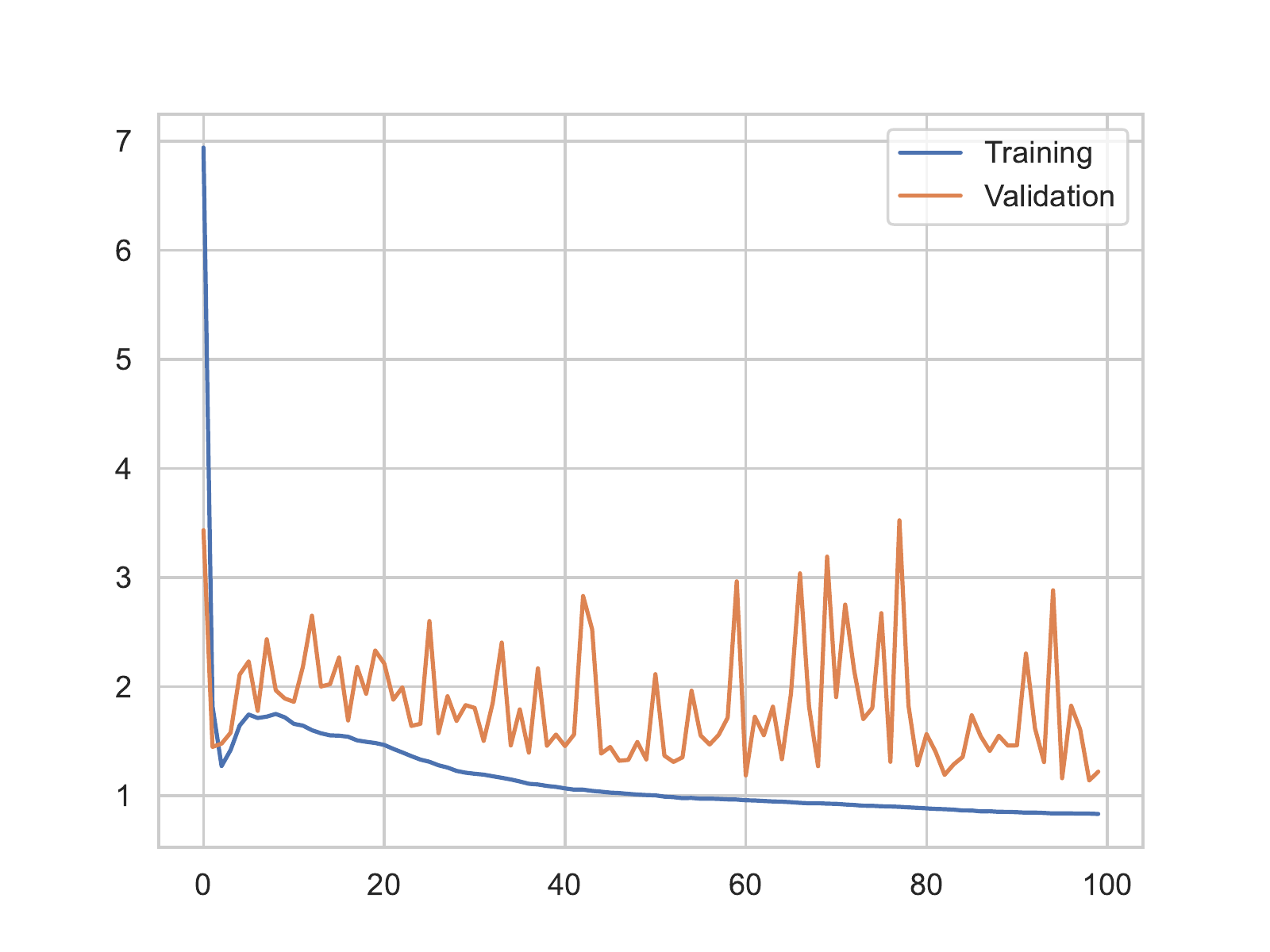}
        \subcaption{$\lossgw$, CelebA.}
    \end{minipage}
    \begin{minipage}[t]{\trainingprocesswidth\linewidth}
        \centering
        \includegraphics[width=\linewidth]{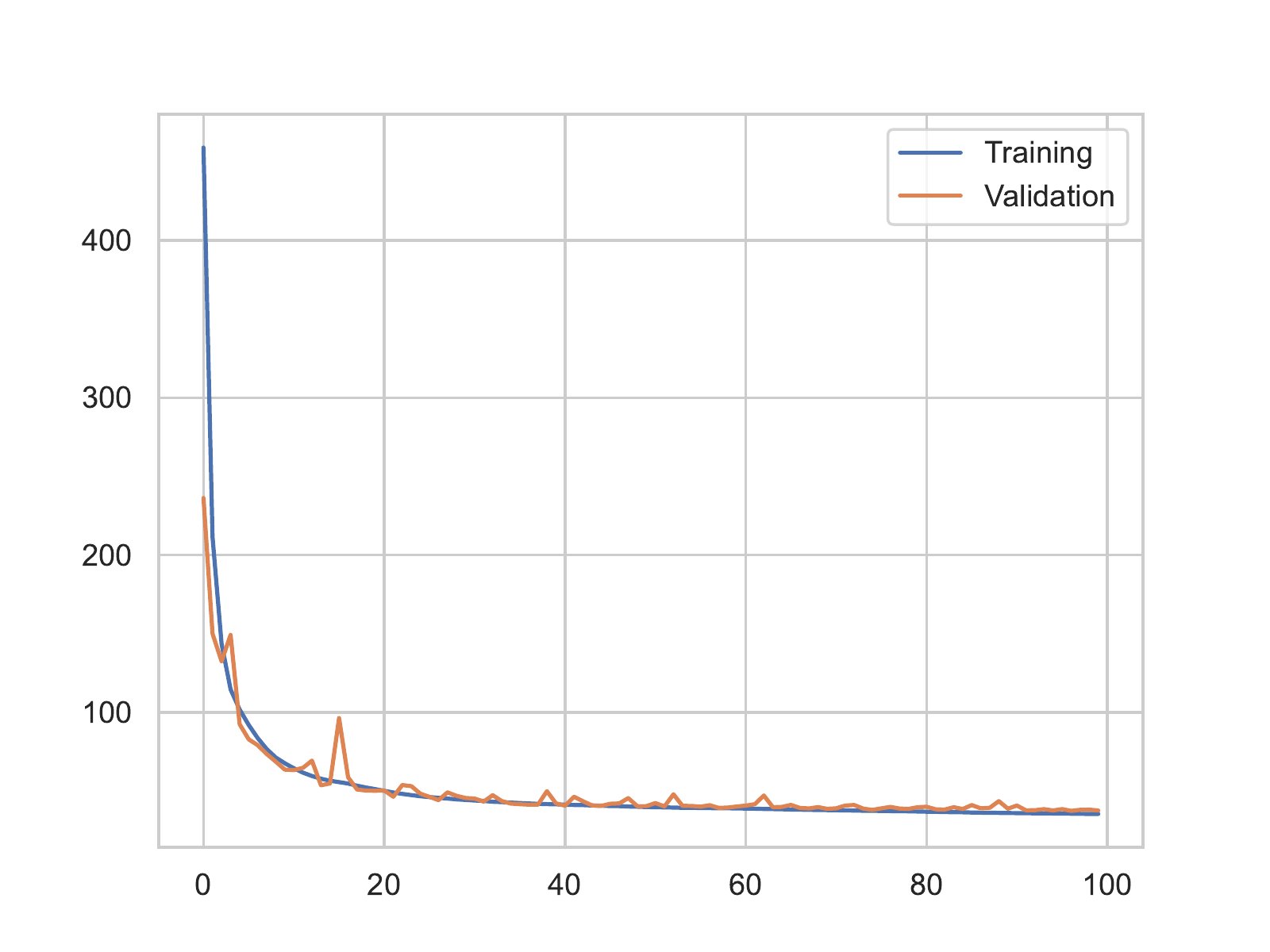}
        \subcaption{$\lossw$, CelebA.}
    \end{minipage}
    \begin{minipage}[t]{\trainingprocesswidth\linewidth}
        \centering
        \includegraphics[width=\linewidth]{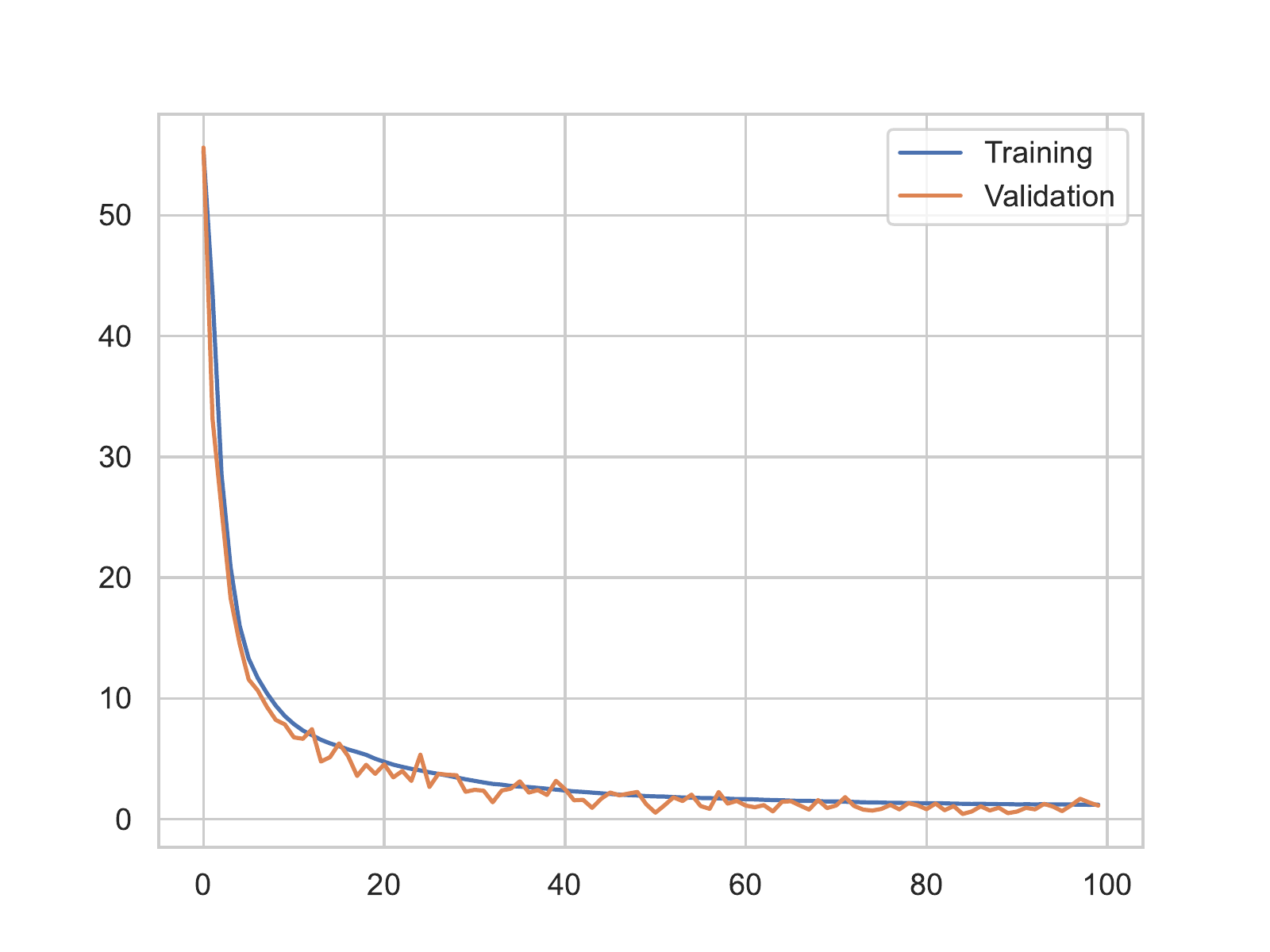}
        \subcaption{$\lossd$, CelebA.}
    \end{minipage}
    \begin{minipage}[t]{\trainingprocesswidth\linewidth}
        \centering
        \includegraphics[width=\linewidth]{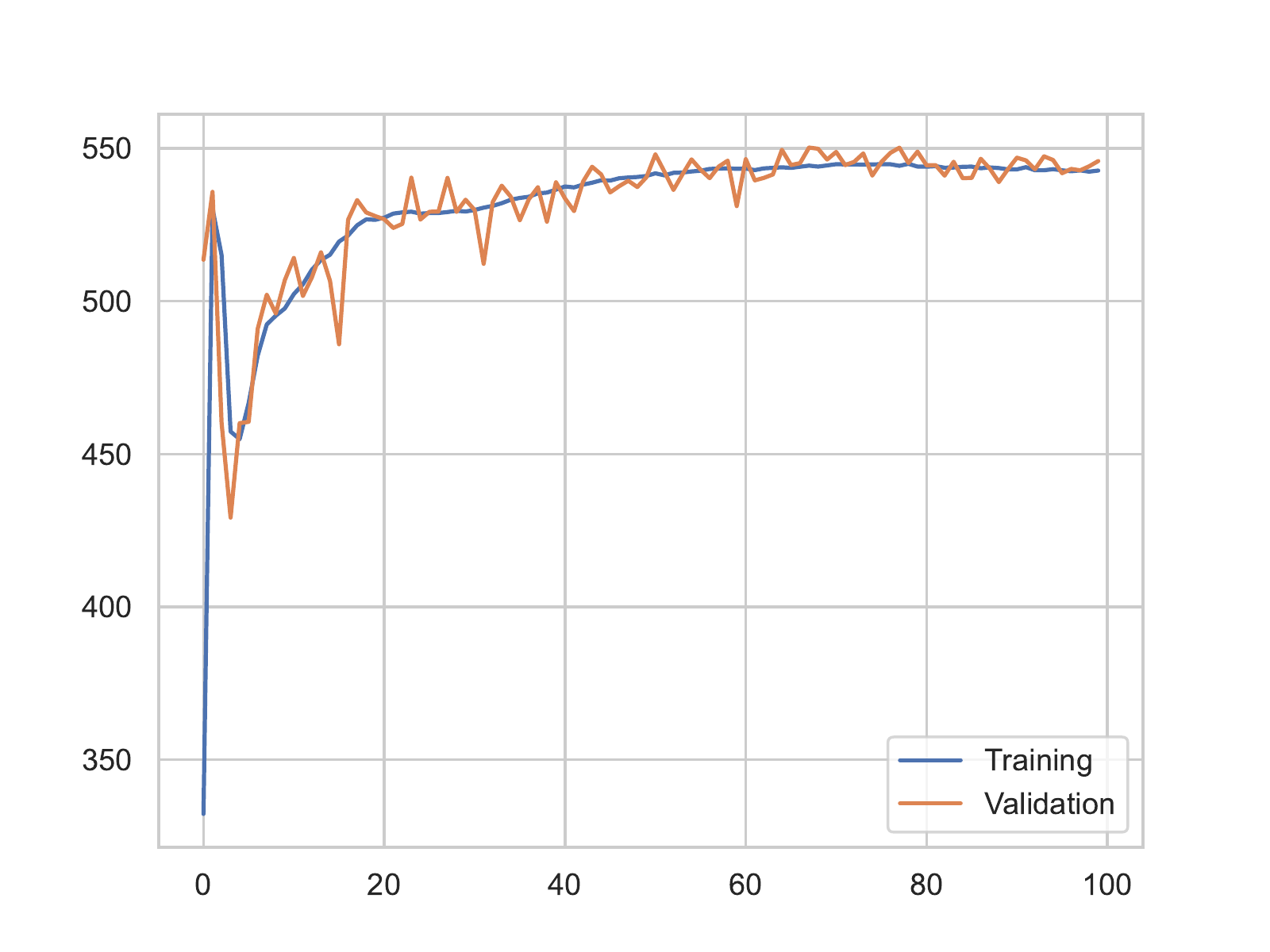}
        \subcaption{$-\lossh$, CelebA.}
    \end{minipage}
    \begin{minipage}[t]{\trainingprocesswidth\linewidth}
        \centering
        \includegraphics[width=\linewidth]{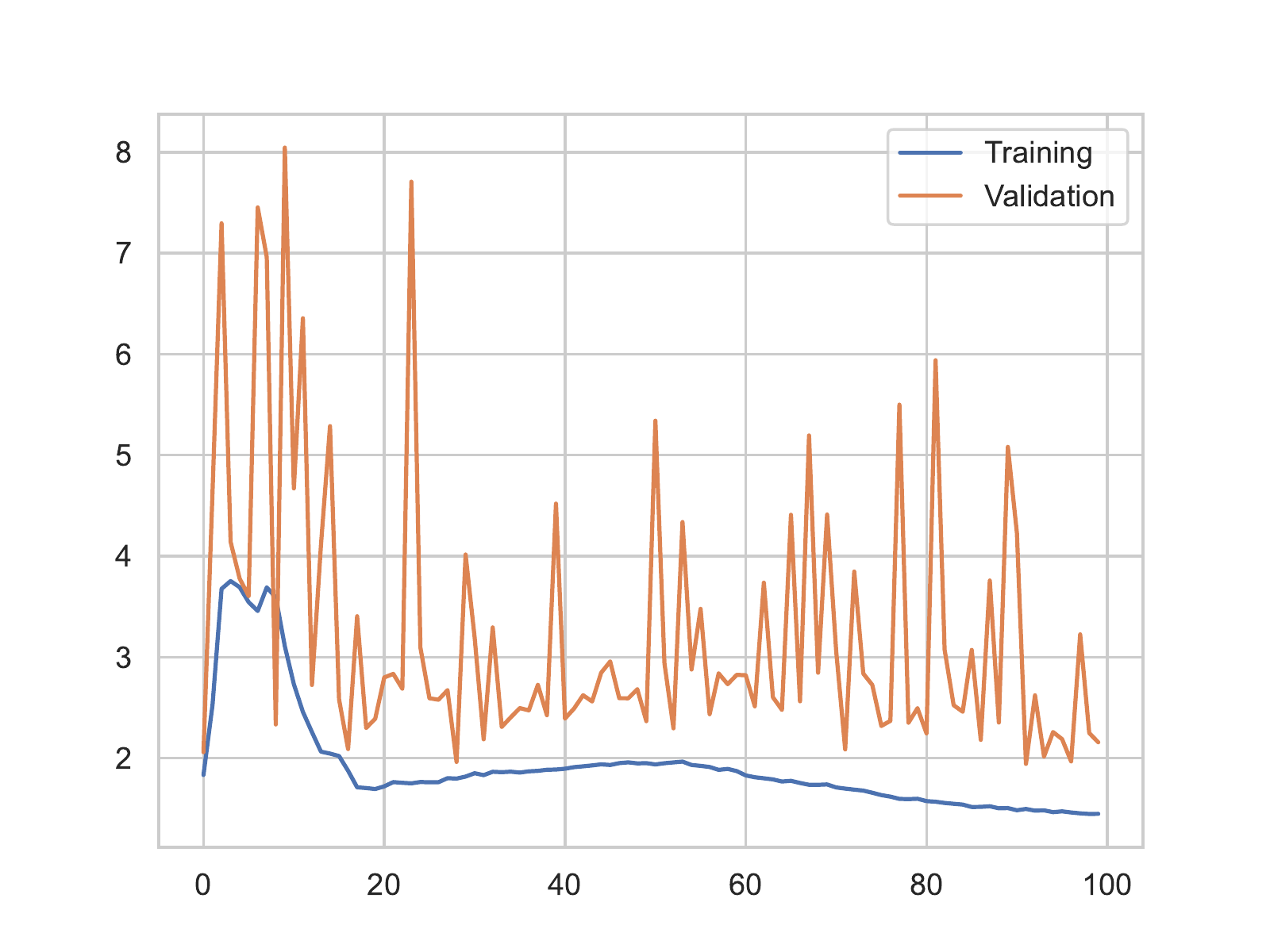}
        \subcaption{$\lossgw$, CIFAR-10.}
    \end{minipage}
    \begin{minipage}[t]{\trainingprocesswidth\linewidth}
        \centering
        \includegraphics[width=\linewidth]{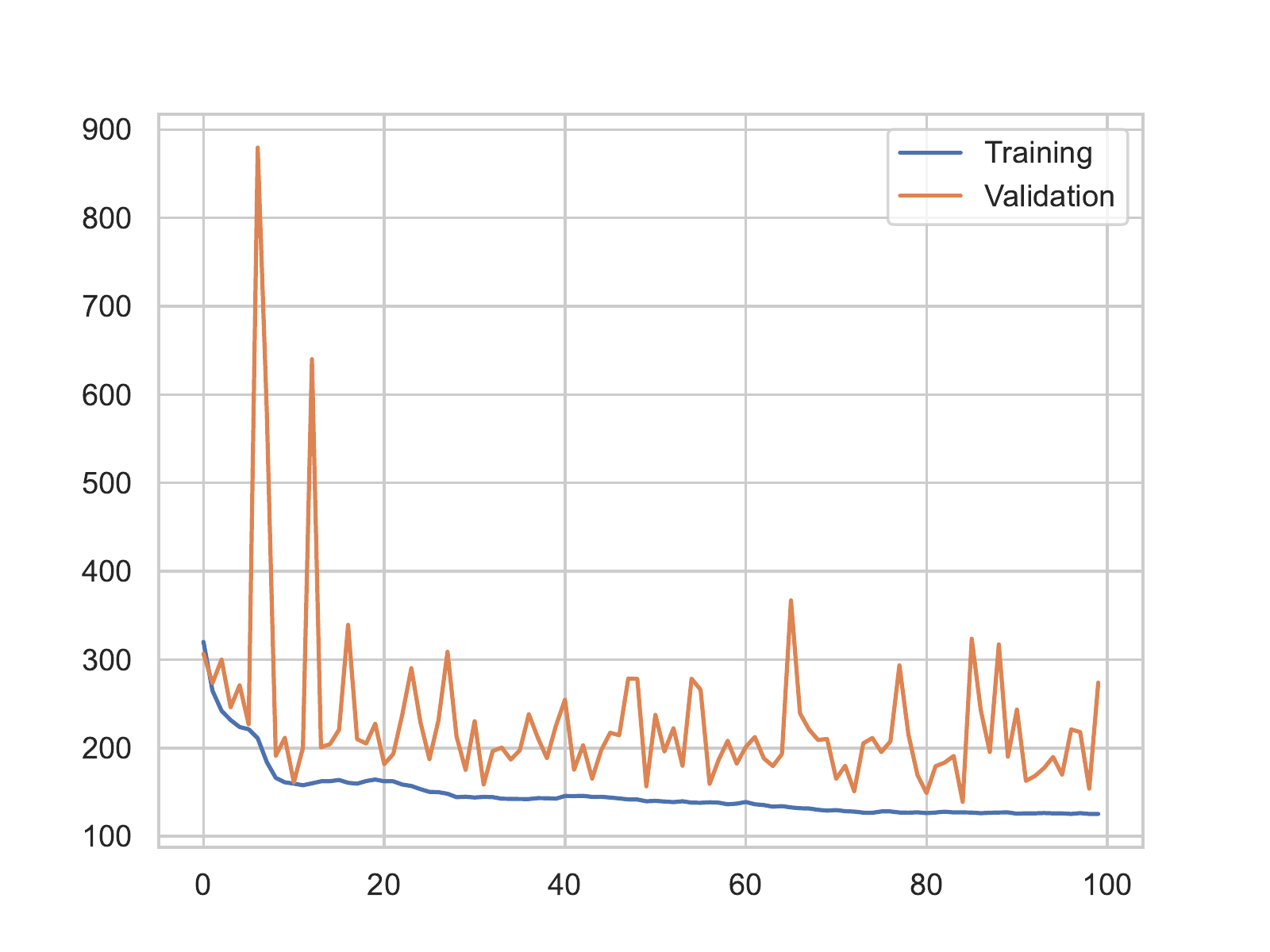}
        \subcaption{$\lossw$, CIFAR-10.}
    \end{minipage}
    \begin{minipage}[t]{\trainingprocesswidth\linewidth}
        \centering
        \includegraphics[width=\linewidth]{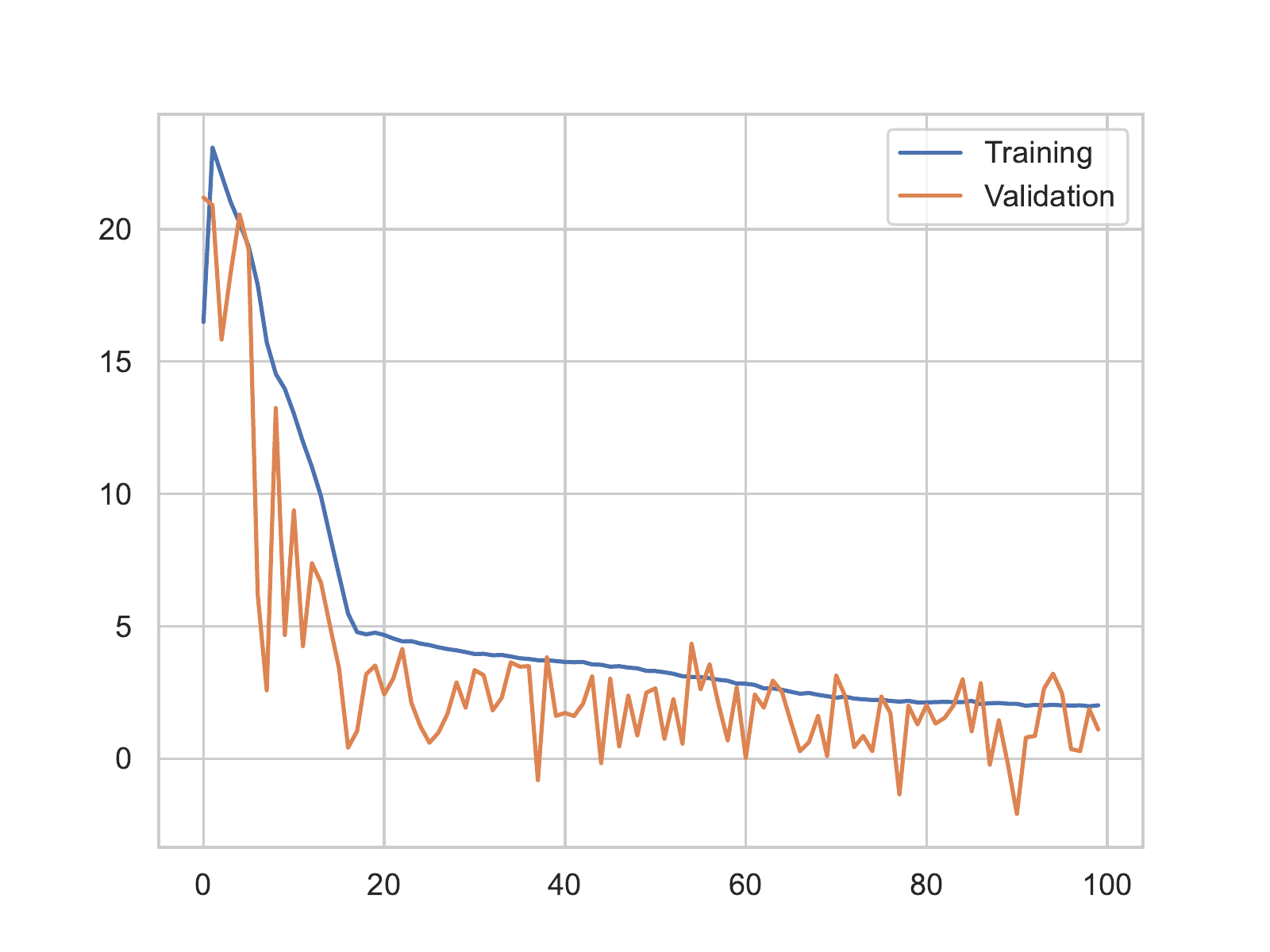}
        \subcaption{$\lossd$, CIFAR-10.}
    \end{minipage}
    \begin{minipage}[t]{\trainingprocesswidth\linewidth}
        \centering
        \includegraphics[width=\linewidth]{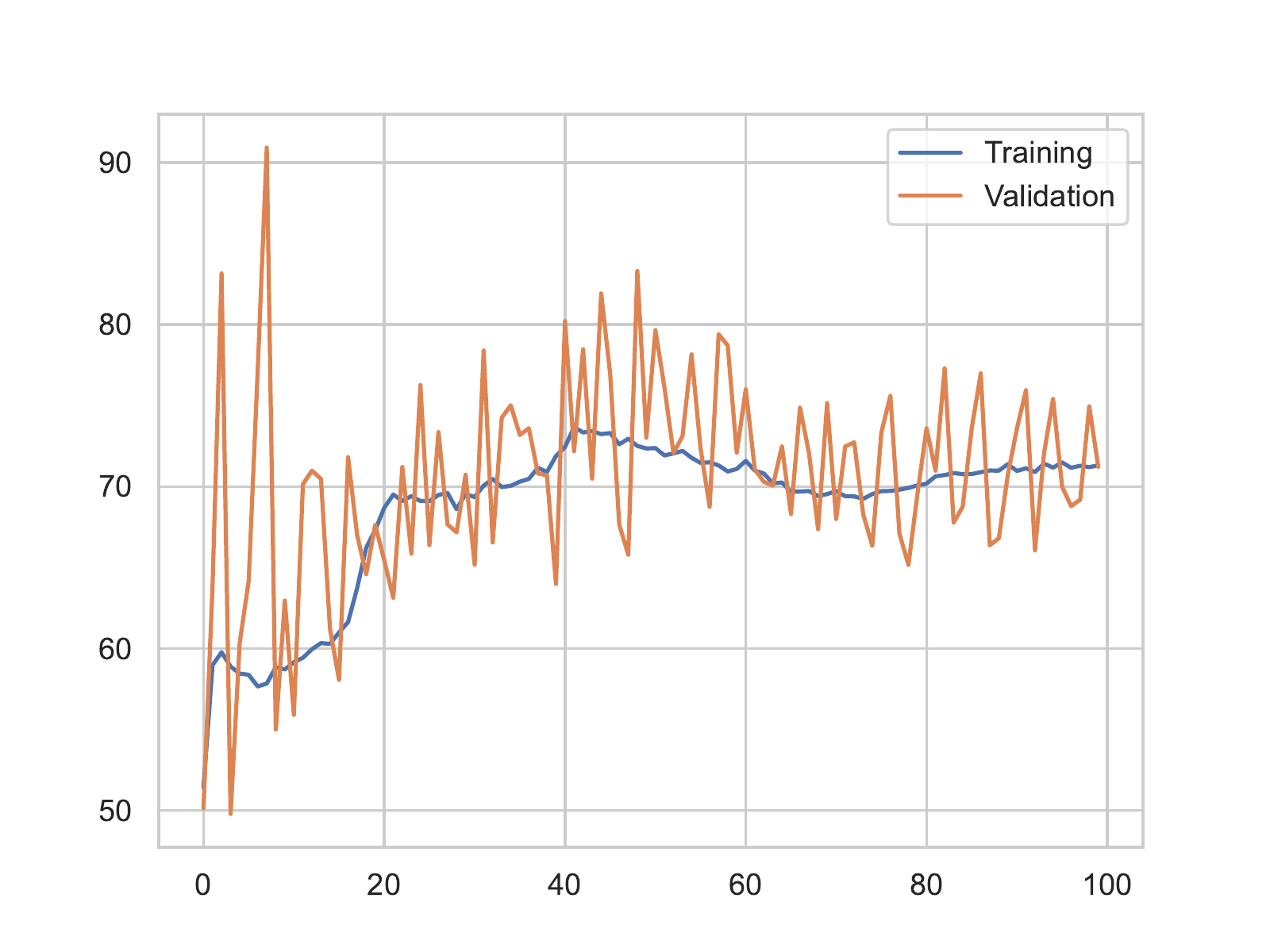}
        \subcaption{$-\lossh$, CIFAR-10.}
    \end{minipage}
    \caption{\revise{The training process of a GWAE model. The model is trained using NP and $\coeffh=\coeffw=\coeffh=1$. The plot (a)--(d) are training curves during one trial of training using CelebA~\cite{CelebA}, and (e)--(h) are using CIFAR-10~\cite{CIFAR10}. In each plot, the horizontal axis represents the number of epochs elapsed, and the vertical axis expresses the loss value. The blue and orange curves represent the training and validation losses, respectively.}}
    \label{fig:training-process}
\end{figure}

\subsection{Prior Family Selection \label{sec:meta-prior-selection}}
\revise{
We show the effect of prior family selection regarding a meta-prior, disentanglement, in \cref{tab:prior-family-selection}.
While GWAE models with the NP and GMP retain the informativeness of the FNP, the other two priors than FNP did not comparably disentangle the latent factors.
Although the NP covers a more general family of prior, these results suggest that choosing a prior family suitable to the postulated meta-prior greatly facilitates learning representations.
}

\begin{table}[!t]
    \centering
    \caption{\revise{The effect of prior family selection in the GWAE model. The same settings in \cref{tab:disentanglement} are applied to all the reported models.}}
    \label{tab:prior-family-selection}
    \begin{tabular}{lccc}
        \toprule
        Model & DCI-C $\uparrow$ & DCI-D $\uparrow$ & DCI-I $\uparrow$ \\
        \midrule
        GWAE~(NP)  & 0.3966 & 0.3113 & 0.9403 \\
        GWAE~(FNP) & \textbf{0.9080} & \textbf{0.7024} & \textbf{0.9966} \\
        GWAE~(GMP) & 0.4247 & 0.4373 & 0.9655 \\
        \bottomrule
    \end{tabular}
\end{table}


\subsection{Qualitative Evaluations of Generation and Reconstruction \label{sec:appendix-qualitative}}

We show the reconstructed images by GWAE and state-of-the-art variational autoencoding methods in \cref{fig:appendix-reconstruction}.
The shown images are the first ten samples of the test split in the CelebA~\cite{CelebA} dataset under the latent size $L=64$.
Compared with the other methods, the reconstruction of the GWAE model tends to retain edges~(see the bottom rows of \cref{fig:appendix-reconstruction}), while VAE-based models generate smooth, blurry images due to the noise injected in the latent space to perform probabilistic modeling and manifold learning.
We also show the reconstruction results of MNIST~\cite{MNIST} in \cref{fig:appendix-reconstruction-mnist} and CIFAR-10~\cite{CIFAR10} in \cref{fig:appendix-reconstruction-cifar10}.
These results support that the GWAE models consistently perform autoencoding also in a more simple dataset (MNIST).
In a more complex dataset (CIFAR-10), the GWAE model attained the best evaluation in generation albeit its reconstruction, suggesting that the GWAE model successfully captured the abstract structure of data rather than reconstructed the given images.
This difference highlights the difference in their objectives, \ie, the GW objective aims at distribution matching in the latent space, while the $\beta$-VAE~\cite{BetaVAE} objective with $\beta<1$ puts weight on reconstruction.

We further study the generated images by GWAE and state-of-the-art VAE-based generative models in \cref{fig:appendix-generation}.
These qualitative results show that the GWAE generation successfully obtains a diverse set of images compared with those of state-of-the-art autoencoding generative models.
The ALI model~\cite{ALI} (\cref{fig:appendix-generation}~(a)) also generates various images by the distribution matching of bidirectional models, but the generated images have wavy contours, failing at composing images with a consistent appearance owing to the lack of an autoencoding process.
Although the VAE-GAN model~\cite{VAEGAN} (\cref{fig:appendix-generation}~(b)) adequately yields organized images with smooth textures, the azimuth of these images is less diverse, \ie, the great majority of the images are facing forward or looking slightly sideways.
The images generated by 2-Stage VAE~\cite{Dai2019} (\cref{fig:appendix-generation}~(c)) have diverse azimuth, color, and background; however, these images tend to incline toward the majority attributes, \eg, not wearing eyeglasses or sunglasses.
The GWAE model~(\cref{fig:appendix-generation}~(d)) successfully generates facial images with various skin colors, diversified backgrounds, and assorted facial expressions~(\eg, wearing a mustache).
These results imply that the GWAE models also function as generative models while it has been built as a representation learning method owing to the collateral condition~$\gjoint{\x}{\z} \approx \ijoint{\x}{\z}$ in \cref{eq:loss-d} and the generative modeling~$\gmodel{\x} \approx \empir{\x}$ as its necessary condition.

\newcommand{\reconstructionminipagewidth}{0.8}
\newcommand{\reconstructionskip}{0.5em}
\newcommand{\generationminipagewidth}{0.49}

\begin{figure}[!t]
    \centering
    \begin{minipage}[t]{\reconstructionminipagewidth\linewidth}
        \centering
        \includegraphics[width=\linewidth]{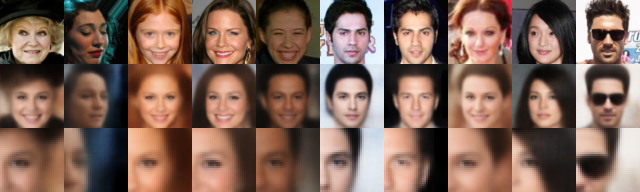}
        \subcaption{VAE.}
        \vspace{\reconstructionskip}
    \end{minipage}
    \begin{minipage}[t]{\reconstructionminipagewidth\linewidth}
        \centering
        \includegraphics[width=\linewidth]{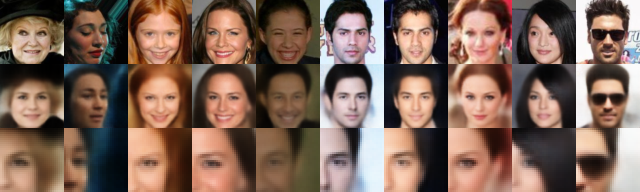}
        \subcaption{$\beta$-VAE~\cite{BetaVAE} ($\beta$=0.1).}
        \vspace{\reconstructionskip}
    \end{minipage}
    \begin{minipage}[t]{\reconstructionminipagewidth\linewidth}
        \centering
        \includegraphics[width=\linewidth]{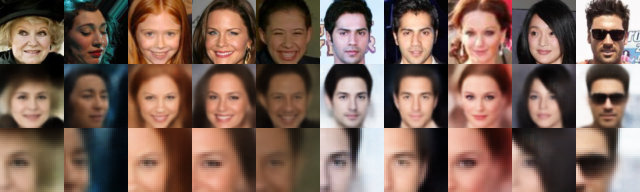}
        \subcaption{AVB~\cite{AVB} ($\beta$=1).}
        \vspace{\reconstructionskip}
    \end{minipage}
    \begin{minipage}[t]{\reconstructionminipagewidth\linewidth}
        \centering
        \includegraphics[width=\linewidth]{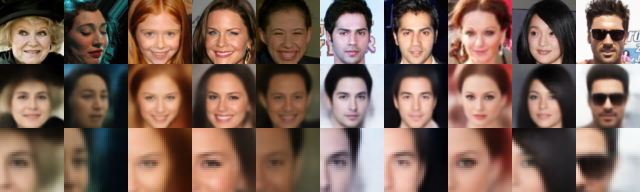}
        \subcaption{WAE~($\lambda$=100).}
        \vspace{\reconstructionskip}
    \end{minipage}
    \begin{minipage}[t]{\reconstructionminipagewidth\linewidth}
        \centering
        \includegraphics[width=\linewidth]{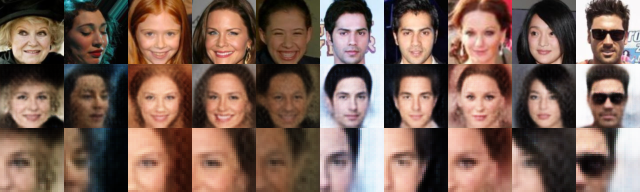}
        \subcaption{GWAE~(NP, $\coeffd$=1, $\coeffw$=10, $\coeffh$=0.0001).}
    \end{minipage}
    \caption{
        Reconstructed images in CelebA~\cite{CelebA}.
        The images denote original data samples~(top rows), reconstructed images~(middle rows), and zoomed reconstructions~(bottom rows).
        Each column corresponds to one data instance in the test set.
    }
    \label{fig:appendix-reconstruction}
\end{figure}

\begin{figure}[!t]
    \centering
    \begin{minipage}[t]{\reconstructionminipagewidth\linewidth}
        \centering
        \includegraphics[width=\linewidth]{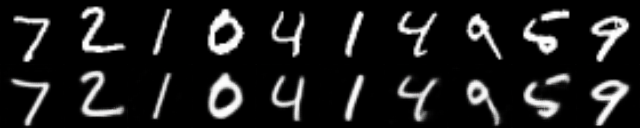}
        \subcaption{VAE. \revise{FID: 16.8, PSNR: 23.66 dB.}}
        \vspace{\reconstructionskip}
    \end{minipage}
    \begin{minipage}[t]{\reconstructionminipagewidth\linewidth}
        \centering
        \includegraphics[width=\linewidth]{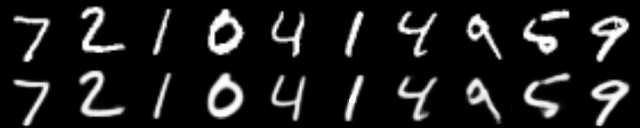}
        \subcaption{$\beta$-VAE~\cite{BetaVAE} ($\beta$=0.1). \revise{FID: 15.5, PSNR: 25.45 dB.}}
        \vspace{\reconstructionskip}
    \end{minipage}
    \begin{minipage}[t]{\reconstructionminipagewidth\linewidth}
        \centering
        \includegraphics[width=\linewidth]{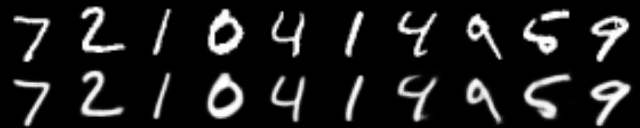}
        \subcaption{AVB~\cite{AVB} ($\beta$=1). \revise{FID: 39.2, PSNR: 24.31 dB.}}
        \vspace{\reconstructionskip}
    \end{minipage}
    \begin{minipage}[t]{\reconstructionminipagewidth\linewidth}
        \centering
        \includegraphics[width=\linewidth]{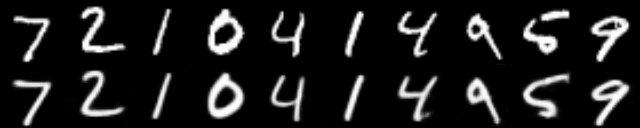}
        \subcaption{WAE~($\lambda$=100). \revise{FID: 16.9, PSNR: 25.28 dB.}}
        \vspace{\reconstructionskip}
    \end{minipage}
    \begin{minipage}[t]{\reconstructionminipagewidth\linewidth}
        \centering
        \includegraphics[width=\linewidth]{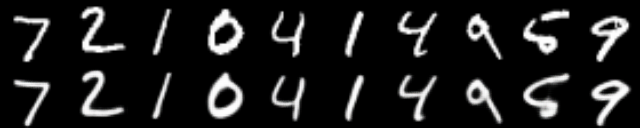}
        \subcaption{GWAE~(NP, $\coeffd$=1, $\coeffw$=10, $\coeffh$=0.0001). \revise{FID: 14.4, PSNR: 26.11 dB.}}
    \end{minipage}
    \caption{
        \revise{
        Reconstructed images in MNIST~\cite{MNIST}.
        The images denote original data samples~(top rows), reconstructed images~(bottom rows).
        Each column corresponds to one data instance in the test set.
        }
    }
    \label{fig:appendix-reconstruction-mnist}
\end{figure}

\begin{figure}[!t]
    \centering
    \begin{minipage}[t]{\reconstructionminipagewidth\linewidth}
        \centering
        \includegraphics[width=\linewidth]{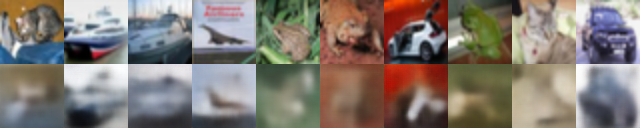}
        \subcaption{VAE. \revise{FID: 111.3, PSNR: 19.84 dB.}}
        \vspace{\reconstructionskip}
    \end{minipage}
    \begin{minipage}[t]{\reconstructionminipagewidth\linewidth}
        \centering
        \includegraphics[width=\linewidth]{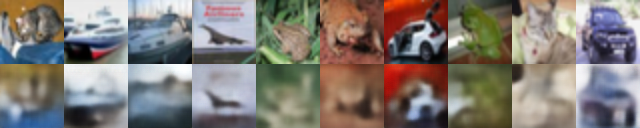}
        \subcaption{$\beta$-VAE~\cite{BetaVAE} ($\beta$=0.1). \revise{FID: 84.5, PSNR: 22.48 dB.}}
        \vspace{\reconstructionskip}
    \end{minipage}
    \begin{minipage}[t]{\reconstructionminipagewidth\linewidth}
        \centering
        \includegraphics[width=\linewidth]{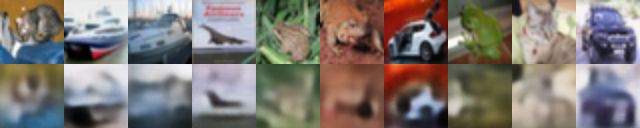}
        \subcaption{AVB~\cite{AVB} ($\beta$=1). \revise{FID: 109.9, PSNR: 21.14 dB.}}
        \vspace{\reconstructionskip}
    \end{minipage}
    \begin{minipage}[t]{\reconstructionminipagewidth\linewidth}
        \centering
        \includegraphics[width=\linewidth]{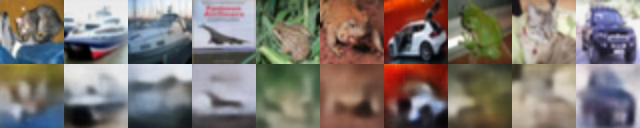}
        \subcaption{WAE~($\lambda$=100). \revise{FID: 87.3, PSNR: 22.45 dB.}}
        \vspace{\reconstructionskip}
    \end{minipage}
    \begin{minipage}[t]{\reconstructionminipagewidth\linewidth}
        \centering
        \includegraphics[width=\linewidth]{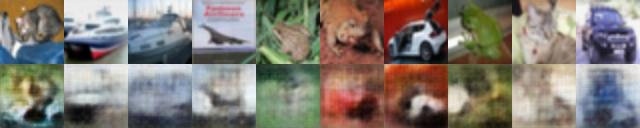}
        \subcaption{GWAE~(NP, $\coeffd$=1, $\coeffw$=10, $\coeffh$=0.0001). \revise{FID: 59.9, PSNR: 17.64 dB.}}
    \end{minipage}
    \caption{
        \revise{
        Reconstructed images in CIFAR-10~\cite{CIFAR10}.
        The images denote original data samples~(top rows), reconstructed images~(bottom rows).
        Each column corresponds to one data instance in the test set.
        }
    }
    \label{fig:appendix-reconstruction-cifar10}
\end{figure}

\begin{figure}[!t]
    \centering
    \begin{minipage}[t]{\generationminipagewidth\linewidth}
        \centering
        \includegraphics[width=\linewidth]{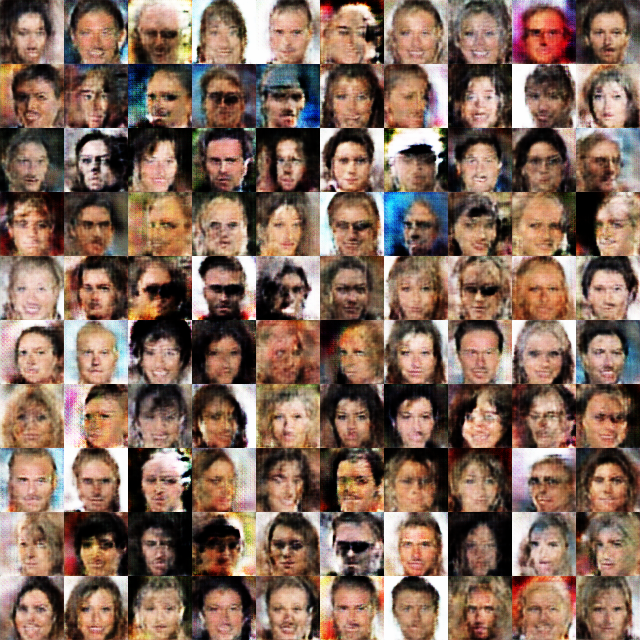}
        \subcaption{ALI~\cite{ALI}.}
    \end{minipage}
    \begin{minipage}[t]{\generationminipagewidth\linewidth}
        \centering
        \includegraphics[width=\linewidth]{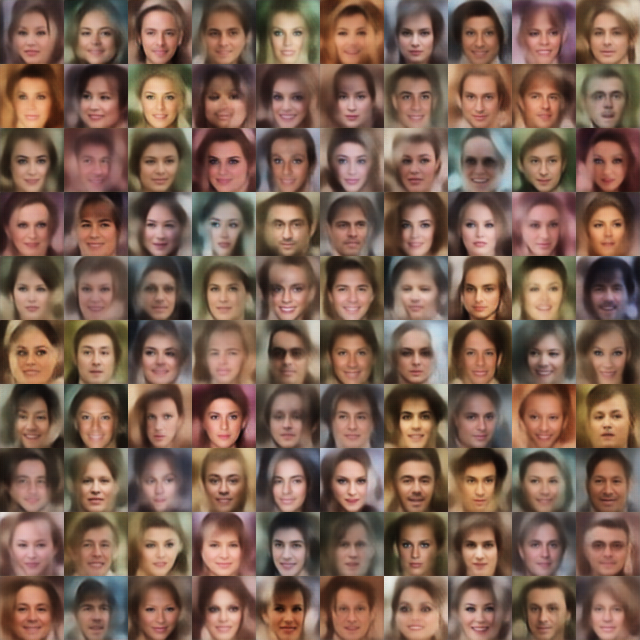}
        \subcaption{VAE-GAN~\cite{VAEGAN}~($\gamma$=1).}
    \end{minipage}
    \begin{minipage}[t]{\generationminipagewidth\linewidth}
        \centering
        \includegraphics[width=\linewidth]{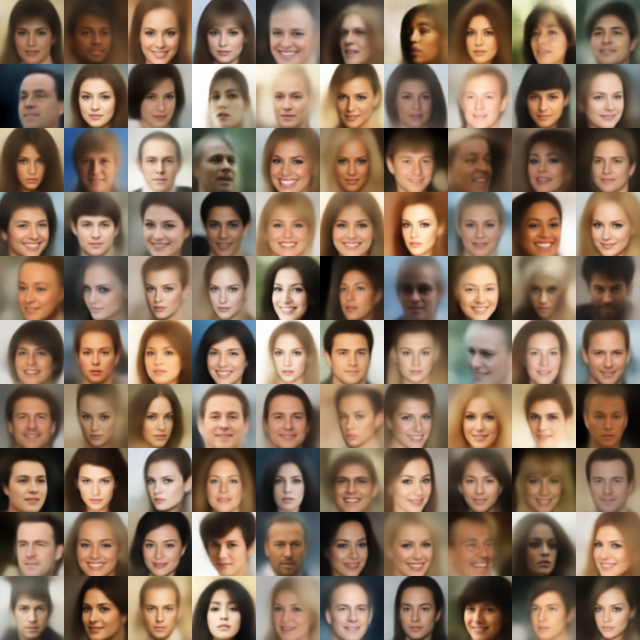}
        \subcaption{2-Stage VAE~\cite{Dai2019}.}
    \end{minipage}
    \begin{minipage}[t]{\generationminipagewidth\linewidth}
        \centering
        \includegraphics[width=\linewidth]{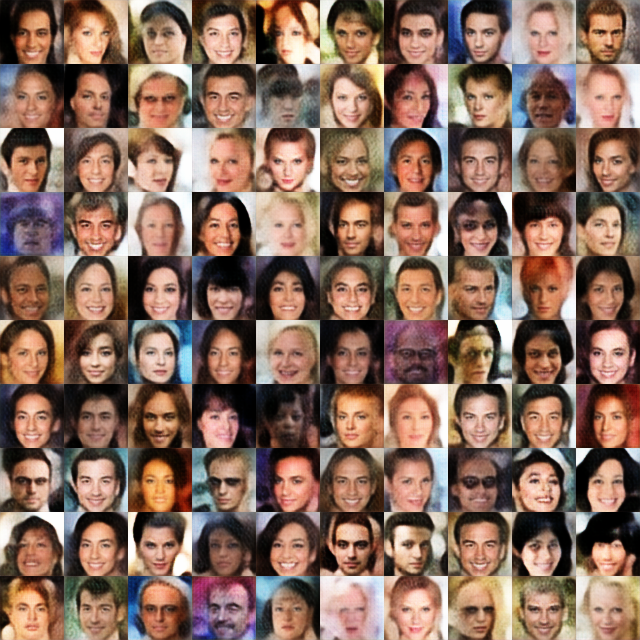}
        \subcaption{GWAE (NP, $\coeffd$=1, $\coeffw$=10, $\coeffh$=0.0001).}
    \end{minipage}
    \caption{
        Generated images in CelebA~\cite{CelebA}.
        We show 100 images sampled from the generative model~$\gmodel{\x}$ without conducting cherry-picking.
    }
    \label{fig:appendix-generation}
\end{figure}

\subsection{Ablation Study \label{sec:ablation-study}}

\begin{table}[!t]
    \centering
    \caption{\revise{The ablation study on generation and reconstruction in CelebA~\cite{CelebA}. The same settings as \cref{tab:fid} are applied in these experiments.}}
    \label{tab:ablation-study}
    \begin{tabular}{lcc}
        \toprule
        Model & FID~\cite{FID} $\downarrow$ & PSNR [dB] $\uparrow$ \\
        \midrule
        GWAE~(NP) & \textbf{45.3} & \textbf{22.82} \\
        GWAE~(NP) w/o $\lossw$ & 233.7 & 9.80 \\
        GWAE~(NP) w/o $\lossd$ & 403.8 & 18.63 \\
        GWAE~(NP) w/ MMD $\lossd$ & 158.4 & 22.61 \\
        GWAE~(NP) w/ $\zsp$-only critic & 102.4 & 22.41 \\
        GWAE~(NP) w/o $\lossh$ & 179.6 & 21.57 \\
        GWAE~(NP, $\gworder=\worder$) & 123.5 & 16.03 \\
        \bottomrule
    \end{tabular}
\end{table}

\begin{figure}[!t]
    \centering
    \includegraphics[width=\linewidth,clip]{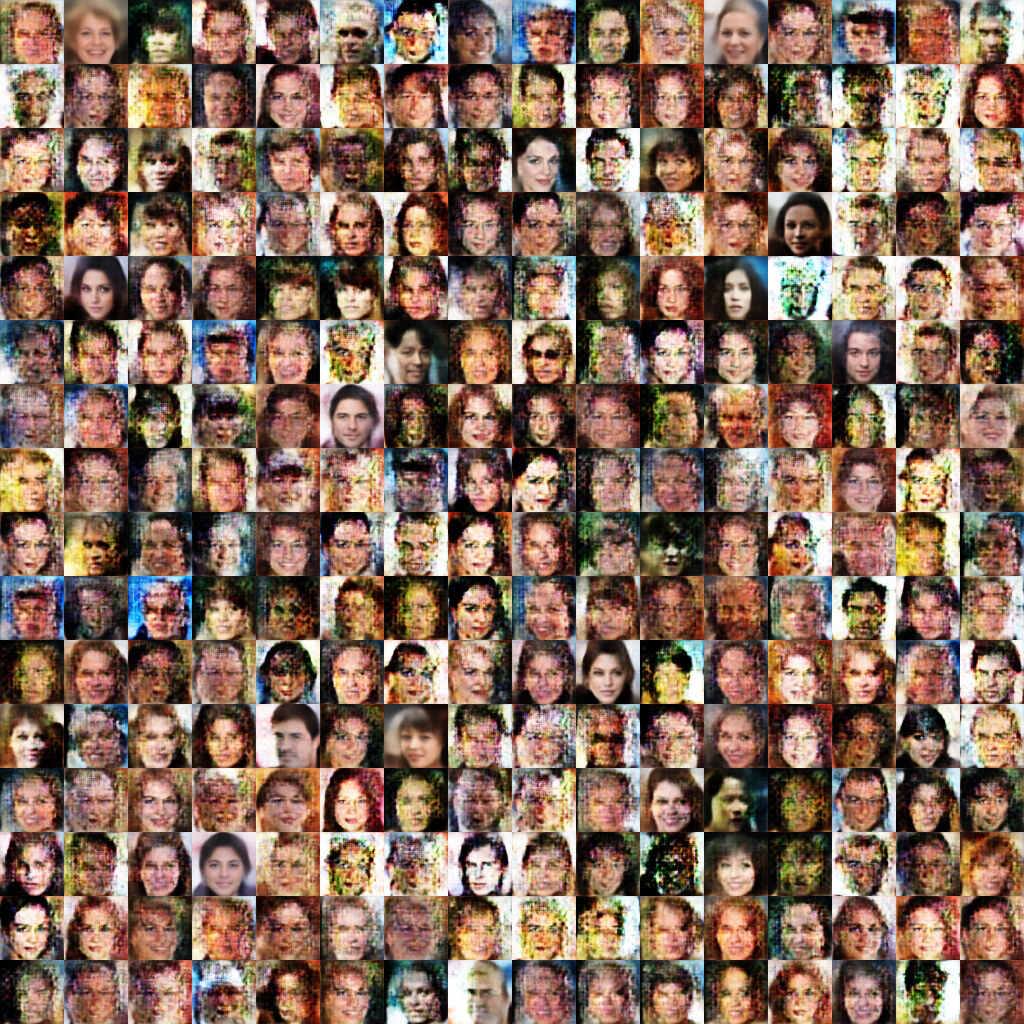}
    \caption{Generated images in CelebA~\cite{CelebA} using the GWAE model without the regularization term~$\lossh$.}
    \label{fig:generation-ablation}
\end{figure}

\begin{figure}[!t]
    \centering
    \includegraphics[width=\linewidth,clip]{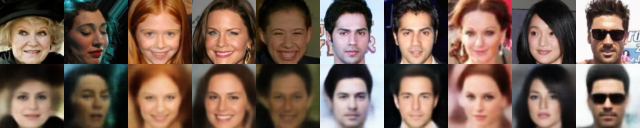}
    \caption{\revise{Reconstructed images in CelebA~\cite{CelebA} using the GWAE model without the regularization term~$\lossh$. Each column corresponds to one test data instance. The rows denote original~(top) and reconstructed~(bottom) images.}}
    \label{fig:reconstruction-ablation}
\end{figure}

\revise{We conducted the ablation study of the losses and regularizations introduced in \cref{eq:total-loss}.
\cref{tab:ablation-study} shows the results of the ablation study of the three sub-constraints~$\lossw$, $\lossd$, and~$\lossh$.
The ablations yielded the performance degradation of GWAE, especially in~$\lossw$.
These results suggest the necessity of each regularization term and reveal their roles in representation learning.

\textbf{Ablation of~$\lossw$.} The ablation of the term~$\lossw$ brought low-quality reconstruction, which suggests that~$\lossw$ works as the autoencoding constraint as can be seen from taking the reconstruction loss in~$\lossw$.
It also reduced generation capability as well as reconstruction, suggesting that the generative modeling via autoencoding is inherited from the variational autoencoding architecture of VAEs~\cite{VAE}.

\textbf{Ablation of~$\lossd$.} Without the term~$\lossd$, the GWAE models suffer from the lack of distribution matching in data generation, while it successfully conducted data reconstruction.
These phenomena could be caused by the discrepancy between the encoded latent distribution~$\iagg{\z}$ and the prior~$\trainablePrior{\z}$.
Similar results are also obtained in the ablation of the merged sufficient condition~(see \cref{eq:loss-d}) for the regularization~$\lossd$, where $\lossd$ is defined as the MMD loss between the prior~$\trainablePrior{\z}$ and the encoded latent~$\iagg{\z}$, as in the WAE-MMD model~\cite{WAE}.
This choice of $\lossd$ on the low-dimensional space~$\zsp$ appears to be a replacement for the Kantorovich potential adversarially learned in the high-dimensional joint space~$\xsp\times\zsp$; however, lacking the merged sufficient condition seems to have caused the crucial reduction of generation performance as in the gross ablation of $\lossd$.
These results imply that the term~$\lossd$ with adversarial learning regularizes the generative model~$\gjoint{\x}{\z}$ to match the inference~$\ijoint{\x}{\z}$.

\textbf{Ablation of~$\lossh$.} Removing~$\lossh$ slightly increased the reconstruction error but deteriorated the generation quality.
To confirm this behavior, we also show the samples generated by the GWAE model without the regularization~$\lossh$ in \cref{fig:generation-ablation} and its reconstruction in \cref{fig:reconstruction-ablation}.
These qualitative results that the decoder without~$\lossh$ successfully reconstructs the images from the inference~$\iagg{\z}$ but generates corrupted images from the prior~$\trainablePrior{\z}$.
It suggests the ``hole'' problem~\cite{TamingVAEs} in the degenerate solution, where each data point is mapped at a single latent point to cover the zero-measure area of the latent space and the latent space is almost everywhere not covered by the inference~$\iagg{\z}$.
Thus, the entropy regularization~$\lossh$ seems to have worked for retaining the probabilistic mappings in the encoder~$\enc{\z}{\x}$ to avoid this phenomenon.
}

In addition, for ablating $\gworder=1$, we also experimented with the $\gworder=\worder$ settings that appear to be intuitively natural although causing an unstable training process due to the outlier samples in $\lossgw$.
The GWAE model with $\gworder=\worder$ suffered from performance degradation both in the generation and reconstruction, suggesting that our settings~$\gworder=1 \le \worder$ affect the learning process of the entire model.

\subsection{The Meta-Prior Effect on GW Minimization and Estimation}
\begin{figure}[!t]
    \centering
    \hspace*{\fill}
    \begin{minipage}[t]{\linewidth}
        \centering
        \includegraphics[width=\linewidth,clip]{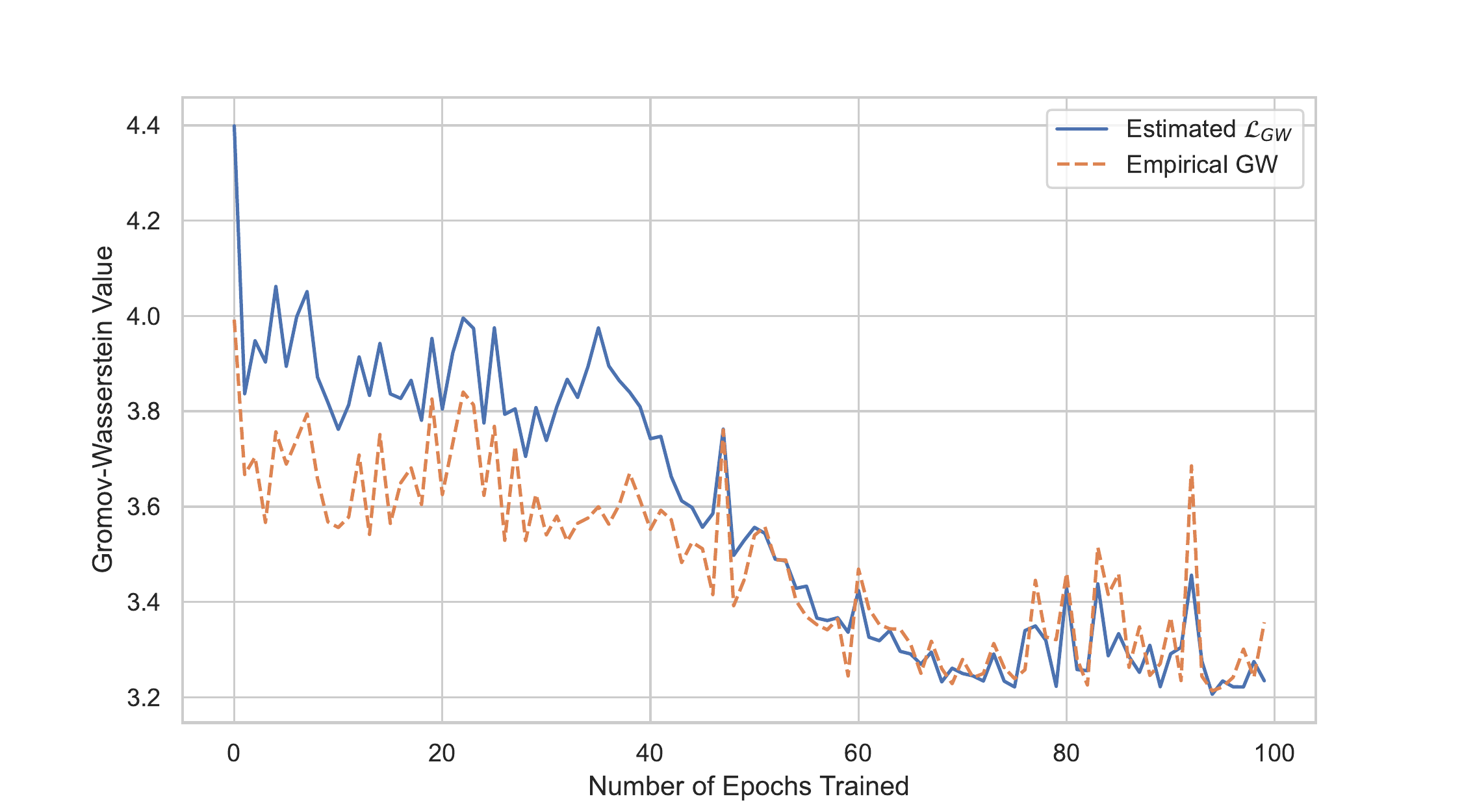}
        \subcaption{The estimation of the GW metric using FNP.}
        \label{fig:gw-estmin-fnp-est}
    \end{minipage}
    \hfill
    \begin{minipage}[t]{\linewidth}
        \centering
        \includegraphics[width=0.7\linewidth,clip]{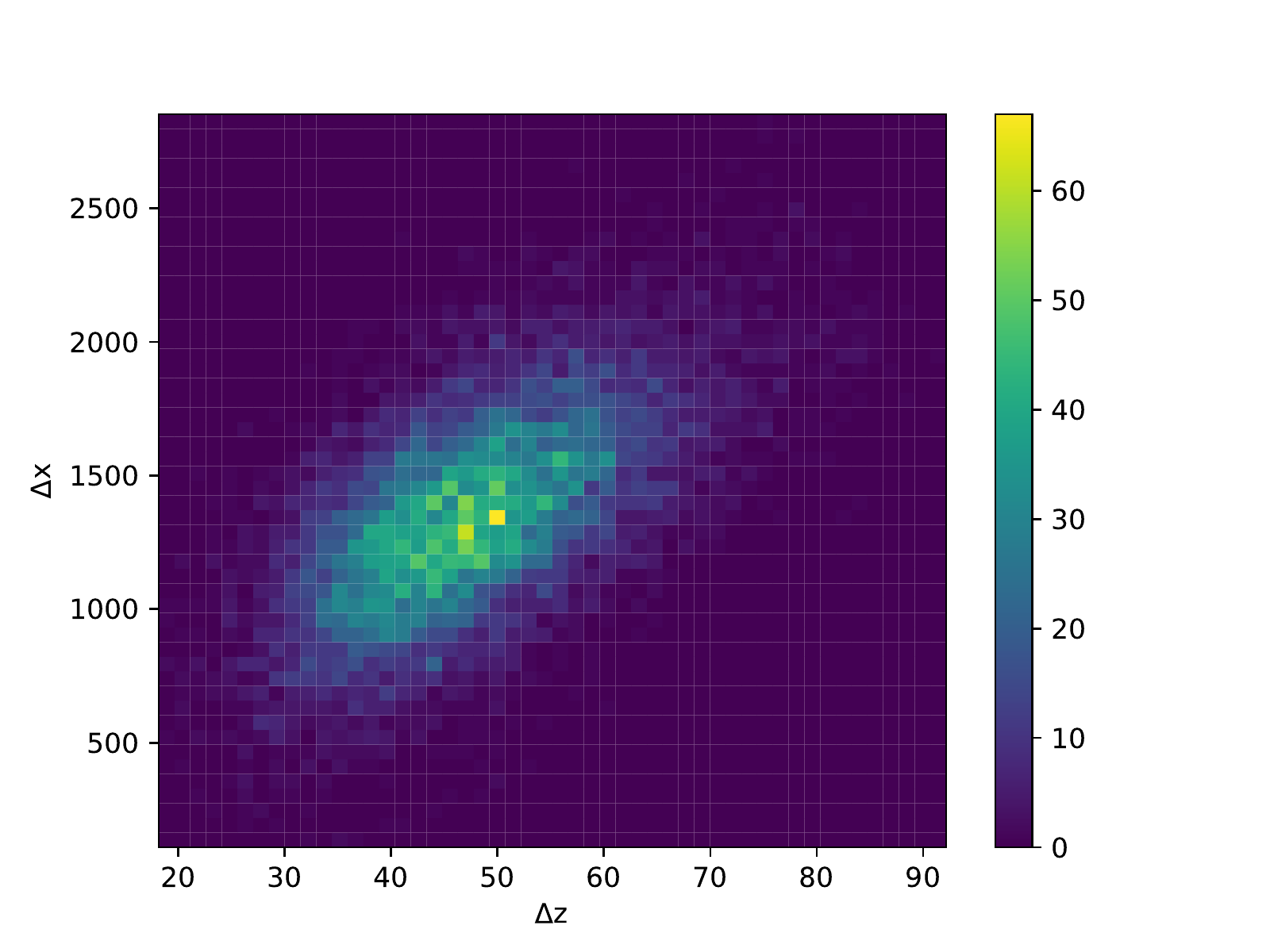}
        \subcaption{The isometry in GWAE with FNP.}
        \label{fig:gw-estmin-fnp-min}
    \end{minipage}
    \hspace*{\fill}
    \caption{
        The estimation and minimization of the GW metric. This trial of training is conduced in GWAE~(\textbf{FNP}, $\coeffd$=1, $\coeffw$=1, $\coeffh$=1) using the MNIST~\cite{MNIST} dataset, which is the same settings as \cref{fig:gw-estmin} except for FNP.
            (a) The curves show the GW values estimated by the loss term~$\lossgw$ (solid, blue) and the empirical GW computed by the POT package~\cite{Flamary2021POT} (dashed, orange).
            The values are computed using the validation set.
            (b) The axes~$\Delta x = \distx(\x,\x')$~(vertical) and $\Delta z = \distz(\z, \z')$~(horizontal) respectively denote the difference in the data and latent spaces between generated samples~$(\x,\z), (\x',\z') \sim \gjoint{\x}{\z}$.
            The histogram contains 10,000 generated sample pairs.
    }
    \label{fig:gw-estmin-fnp}
\end{figure}

For a further inspection of \cref{sec:experiments-isometry}, we also studied the GW minimization and estimation using FNP in \cref{fig:gw-estmin-fnp}.
Compared with the NP case in \cref{fig:gw-estmin}, GWAE with FNP presents less stable and more biased estimation and minimization.
The learning curve of $\lossgw$ in \cref{fig:gw-estmin-fnp-est} is largely biased in the first 40 epochs and then seems to be converged at approximately 3.2, a higher value than that of \cref{fig:gw-estmin-est} (lower than 2).
The isometry histogram also suggests the degradation of GW minimization in FNP.
In \cref{fig:gw-estmin-fnp-min}, more samples fell in off-diagonal areas, showing that the isometry is less tight than that of \cref{fig:gw-estmin-min}.
These results are presumably due to the mismatch of disentanglement meta-prior in MNIST~\cite{MNIST} because one of the major generative factors of MNIST images is the kind of digits, a categorical variable typically learned as one-hot variables in contrast to the factorization imposed by FNP.

\subsection{Priors in Clustering Structure} \label{sec:appendix-ood-prior}
\begin{figure}[!t]
    \centering
    \begin{minipage}[t]{0.8\linewidth}
        \centering
        \includegraphics[width=0.49\linewidth,clip]{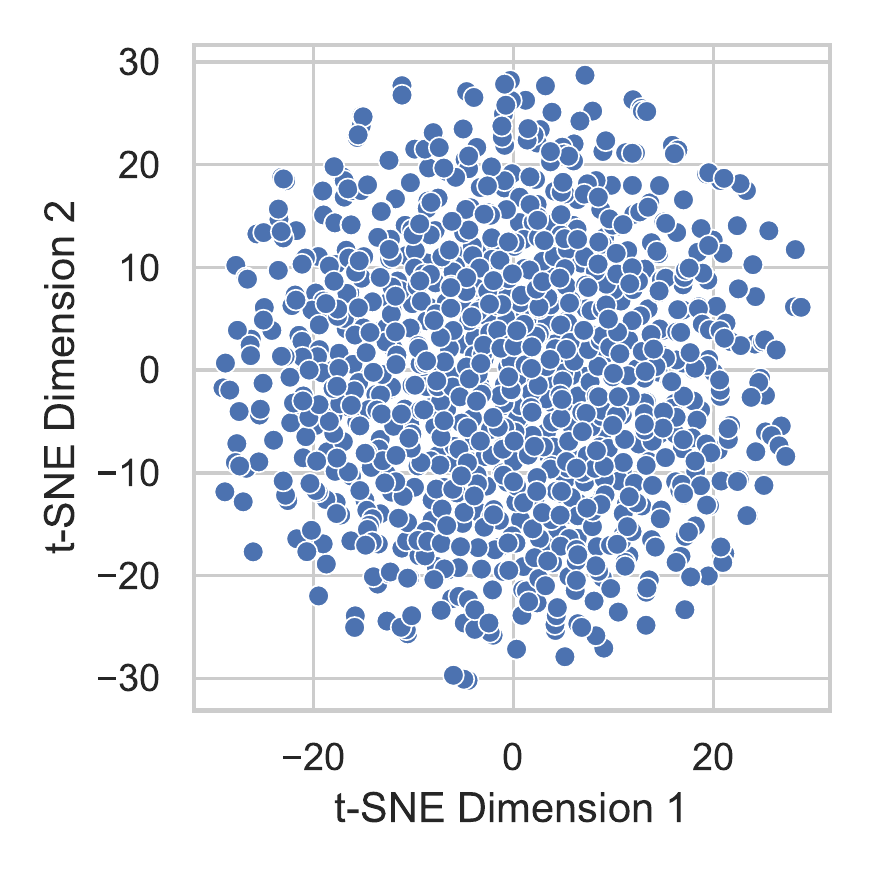}
        \includegraphics[width=0.49\linewidth,clip]{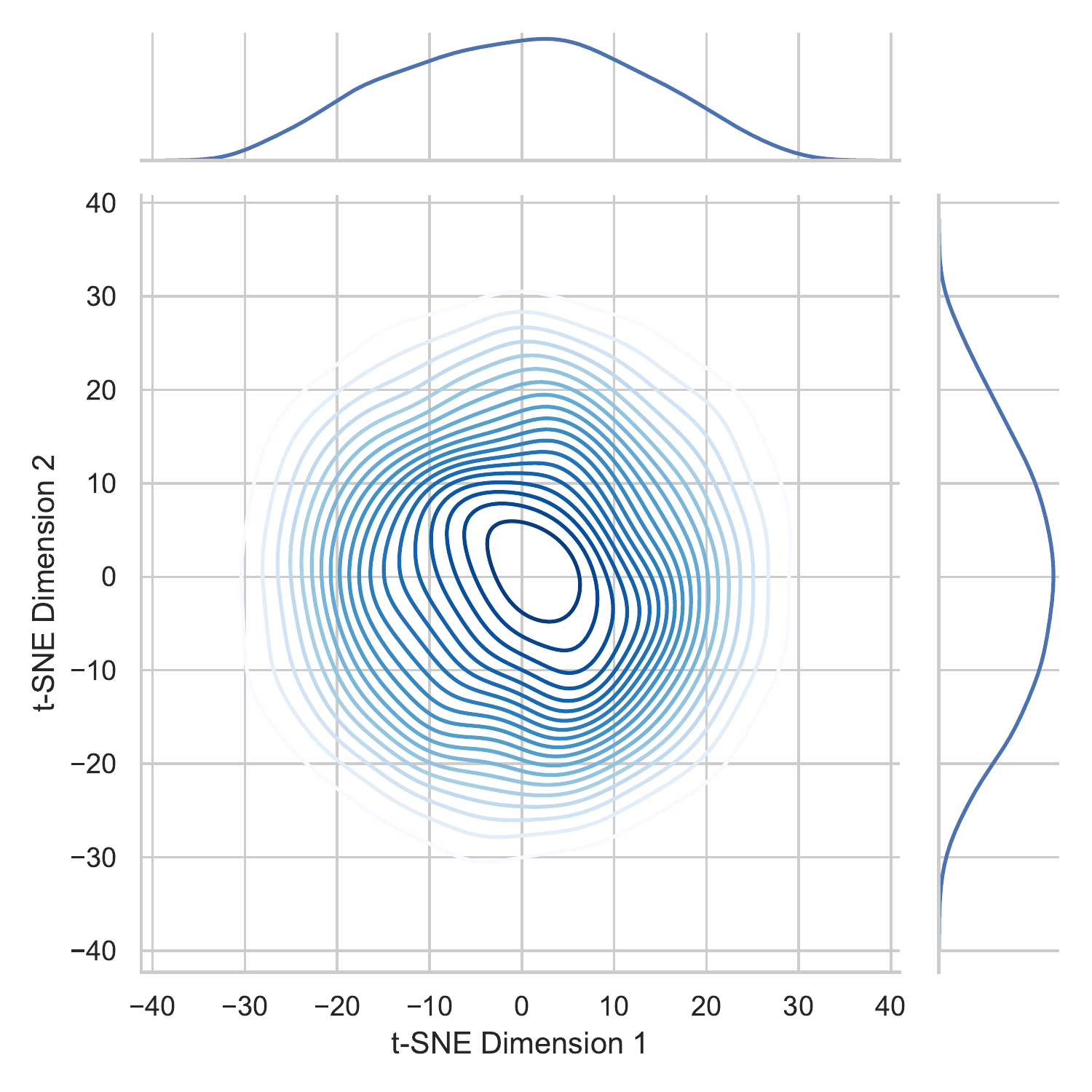}
        \vspace{-0.3em}
        \subcaption{VAE~\cite{VAE}.}
        \vspace{1em}
    \end{minipage}
    \begin{minipage}[t]{0.8\linewidth}
        \centering
        \includegraphics[width=0.49\linewidth,clip]{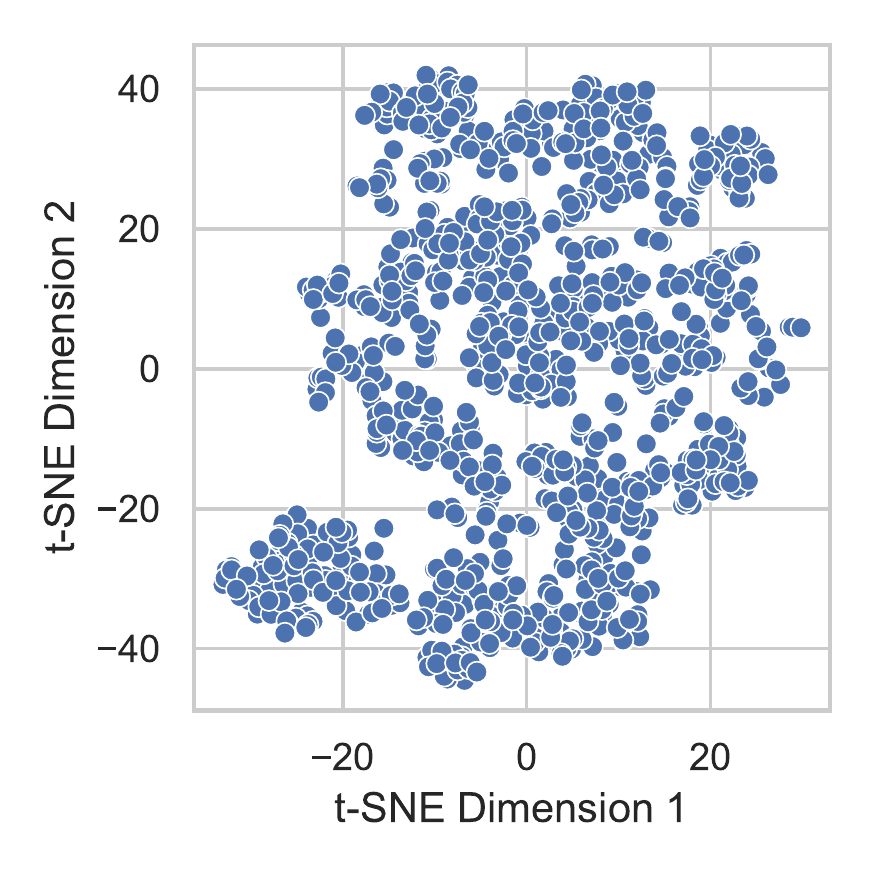}
        \includegraphics[width=0.49\linewidth,clip]{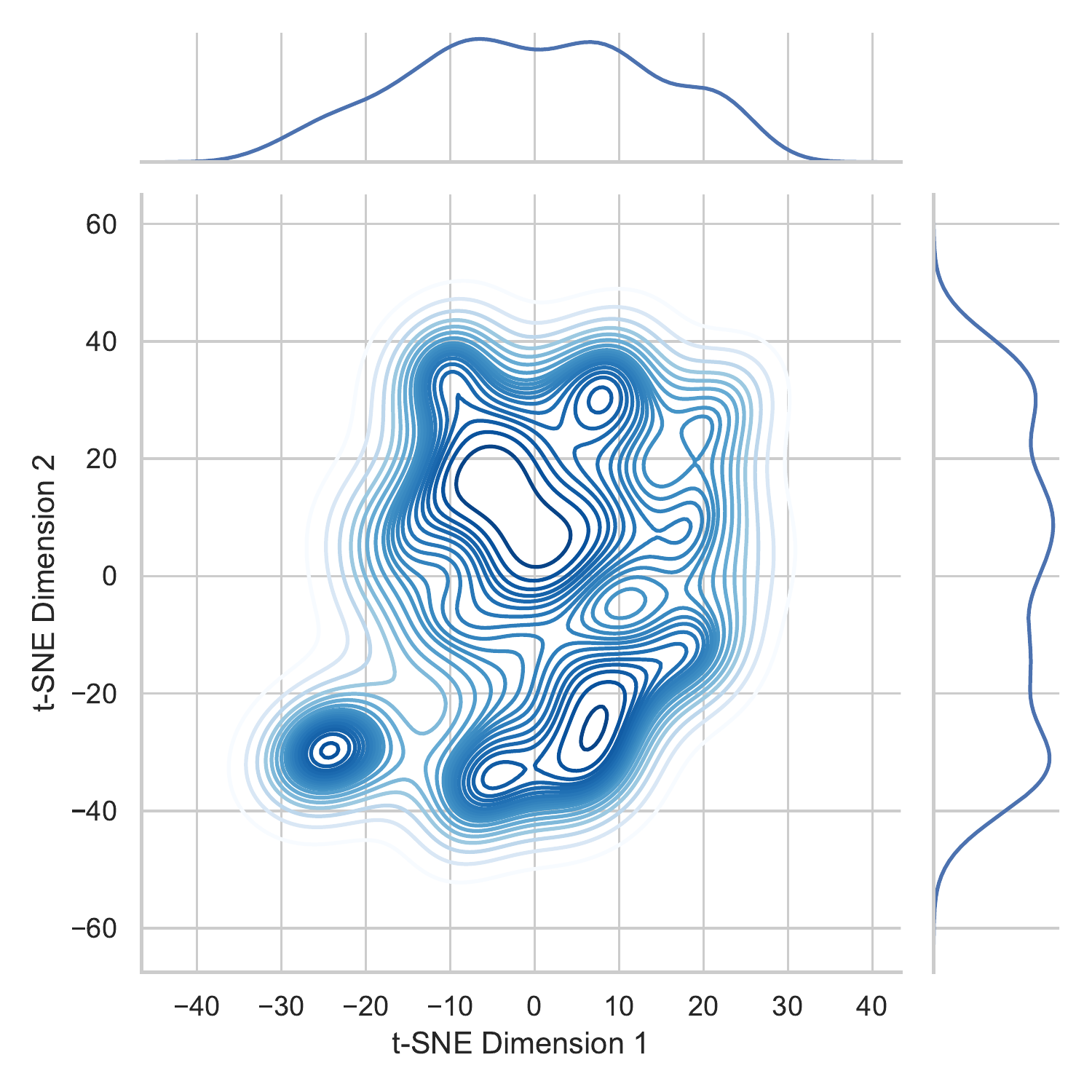}
        \vspace{-0.3em}
        \subcaption{DAGMM~\cite{DAGMM}.}
        \vspace{1em}
    \end{minipage}
    \begin{minipage}[t]{0.8\linewidth}
        \centering
        \includegraphics[width=0.49\linewidth,clip]{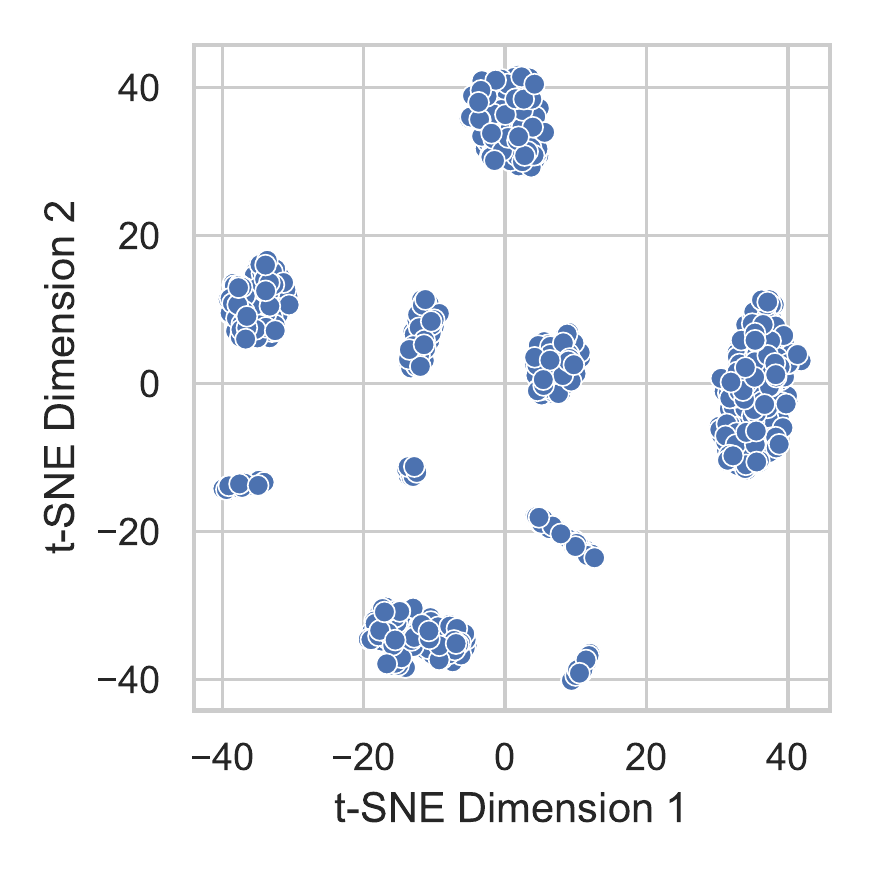}
        \includegraphics[width=0.49\linewidth,clip]{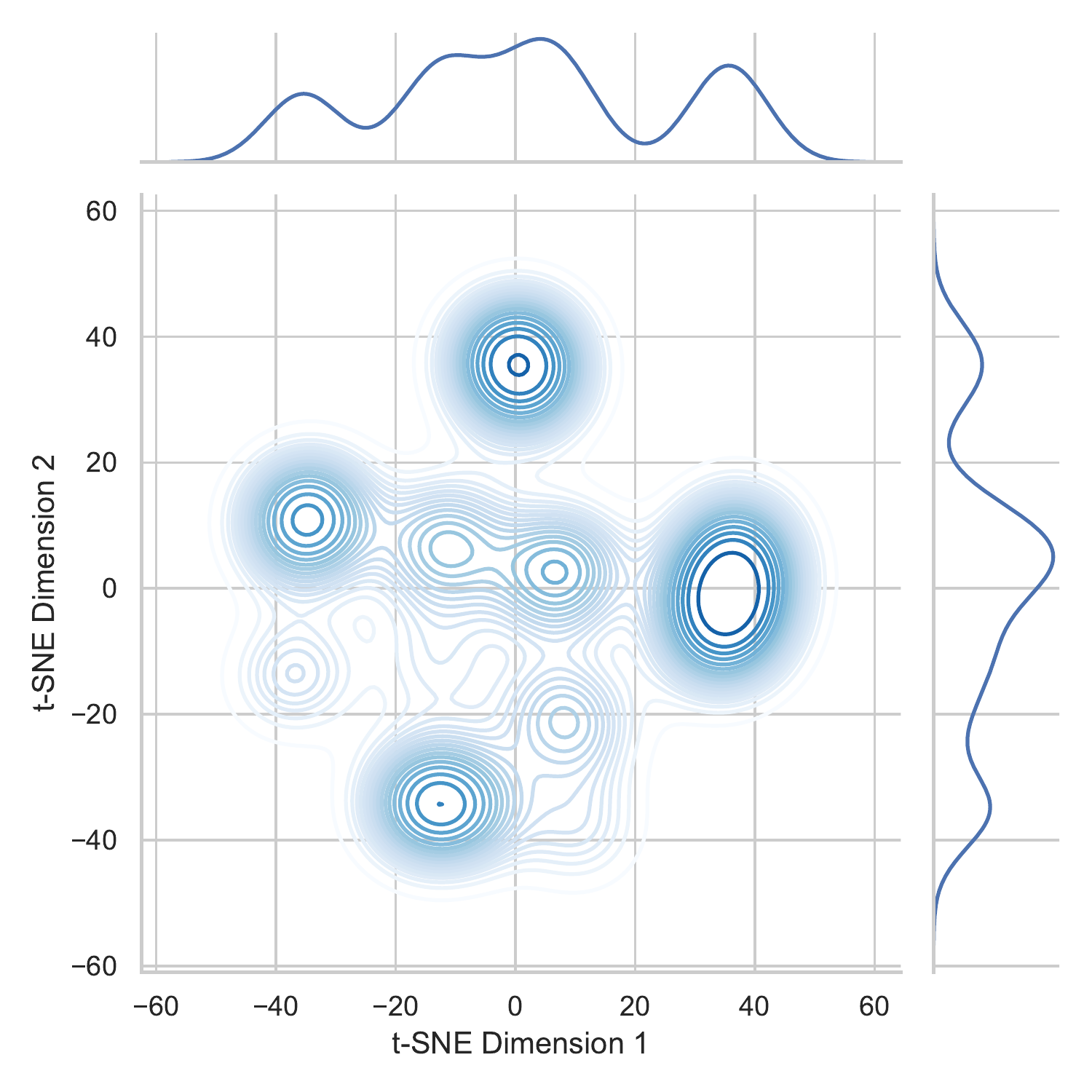}
        \vspace{-0.3em}
        \subcaption{GWAE~(GMP).}
        \vspace{1em}
    \end{minipage}
    \caption{
        The t-SNE visualizations~\cite{tSNE} for latent space samples~$\z\sim\trainablePrior{\z}$ for the OoD detection in \cref{fig:ood}.
        The left plot presents the sampled points of the t-SNE embeddings, and the right one presents the kernel density estimation (KDE) of these embeddings.
        The sample size is equally 1,024 in each reported model.
    }
    \label{fig:ood-priors}
\end{figure}

For more detailed investigation of the capture of clustering structure studied in \cref{fig:ood}, we further study the latent spaces of VAE~\cite{VAE}, DAGMM~\cite{DAGMM}, and GWAE with GMP.
The t-SNE visualization~\cite{tSNE} of the latent spaces are shown in \cref{fig:ood-priors}, which suggests that the GWAE model with GMP clearly captured the clustering structure in its latent space.
The prior of VAE~\cite{VAE} is defined as the standard Gaussian~$\mathcal{N}(\bm{0},\mathbf{I}_L)$ which does not consist of multiple clusters.
The learned prior of DAGMM contains multiple clusters; however, adjacent clusters were overlapping to some extent.
From the learned prior in GWAE, we can observe clear clusters densely concentrating themselves and separating each other.
These results support the quantitative OoD results in \cref{fig:ood}, in which the GWAE model outperforms the other two models with and without explicit clustering modeling, respectively.

\end{document}